%% file: main.tex
\documentclass[nofootinbib,%
 reprint,
 amsmath,amssymb,
 aps,
]{revtex4-2}
\usepackage{bbm}
\usepackage[dvipsnames]{xcolor}
\usepackage{hyperref}
\usepackage{soul}
\usepackage[most]{tcolorbox}
\usepackage{graphicx}
\usepackage{refcount}
\usepackage{amsmath}
\usepackage{mathtools}
\usepackage{makecell}
\usepackage{tikz}
\usepackage{footnote}
\usetikzlibrary{plotmarks}
\tikzset{>=latex}
\usetikzlibrary{arrows.meta}
\usetikzlibrary{positioning}
\usepackage{dcolumn}
\usepackage{bm}

\usepackage{pgfplots}
\usepackage{pgfplotstable}
\usepackage{algorithm,algpseudocode}
\PassOptionsToPackage{noend}{algpseudocode}
\usepackage{algpseudocode}
\renewcommand{\algorithmiccomment}[1]
{\bgroup\hfill$\blacktriangleright$~#1\egroup}

\makeatletter
\def\BState{\State\hskip-\ALG@thistlm}
\makeatother
\algnewcommand{\algorithmicforeach}{\textbf{for each}}
\algdef{SE}[FOR]{ForEach}{EndForEach}[1]
  {\algorithmicforeach\ #1\ }
  {\algorithmicend\ \algorithmicforeach}
\usetikzlibrary{shapes.misc}
\begin{document}

\preprint{APS/123-QED}

\title{Towards a framework on tabular synthetic data generation: a minimalist approach: theory, use cases, and limitations}
\thanks{The views expressed in the paper are those of the authors and do not represent the views of Wells Fargo}%

 \author{Yueyang Shen}
 \altaffiliation[]{Computational Medicine and Bioinformatics, Statistical Online Computational Resources, University of Michigan; this work was conducted while the author was a summer intern at Wells Fargo}
  \email{petersyy@umich.edu}
  \author{Agus Sudjianto}
 \altaffiliation[]{H2O.ai and University of North Carolina at Charlotte; this work was conducted while the author was at Wells Fargo. }
\author{Arun Prakash R, Anwesha Bhattacharyya, Maorong Rao, Yaqun Wang, Joel Vaughan, Nengfeng Zhou}%
\altaffiliation[]{Wells Fargo }

\date{\today}



\input{abstract.tex}

\maketitle

\input{intro_backg.tex}

\input{theo_foundation.tex}

\input{related_work.tex}
\input{evaluation_metrics.tex}

\input{experiments.tex}

\input{case_study.tex}

\section{Conclusion}
In this study, we introduced a minimalist framework for synthetic tabular data generation leveraging SparsePCA for encoding and XGBoost for nonlinear recovery. This approach demonstrates simplicity, interpretability, and computational efficiency, making it particularly suited for high-dimensional and heterogeneous tabular data. Through rigorous evaluations on both low-dimensional toy datasets and high-dimensional simulated credit data, we highlighted the advantages and limitations of the proposed method.

The SparsePCA+XGBoost pipeline provides an interpretable mechanism for feature compression and data reconstruction, avoiding the complexity and overfitting challenges often associated with autoencoders and variational autoencoders. Additionally, the model-based perturbation strategy introduced in this work offers a robust alternative to raw and quantile-based perturbations for downstream robustness testing. This framework ensures flexibility while maintaining the fidelity of the synthetic data to the original distribution.

Despite its strengths, our approach has certain limitations, particularly in handling isotropic and symmetric structures during encoding, which may lead to information loss in specific data scenarios. Future work will aim to address these limitations by exploring more sophisticated encoding techniques, such as incorporating independent component analysis (ICA) or leveraging nonlinear disentanglement methods. Moreover, extending the framework to seamlessly handle diverse data types and distributions remains an open challenge.

Overall, the proposed methodology presents a practical and effective solution for generating synthetic tabular data, with applications ranging from data privacy and robustness testing to model evaluation and beyond. It strikes a balance between simplicity and performance, making it a valuable tool for practitioners in data-centric fields.

\input{appendix}

\nocite{*}

\bibliography{apssamp}

\end{document}

%% file: abstract.tex
\begin{abstract}
\textbf{Abstract:} We propose and study a minimalist approach towards synthetic tabular data generation. The model consists of a minimalistic unsupervised\footnote{From the perspective of linear model $x=Wz+e,$ PCA is the optimal self-supervised error construction algorithm.} SparsePCA encoder (with contingent clustering step or log transformation to handle nonlinearity) and Xgboost decoder which is SOTA for structured data regression and classification tasks \cite{borisov2022deep}. We study and contrast the methodologies with (variational) autoencoders in several toy low dimensional scenarios to derive necessary intuitions. The framework is applied to high dimensional simulated credit scoring data which parallels real-life financial applications. We applied the method to robustness testing to demonstrate practical use cases. The case study result suggests that the method provides an alternative to raw and quantile perturbation for model robustness testing. We show that the method is simplistic, guarantees interpretability all the way through, does not require extra tuning and provide unique benefits.

\end{abstract}

%% file: intro_backg.tex
\section{Introduction \& Backgrounds}
Intelligent machines cannot be based purely on decision-making machines, it must be endowed with the ability to reason, learn succinct representation that resembles underlying physical processes, and express understanding and uncertainty. The framework that gives machines such abilities can often be cast as a generative modeling problem. The current generative framework can be classified into four categories \cite{tomczak2021deep}: 1) Autoregressive models. 2) Flow Based Models. 3) Latent variable compression-decompression formulated with information bottleneck. 4) Energy based models.

The lossy compression-decompression modeling offers stable, potentially robust feature representation invariant to noise and nuance parameter \cite{achille2018emergence} by enforcing on an information bottleneck \cite{tishby2015deep}. This paradigm parallels many machine learning two-stage pipelines: representation learning (feature extraction) and downstream optimization. This former process of extracting summary statistics as a lower dimensional feature from raw dimension can often be re-casted as a dimension reduction problem.  Some of the technologies that tackles this problem are summarized in the following table (Table \ref{tab:dimredbenchmark}) \cite{kruskal1964multidimensional,sammon1969nonlinear,tenenbaum2000global,belkin2003laplacian,silva2002global,huang2022towards,ghojogh2020multidimensional}.
\begin{table*}
\caption{\label{tab:dimredbenchmark}A short benchmark comparison amongst common (unsupervised) dimension reduction algorithms}
\begin{ruledtabular}
\begin{tabular}{cccccccc}
 &\multicolumn{7}{c}{Dimension Reduction Algorithms}\\
\cline{2-8}
 Features&PCA&Classical MDS & Sammon Mapping&tSNE &Isomap &UMAP &Laplacian Eigenmap\\ \hline
 Matrix Factorization&\checkmark &\checkmark&&&&&\checkmark \\
 Nonlinear&&&\checkmark&\checkmark&\checkmark&\checkmark&\checkmark\\
 Graph construction&&&&\checkmark&\checkmark&\checkmark&\checkmark\\
 Robust to parameter choices&\checkmark&\checkmark&\checkmark&&\checkmark&&\\
 Global Structure Preservation& \checkmark&\checkmark&&&\checkmark&\checkmark&\\
\end{tabular}
\end{ruledtabular}
\end{table*}
\\
\indent From the latent standpoint, linear methodologies learn a low-dimensional linear subspace projection assuming that the latent feature spans a linear subspace, while the nonlinear methods employ nonlinear projection that preserves local information along some low dimensional data manifold, assuming the validity of the manifold hypothesis \cite{fefferman2016testing}. The latter is often referred to as manifold learning. Local approaches preserve the local geometry of raw observational data by mapping nearby points on the manifold to nearby points in the low dimensional latent space, while global approaches preserve geometry and structure not only for nearby points in high dimensional space but also for faraway points. Many algorithms (e.g. tSNE\cite{van2008visualizing}, UMAP\cite{mcinnes2018umap}) can also be sensitive to parameter (and hence to preprocessing) choices leading to loss of meaningful informational structures in the data \cite{huang2022towards}.

In this short manuscript, we review and investigate the empirical features for encoder-decoder frameworks operating on tabular data: (variational) autoencoders \cite{kingma2013auto}, with a special emphasis on a sparsePCA\cite{wold1987principal,d2004direct}+XGboost\cite{chen2016XGboost} framework. We provide concrete use cases for the proposed algorithm and demonstrate its use cases in robustness testing. Other encoding choices' performance is left as future work. The key feature of the proposed sparsePCA+XGboost pipeline is that it employs a minimalist approach with few inductive bias on the manifold and guarantees interpretability all the way through.

It is a well-known fact in the community that linear and logistic regression almost surely cannot beat XGboost in benchmarks, however the simpler models offer unique advantages in terms of interpretability, parsimony, trustworthiness, and sheds light on the underlying data generating processes. Likewise, we won't argue the proposed sparsePCA+XGboost method consistently outperform over autoencoders/VAEs on homogeneous data generation. However, we provide explicit case scenarios (especially in tabular cases where VAE deployment requires tuning and extra care in face of heterogeneous tabular data), highlighting the benefits of simplicity (and hence less overfitting) and interpretability.

\begin{figure}[!htb]
\includegraphics[width=8.5cm,height=5cm]{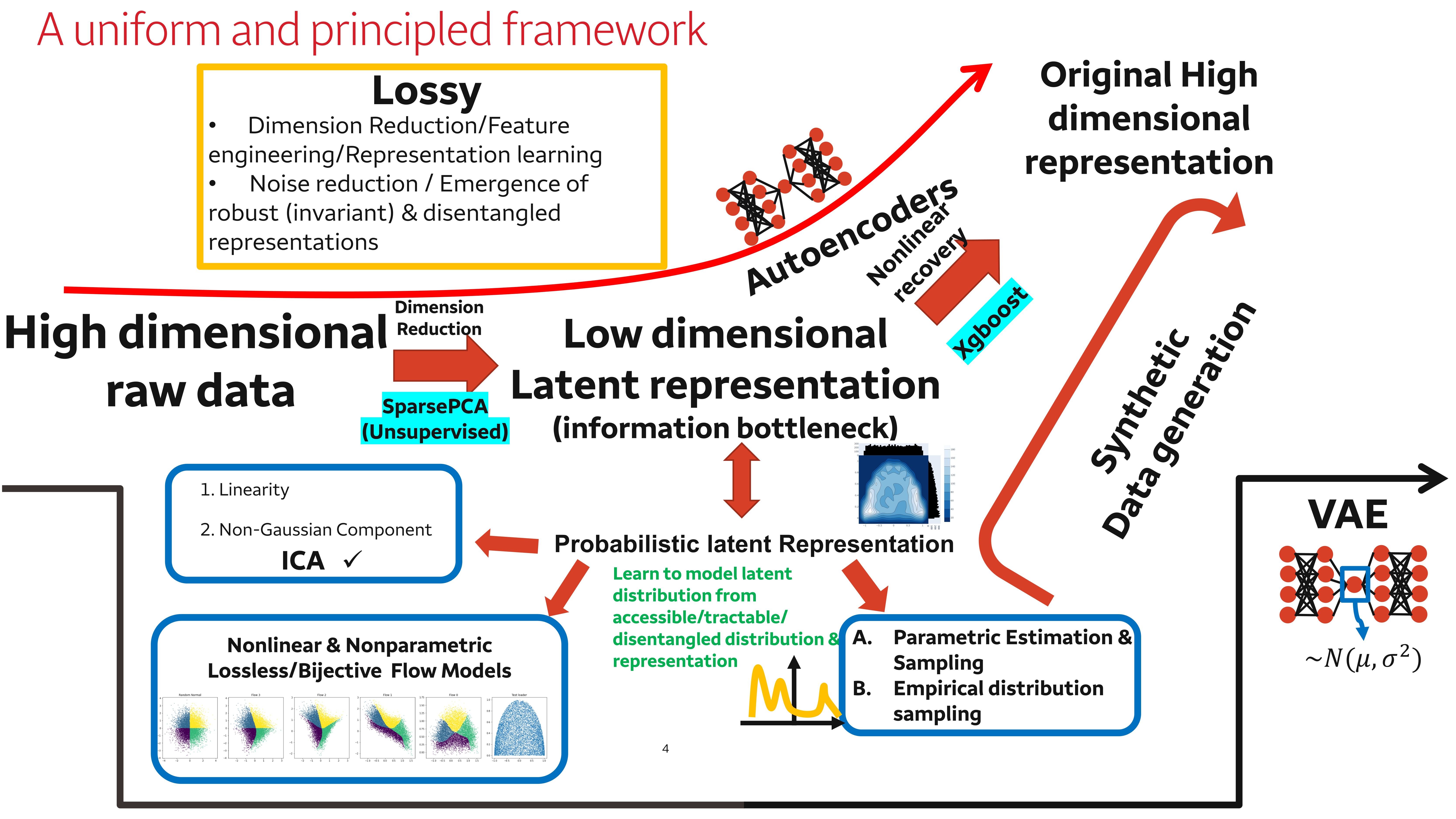}
\caption{\label{fig:landscape} The workflow and landscape of latent representation learning for synthetic data generation.}
\end{figure}

To generate synthetic tabular data, we often first need to compress information from high dimensional data. This can be done either via (unsupervised) dimension reduction followed by nonlinear recovery or straightforwardly via autoencoders (self-supervised). To generate synthetic data, not only do we need a latent representation, we also need a mechanism to construct a probabilistic sampling scheme from the latent space. From sampling the latent distribution and feeding the sampled result into a nonlinear recovery method, we obtain instances of synthetic data that mimics the original data (See Fig \ref{fig:landscape}).

\subsection{Proposed Algorithm}
We propose and study a minimalist approach towards synthetic tabular data generation. The model consists of a minimalistic unsupervised SparsePCA encoder (with contingent clustering preprocessing step to handle nonlinearity) and XGboost decoder which is SOTA for structured data regression and classification tasks \cite{borisov2022deep}\footnote{We are also aware that some benchmarks in a followed up study \cite{gorishniy2023tabr} in the paper writing process}. The proposed minimalist framework is as follows (Algorithm \ref{proposed-alg})
\input{proposed_algo.tex}

\subsection{Possible Use Cases \label{mod_algo}}
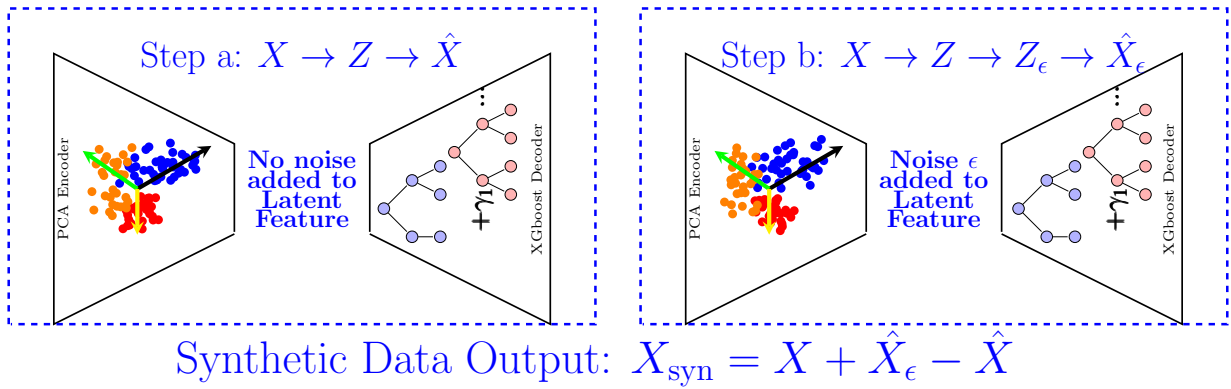
\begin{figure*}
    \input{figures/robust_testing}
    \caption{The robustness analysis pipeline for computational tractability in perturbation and to facilitate robustness analysis.}
    \label{robust_alg}
\end{figure*}
The above algorithmic framework provides an obfuscation strategy to desensitize the dataset $X$, but the sampling from uniform distribution condition in Step 4 compromises computation tractability for robustness testing. For controllable robustness testing and preservance of individual row-instance characteristic, we propose an alternative Step \textcolor{blue}{$4^*$} by inputting the PCA latents with added noise of size $\epsilon$ into the trained XGboost obtaining a dataset $\hat{X}_{\epsilon}$. The dataset $\hat{X}_{\epsilon}$ preserves correspondence in each row entries. Furthermore, we add back the residuals obtaining $X+\hat{X}_{\epsilon}-\hat{X}_{\epsilon=0}=\hat{X}_{\epsilon}+(X-\hat{X})$\footnote{In the following experiments reported in the paper, we use log transformation instead of clustering to handle nonlinearity.}, which is used for robustness testing. Notably, when $\epsilon=0$, no perturbation is applied. \textit{The epsilon factor effectively becomes a controllable factor to control the amount of model-based noise applied to the dataset} (See Figure \ref{robust_alg} for a schematic). 

%% file: proposed_algo.tex
\begin{minipage}{0.5\textwidth}
\renewcommand*\footnoterule{}
\begin{algorithm}[H]
  \caption{Proposed Algorithm Framework}
  \label{proposed-alg}
  \hspace*{\algorithmicindent} \textbf{Input}: Input tabular data matrix $\pmb{X}\in \mathbb{R}^{n\times p}$, Cluster size $c$, Latent dimension $l$ \\
 \hspace*{\algorithmicindent} \textbf{Output}: Output synthetic data $\pmb{\tilde{X}},\pmb{X_{\text{syn}}}\in \mathbb{R}^{n\times p}$
  \begin{algorithmic}[1]\\
\textit{\underline{Interpretable Nonlinearity handling}}: Apply \textbf{clustering} to kill isotropicity and deal with nonlinearity.\\
\textit{\underline{Interpretable Minimal Compression}}: \textbf{For each }cluster, apply sparsePCA to learn a linear subspace.\\
\textit{\underline{Interpretable Nonlinear Recovery}}: \textbf{For each }cluster $i\in\{1,2,...,c\}$ use $p$ (regularized) XGboost trees to learn a nonlinear recovery maps $f_{i,1},f_{i,2},...,f_{i,p}$ from the latent space to the original dataset.\footnote{Adding noise to the input latent space as in VAE can improve the generation quality but sacrifice reconstruction quality.}\\
\textit{\underline{Synthetic data generation} (Ex post density estimation)}:\textbf{For each} cluster $i$ obtain empirical transformations $g_i$ to transform observed latents $\pmb{Z_i}$ to a uniform hypercube $H:[0,1]^l$, $g_i(\pmb{Z_i})\subset H$, where $l$ is the latent dimension.
\par\noindent \textbf{\textit{(a)}} Sample from the uniform distribution $u_j\sim (Unif[0,1])^l$, $\pmb{u}=[u_1,...,u_{s_i}], 1\leq j\leq s_i$ and inverse transform the observed sample back to latent space $\pmb{\hat{Z_i}}=g_i^{-1}(\pmb{u})$, where  the sample size of $s_i$ are proportional to the cluster sizes. \footnote{Other alternative approaches to capture joint relationships are reviewed in \ref{contrasts}.}
\par\noindent \textbf{\textit{(b)}} Obtain $\pmb{\tilde{X}}=(f_{i,1}(\hat{\pmb{Z_i}}),f_{i,2}(\hat{\pmb{Z_i}}),...,f_{i,p}(\hat{\pmb{Z_i}}))_{1\leq i\leq c}$ via XGboost.
\setcounter{ALG@line}{3}
\makeatletter
\algrenewcommand\alglinenumber[1]{\textcolor{blue}{#1*}} 
\makeatother
\State \textit{\underline{Controllable Robustness Testing} }
\par\noindent \textbf{\textit{(a)}} Fit an XGboost model for each cluster latent $\pmb{Z_i}$ to the target $\pmb{X_i}$. Denote the fitted XGboost models as $h_{i}(\cdot)$, the reconstruction residual is $\pmb{\hat{e}}_i=\pmb{X_i}-\pmb{\hat{X_i}}=\pmb{X_i}-h_{i}(\pmb{Z_i})$
\par\noindent\textbf{\textit{(b)}} Adding noise with size $\pmb{\epsilon}$ to the latents $\pmb{Z_i}$, generating $\pmb{Z_{i,\epsilon}}$. Output the synthetic data for robustness testing with $\pmb{X_{\text{syn}}}=(h_i(\pmb{Z_{i,\epsilon}})+\pmb{\hat{e_i}})_{1\leq i\leq c}=(\hat{\pmb{X_{i,\epsilon}}}+\pmb{X_i}-\pmb{\hat{X_i}})_{1\leq i\leq c}$

\end{algorithmic}

\end{algorithm}

\end{minipage}

%% file: figures/robust_testing.tex
\vspace{4cm}
\begin{center}
    \begin{tikzpicture}[>=stealth,
    shorten >=1pt,
    node distance=2cm,
    on grid,
    auto,
    scale=0.75,
    every node/.style={scale=1},
    align=center,
    font=\scriptsize,
   level distance=1cm,
  level 1/.style={sibling distance=2cm},
  level 2/.style={sibling distance=1cm}
]
\begin{scope}[shift={(-7.5,-0.7)},transform canvas={scale=0.5},rotate=90]
\node [circle, draw, fill=blue!30] {}
  child {node [circle, draw, fill=blue!30] {}
    child {node [circle, draw, fill=blue!30] {}}
  }
  child {node [circle, draw, fill=blue!30] {}
    child {node [circle, draw, fill=blue!30] {}}
    child {node [circle, draw, fill=blue!30] {}}
  };
  \begin{scope}[shift={(2,-2.5)}]
    \node[rotate=90] at (-2,-1) {\huge $\pmb{+\gamma_1}$};
   \node [circle, draw, fill=red!30] {}
  child {node [circle, draw, fill=red!30] {}
    child {node [circle, draw, fill=red!30] {}}
    child {node [circle, draw, fill=red!30] {}}
  }
  child {node [circle, draw, fill=red!30] {}
    child {node [circle, draw, fill=red!30] {}}
    child {node [circle, draw, fill=red!30] {}}
  };
  \node[rotate=90]  at (2,-1) {\huge $\pmb{...}$};
  \end{scope}
\end{scope}
\begin{scope}[shift={(15,-0.7)},transform canvas={scale=0.5},rotate=90]
\node [circle, draw, fill=blue!30] {}
  child {node [circle, draw, fill=blue!30] {}
    child {node [circle, draw, fill=blue!30] {}}
  }
  child {node [circle, draw, fill=blue!30] {}
    child {node [circle, draw, fill=blue!30] {}}
    child {node [circle, draw, fill=blue!30] {}}
  };
  \begin{scope}[shift={(2,-2.5)}]
    \node[rotate=90] at (-2,-1) {\huge $\pmb{+\gamma_1}$};
   \node [circle, draw, fill=red!30] {}
  child {node [circle, draw, fill=red!30] {}
    child {node [circle, draw, fill=red!30] {}}
    child {node [circle, draw, fill=red!30] {}}
  }
  child {node [circle, draw, fill=red!30] {}
    child {node [circle, draw, fill=red!30] {}}
    child {node [circle, draw, fill=red!30] {}}
  };
  \node[rotate=90]  at (2,-1) {\huge $\pmb{...}$};
  \end{scope}
\end{scope}
\begin{scope}[shift={(-8,0)},transform canvas={scale=0.8}]
\begin{scope}[shift={(-7,-1)},transform canvas={scale=0.7}]
\foreach \i in {1,...,40} {
        \fill[blue, mark size=15pt] (1+rnd*2,2+rnd*1,1+rnd*2) circle (4pt);
    }
    \foreach \i in {1,...,30} {
        \fill[red, mark size=15pt] (0.3+rnd,rnd,rnd) circle (4pt);
    }
    \foreach \i in {1,...,30} {
        \fill[orange, mark size=15pt] (-1+rnd*1,rnd*2,-1+rnd) circle (4pt);
    }

    \draw[line width=1mm,black,->] (0.5,1,0) -- (2.5,2,-1) ;
    \draw[line width=1mm,green,->] (0.5,1,0) -- (-0.5,3,2) ;
    \draw[line width=1mm,yellow,->] (0.5,1,0) -- (0.5,-0.5,0);
\end{scope}
\draw[thick] (0,-1) -- (-4,-3) -- (-4,3) -- (0,1) -- (0,-1);
\draw[dashed, very thick, blue] (-5,-3) -- (-5,4) -- (8,4) -- (8,-3) -- (-5,-3);
\draw[thick] (3,-1) -- (7,-3) -- (7,3) -- (3,1) --  (3,-1);

\node at (-3.8,0)[rotate=90] {PCA Encoder};
\node at (6.7,0) [rotate=90]{XGboost Decoder};
\node at (1.5,0) [blue, thick]{\large\textbf {No noise}\\\large\textbf{added to}\\\large\textbf {Latent}\\ \large\textbf{Feature}};
\node at (1.5,3) [blue]{\LARGE Step a: $X\to Z\to \hat{X}$};
\end{scope}

\begin{scope}[shift={(6,0)},transform canvas={scale=0.8}]
\begin{scope}[shift={(-1,-1)},transform canvas={scale=0.7}]
\foreach \i in {1,...,40} {
        \fill[blue, mark size=15pt] (1+rnd*2,2+rnd*1,1+rnd*2) circle (4pt);
    }
    \foreach \i in {1,...,30} {
        \fill[red, mark size=15pt] (0.3+rnd,rnd,rnd) circle (4pt);
    }
    \foreach \i in {1,...,30} {
        \fill[orange, mark size=15pt] (-1+rnd*1,rnd*2,-1+rnd) circle (4pt);
    }

    \draw[line width=1mm,black,->] (0.5,1,0) -- (2.5,2,-1) ;
    \draw[line width=1mm,green,->] (0.5,1,0) -- (-0.5,3,2) ;
    \draw[line width=1mm,yellow,->] (0.5,1,0) -- (0.5,-0.5,0);
\end{scope}
\draw[thick] (0,-1) -- (-4,-3) -- (-4,3) -- (0,1) -- (0,-1);
\draw[dashed, very thick, blue] (-5,-3) -- (-5,4) -- (8,4) -- (8,-3) -- (-5,-3);
\draw[thick] (3,-1) -- (7,-3) -- (7,3) -- (3,1) --  (3,-1);

\node at (-3.8,0)[rotate=90] {PCA Encoder};
\node at (6.7,0) [rotate=90]{XGboost Decoder};
\node at (1.5,0) [blue, thick]{\large\textbf {Noise $\epsilon$}\\\large\textbf{added to}\\\large\textbf {Latent}\\ \large\textbf{Feature}};
\node at (1.5,3) [blue]{\LARGE Step b: $X\to Z\to Z_{\epsilon} \to \hat{X_\epsilon}$};
\end{scope}
\node at (0,-3) [blue]{\LARGE Synthetic Data Output: $X_{\text{syn}}=X+ \hat{X_{\epsilon}}-\hat{X}$};
\end{tikzpicture}
\end{center}

%% file: theo_foundation.tex
\section{Theoretical foundations of the proposed method}
\subsection{Dimension Reduction and Manifold Hypothesis}
The family of dimension reduction algorithms relies on the \underline{\textit{hypothesis}} that many real-life data are generated from or at least can be embedded into low-dimensional latent manifolds. However, without explicit domain knowledge, it is not clear a priori what structure we should endow these manifolds with. The sparsePCA+XGboost approach presents unique advantage by employing a minimalist approach on the encoding part. The nonlinear recovery for the latent representation uses SOTA tabular data prediction method XGboost, which can deal with generation of various data types including: binary, categorical, and continuous data gracefully (See Figure \ref{half-circle-figure} for a graphical illustration and following Section \ref{fund_lim}).
\input{figures/half_circles.tex}

\begin{figure*}
\begin{minipage}[t]{.45\textwidth}
        \centering
        \includegraphics[width=\textwidth]{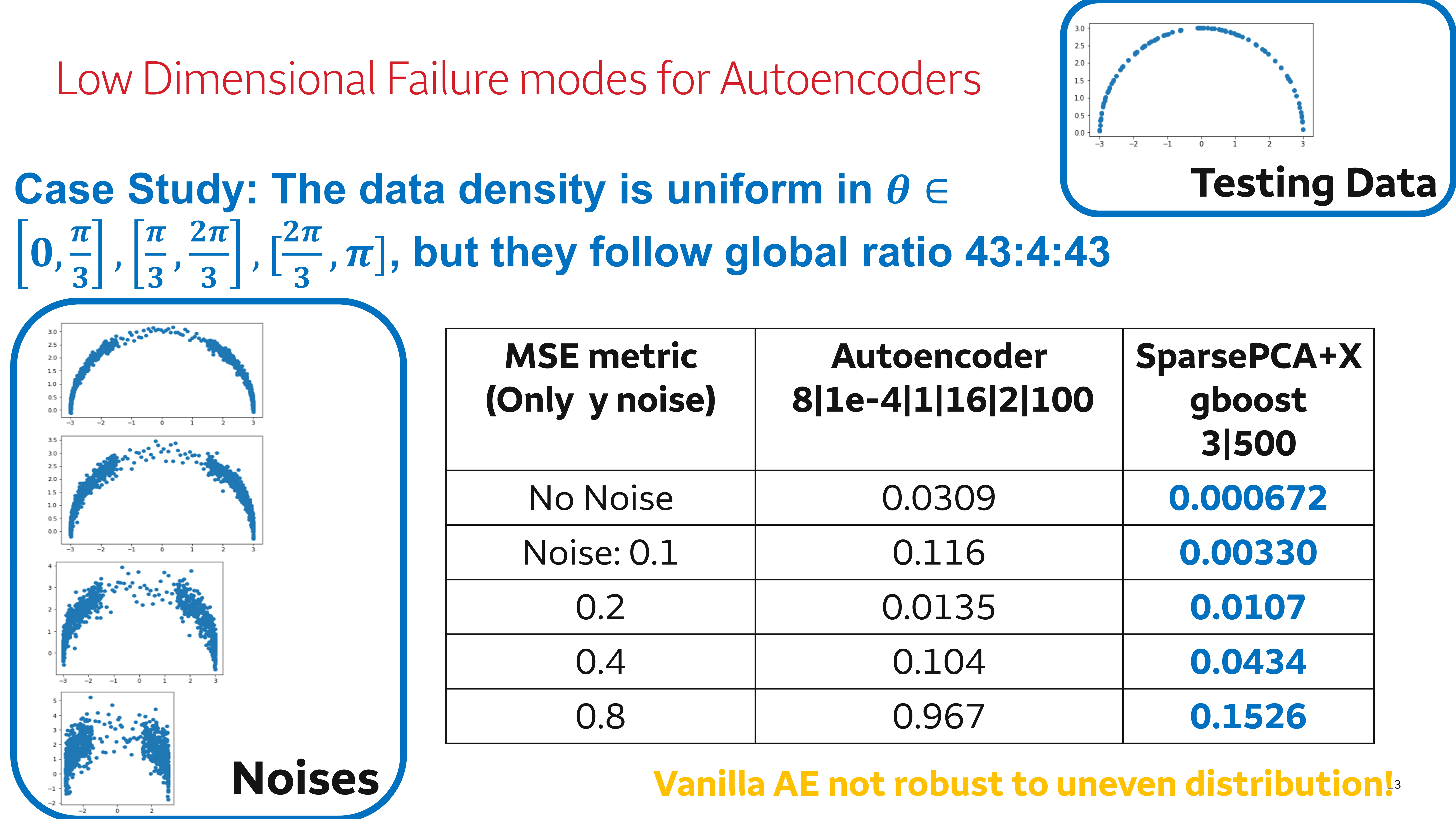}
    \end{minipage}
    \hfill
    \begin{minipage}[t]{.45\textwidth}
        \centering
        \includegraphics[width=\textwidth]{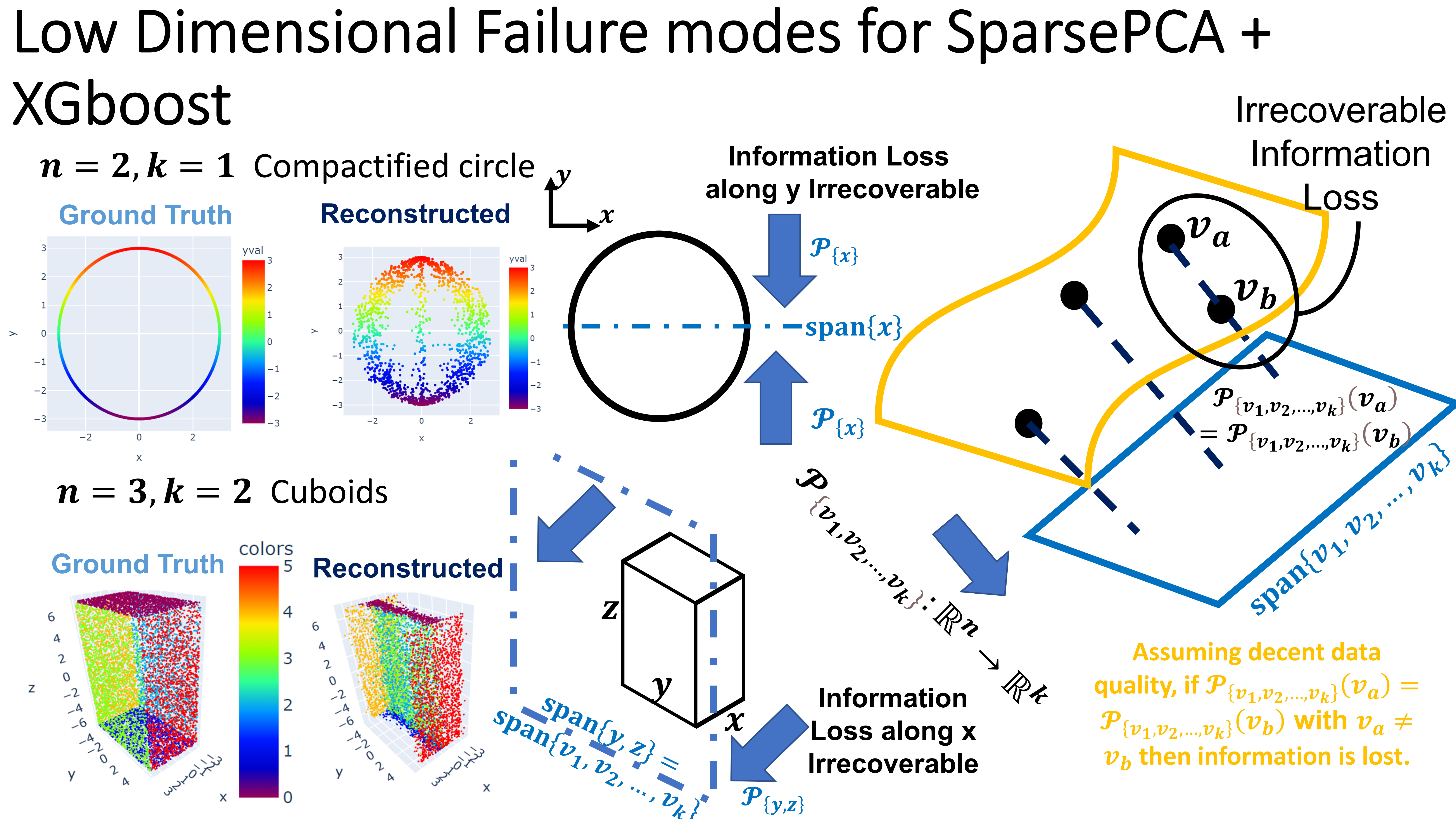}
    \end{minipage}  
\caption{(left). A failure mode for autoencoders: no uniform sampling from 0 to $\pi$. In the example demonstrated, a sparse region was provided in $\theta\in(\frac{\pi}{3},\frac{2\pi}{3})$. This case autoencoders underperforms compared to the proposed method. (right). Characterizing the possible low dimensional failure modes for the pipeline possibly caused by the linear subspace encoder learning from PCA. As illustrated from the circle and cuboids example, the singular dimension that is discarded contains significant information in both examples. In compactified circle example, the $y$ direction contains same information as $x$ and discarding either $x$ or $y$ kills the structure of the data in reconstruction. In the cuboid (surface) example, the smallest $x$ dimension is projected out and this was followed by the misconstruction of four facets. The left and right sides are collapsed onto one plane and lower facets are collapsed into a line.}
\label{fig:failmod}
\end{figure*}

Principal components optimizes for the directions that maximizes the variance which are essentially the singular vectors. This projection presents inherent limitations (See Figure \ref{fig:failmod} right):\\
\subsection{Fundamental limitation in PCA\label{fund_lim}} 
Assuming the principal components dimension being $k$, with the first $k$ singular vectors for the data matrix A given by $\{v_1,v_2,...,v_k \}$. Then the projected linear subspace is $\text{span}\{v_1,v_2,...,v_k \}$. Furthermore, denote $P_{v_1,v_2,...,v_k }:\mathbb{R}^n\to \mathbb{R}^k$ as the projection operator that maps the original Euclidean observed data $v$ to the element $P_{v_1,v_2,...,v_k }(v)$ in the linear subspace. Then, the downstream nonlinear recovery model (XGboost) cannot recover meaningful information represented in $P_{v_1,v_2,...,v_k }  (v_a )=P_{v_1,v_2,...,v_k }  (v_b)$ with $v_a\neq v_b$.
\subsection{Mechanism for gradient boosted trees}
XGboost are inherently interpretable. For example, when it consists of depth 1 trees, it is a generalized additive model (GAM)\footnote{\url{https://selfexplainml.github.io/PiML-Toolbox/_build/html/guides/models/xgb1.html}}
\begin{equation}
    g(\mathbbm{E}(y\mid x))=\mu+\sum_j h_j(x_j)
\end{equation}
When it consists of depth 2 trees, it is a generalized additive model with interaction (GAMI) \footnote{\url{https://selfexplainml.github.io/PiML-Toolbox/_build/html/guides/models/xgb2.html}}
\begin{equation}
    g(\mathbbm{E}(y\mid x)) = \mu+\sum_j h_j(x_j)+\sum_{j<k}f_{jk}(x_j,x_k)
\end{equation}
\subsection{Characterizing the failure modes of Autoencoders and our minimalist model in reconstruction tasks}

We contrast the performance of autoencoders and the performance of our model by providing two explicit scenarios to explicate the bottleneck in these approaches. We note that the main bottleneck for autoencoder is the overfitting issue in handling non-uniform dataset, and the bottleneck in the proposed method is coping with isotropic and symmetric data in the encoding process (See Figure \ref{fig:failmod}). Due to the unsupervised encoder nature where there is no feedback mechanism from decoder as gradient can't propagate, there are limitations with our approach compared to autoencoders for example when dealing with data from \textit{a toy compactified circle}, where sparsePCA is bound to lose information if we need to perform compression. This also coincides with our empirical observations on credit data where for high dimensional data increasing PC dimension properly can lead to huge performance improvements and benefits. 

%% file: figures/half_circles.tex
\begin{figure}[!htbp]
\begin{center}
\vspace{-0.5cm}
\tikzset{cross/.style={cross out, draw=black, minimum size=2*(#1-\pgflinewidth), inner sep=0pt, outer sep=0pt},
cross/.default={3pt}}
\begin{tikzpicture}[baseline=(current bounding box.north)]
\begin{scope}
\clip (-3,-0.1) rectangle (3,4);

\draw [cyan,line width=1mm] plot [smooth] coordinates {(0.5,2.8) (1,2.9) (1.5,2.6)(2,2.1)(2.5,2) (2.7,1)(3,-0.1)};

\draw[dashed, thick, red,line width=0.5mm] (0,0) circle(3);
\draw[thick, black] (-3,0)--(3,0);
\draw [blue,line width=1mm] plot [smooth] coordinates {(-0.3,4) (-0.2,3.5) (0.1,3)(-0.2,2.5)(-0.3,2)};
\draw (1.5,0) node[cross] {};
\filldraw [black] (0.205,2.932) circle (2pt);
\filldraw [black] (-0.197, 3.19) circle (2pt);
\filldraw [black] (-0.568, 2.58) circle (2pt);
\filldraw [black] (-0.206, 2.8) circle (2pt);
\filldraw [black] (0.082, 3.43) circle (2pt);
\filldraw [black] ( -0.241, 3.45 ) circle (2pt);
\filldraw [black] (-0.424, 3.15) circle (2pt);
\filldraw [black] (0.021, 3.56) circle (2pt);
\filldraw [black] (-0.386, 2.54) circle (2pt);

\filldraw [black] (1.5,2.6) circle (2pt);
\draw[orange,line width=0.5mm,-{latex[width=2mm,length=3mm]}] (1.5,2.55)--(1.5,0);

\end{scope}
%
\node[below left= 3mm of {(0,2.6)}, blue] {$p(x\lvert z)$};
\node[below right= 1mm of {(3,0)}, thick, red] {$f(z)$};
\node[ right= 1mm of {(2,2.5)}, thick, cyan] {$\hat{f}(z)$};
\node[label={[orange,label distance=0.5cm,text depth=-1ex,rotate=-90]right:Sparse PCA}] at (1.1,2.9) {};
\node[label={[green,label distance=0.5cm,text depth=-1ex,rotate=-90,font=\bf]right:XGboost}] at (2.1,2.5) {};
\node[draw] at (1.5, 0)   (a) {};
\node[draw] at (1.5, 2.6)   (b) {};
\path[green,dashed,line width=0.5mm,-{latex[width=2mm,length=3mm]}]
    (a) edge [bend right] node {} (b);

\end{tikzpicture}

\end{center}

\caption{Schematic for the half circle example. The red dashed line $f(z)$ resembles the underlying ground truth. The black scatter plots are the observed data. The blue distribution resembles the distributional modeling for $p(x\lvert z)$. The orange line explicates the mechanism for sparsePCA which is projecting onto the $x$-axis in the half circle data, and the green dashed line resembles the XGboost reconstruction. The cyan line $\hat{f}(z)$ means the fitted mapping from latent to original space data.}
\label{half-circle-figure}
\end{figure}

%% file: related_work.tex
\section{\label{contrasts}Other related work}
\subsection{Probability mixture, transformation, disentangling, modeling, and blind source separation}
When we need to generate samples that conform to the data distribution, one possible way is to sample from a tractable distribution (say normal) and then apply transformations to obtain target distribution.  In the pure probabilistic picture where we obfuscate individual characteristic (Step 4 in Algorithm \ref{proposed-alg}), we would need to transform the base uniform distribution to the latent distribution to resemble the latent structure. The problem of mimicking the latent distribution can be framed as a density estimation problem. The na\"ive approach for this is to parametrically estimate the distributional parameters in the latent space, and then sample from the fitted parametric distribution. However, in high dimensional latent space, correlational and entanglement often cannot be effectively described by parametric assumptions. Therefore, we need a tractable approach to transform a tractable distribution (say random multivariate normal) to the target latent distribution in the training dataset latent.

Our implemented approach (See Figure \ref{fig:cred_gen1} for empirical result) for desensitized data generation fits an empirical distribution on each latent variables. \textit{The downside of this approach is although the PCA components are uncorrelated, they are not necessary independent, which means marginalization limits expression of uncorrelated but dependent information.} In other words, $\text{Cov}(X,Y)= 0$, $\not\Rightarrow$$\qquad I(X,Y)= 0$, where $I$ is the mutual information. We observe this na\"ive marginalization approach works well empirically, but it may be improved with several alternatives.
\subsubsection{Linear disentangling}
A partial remedy for the non-independence in PCA latents is adding ICA on top of PCA or replace ICA for preprocessing (See \ref{ICAs}). In the \textit{linear mixing} case (i.e., data generated from $Y=AX$ where $X$ contains independent sources on the rows), if the mixture's \ul{fourth order kurtosis spectrum is distinct and strictly ordered}, then the ICA process can effectively generate samples by:
\begin{enumerate}
    \item Sample from the disentangled empirical marginal distribution $\hat{x}$
    \item Multiply the fitted linear mixing matrix $W_{\text{ica}}$ with $\hat{x}$ and obtain the synthetic latent space codes $W_{\text{ica}}\hat{x}$
\end{enumerate}
\subsubsection{The nonlinear case: modeling complex distribution}
\begin{figure}[!htb]
\includegraphics[width=8.5cm,height=5cm]{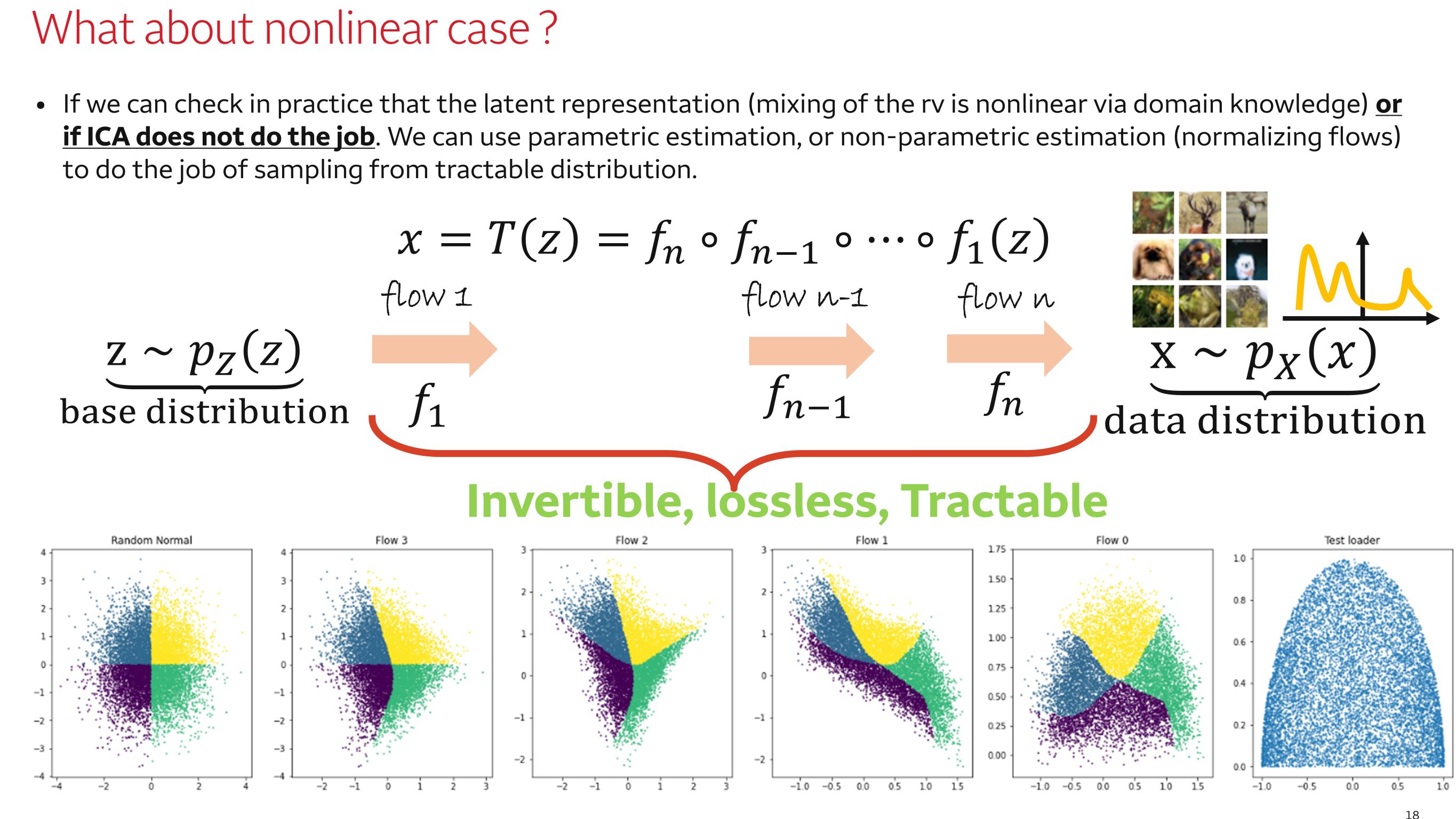}
\caption{ The flow-based landscape of latent representation learning for synthetic data generation.}
\label{fig:normal_flow}
\end{figure}

In the nonlinear case, we have to resort to density estimation schemes where functional approximation of the density is achieved by neural networks. We mainly revisit two types of \underline{\textit{exact}} distribution modeling method: \textit{autoregressive models} and \textit{flow-based methods}\cite{tomczak2021deep}.\\
\textbf{Flow-based method}: The backbone of flow based method relies on the change of variable formula
\begin{equation}
    p_X(\pmb{x}) = p_Z(f^{-1}(\pmb{x}))|\text{det}(\frac{\partial f^{-1}}{\partial \pmb{x}})|
\end{equation}
$\pmb{X} = f(\pmb{Z})$, where $f$ is an differentiable and invertible function with tractable Jacobian, and $ \pmb{Z}\sim p_Z(z) $ samples from a base distribution. 
The canonical base distribution for normalizing flows method is multivariate normal distribution \cite{rezende2015variational}. The power of such models derives from concatenating such flows (i.e., bijective, invertible, and lossless maps  (homeomorphism) ) guarantees interpretability and tractability. The mapping (See Figure \ref{fig:normal_flow}) from a base distribution (simple) to a target distribution (complex) is realized through neural network approximations. The bijection and invertibility implies tractability so that one can model the complex distribution through sampling from simple distribution and go through the coupling layers to entangle the relations of the variable to reach target density functions. Copula flows\cite{kamthe2021copula} captializes on this idea and proposed to use copulas learned from normalizing flows to capture multivariate variable dependence in tabular data.
\\
\noindent\textbf{Autoregressive models}. Autoregressive models have been proven successful in GPT decoder-only language modeling \cite{brown2020language}, especially effective under scaling laws \cite{kaplan2020scaling}. This group of methods perform density estimation iteratively via
\begin{equation}
    p(\pmb{x})=p(x_1)\prod_{i=2}^n  p(x_i\lvert \pmb{x}_{\leq i-1})
\end{equation}
However, this approach is susceptible to estimation ordering of features \footnote{Learning the density $p(x_1,x_2) = \mathcal{N}(x_2\mid 0,4)\mathcal{N}(x_1\mid \frac{1}{4}x_2^2,1)$ in $(x_1,x_2)$ order - namely $p(x_1,x_2)=p(x_1)p(x_2\mid x_1)$ would be problematic \cite{papamakarios2017masked}}, hence potentially limiting its use for tabular data generation. Furthermore, when there isn't sufficient data this approach would not be feasible as conditioning incurs many branching. To ease the branching and data-scarcity problem, an LSTM and RNN\cite{hochreiter1997long} approach can be employed to summarize the previous (hidden) states
\begin{equation}
    p(\pmb{x})=p(x_1)p(x_2\lvert x_1
)\prod_{i=3}^n  p(x_i\lvert f_{RNN}(x_{i-1},h_{i-2}))
\end{equation}
however these methods are subtle to train and subject to systematic forgetting or gradient explosion when handling long range interactions. A convolutional approach may also be employed to model long range interaction, but still an explicitly variable ordering is needed\cite{kalchbrenner2014convolutional}. 
\subsection{Other Methodologies}
\noindent \textbf{Data Imputation approach:} Previous work have approached the problem of synthetic data generation through a data imputation approach \cite{marino2019hdda}. This is similar to employing a masked autoencoder approach (as in training encoder-only models) \cite{germain2015made}. However, a pitfall of this approach is that in financial tabular data one variable could be independent from the others. In this case, when missing at random is introduced, information about the independent variable could be irreversibly lost.
\\
\noindent\textbf{VAEs: }
VAEs are essentially generative autoencoders that can reconstruct and sample complex data distributions. They are deterministic autoencoders with noises injected in the latent layer \cite{ghosh2019variational}, where they could lead to robust representations \cite{achille2018emergence}. Unlike homogeneous image data, tabular data often contains a mixture of categorical, binary, and continuous data. Tuning these features in VAEs can be clumsy in practice. The proposed method is flexible in handling these variable type variations compared to VAEs\footnote{ A related concurrent work \cite{srivastava2020improving} implemented a Multistage VAE demonstrating that VAEs can uncover distinct and meaningful
features of variations on tabular data.}.

%% file: evaluation_metrics.tex
\section{Evaluation Metrics}
The evaluation for the quality of tabular
synthetic data can be classified as \underline{\textbf{qualitative}} or \underline{\textbf{quantitative}}. 
\subsection{Qualitative Comparison}
On the \textit{\underline{qualitative}} side, a first step is to visually inspect the \textit{marginal distribution} and compare that with the \textit{empirical distribution}. This approach bears the same moral with comparison of image generation quality from visual cues. The intuition derived from the evaluation is highly contingent upon the data visualization strategy and data preprocessing and projection methods employed. The main challenge remains as tabular data are often high dimensional, hence it is not possible to succinctly visualize all its underlying structure for comparisons. Therefore, one often plot and compare projections of these generated data where one slice a subset of the dataset.
 
\subsection{Quantitative Comparison}
The \underline{\textit{quantitative}} aspect relies on metrics that produces numerical outputs that measures the quality according to  specific assumptions. We mainly reported the PSI result


\noindent\textbf{PSI distance} The Kullback–Leibler \textit{divergence} measures how far one distribution is to another but it is not a \textit{\underline{distance}} metric because it is not symmetric (i.e., $KL(P\|Q)\neq KL(Q\|P)$). The population stability index (PSI) can be leveraged to solve this problem by symmetrizing KL into a distance measure $\text{PSI}\equiv \text{PSI}(P,Q)=\text{PSI}(Q,P)$ with
\begin{equation}
    \text{PSI} = KL(P\|Q)+ KL(Q\|P) =\sum_i (P_i-Q_i)\log\frac{P_i}{Q_i}
\end{equation}
where the last equality assumes categorical nature (continuous counterpart derivation is similar). As a distance metric, the lower the quantitative value the better the model performance.

%% file: experiments.tex
\section{Experiments}
We first study several low dimensional toy experiments and present a high dimensional credit data example.
\subsection{Low dimensional Toy experiments}
We focus our study on several 2D and 3D scenarios to get an intuition on the capabilities and intuitions of the approach.
\subsubsection{The half circle in 2D\label{half-c-exp}}
\begin{figure*}[!htbp]
\begin{minipage}[t]{.45\textwidth}
        \centering
        \includegraphics[width=\textwidth]{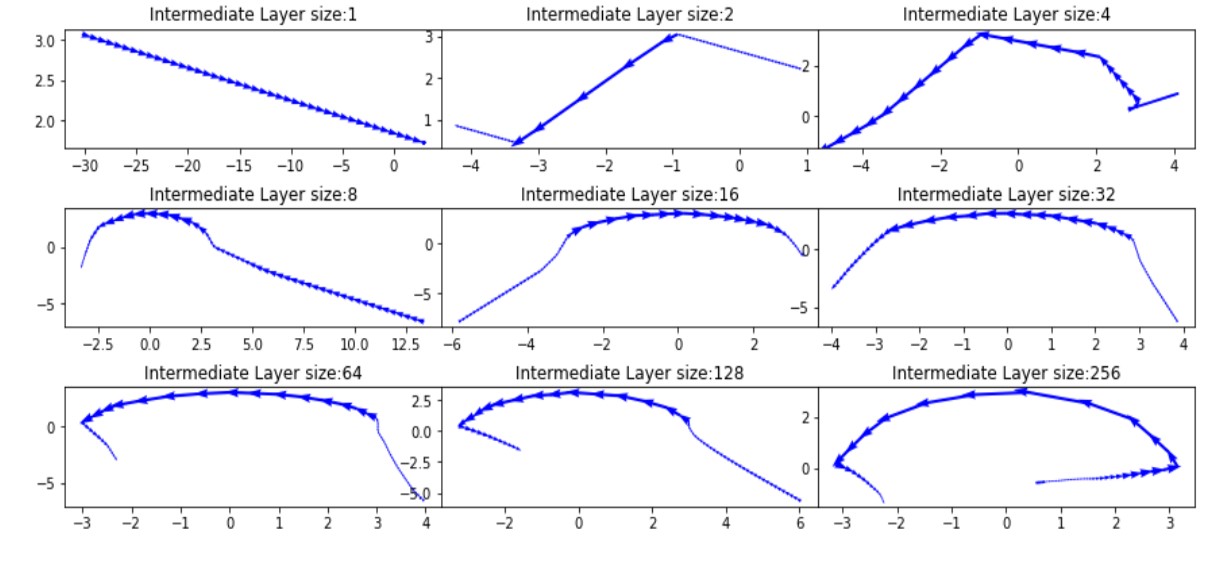}
    \end{minipage}
    \hfill
    \begin{minipage}[t]{.45\textwidth}
        \centering
        \includegraphics[width=\textwidth]{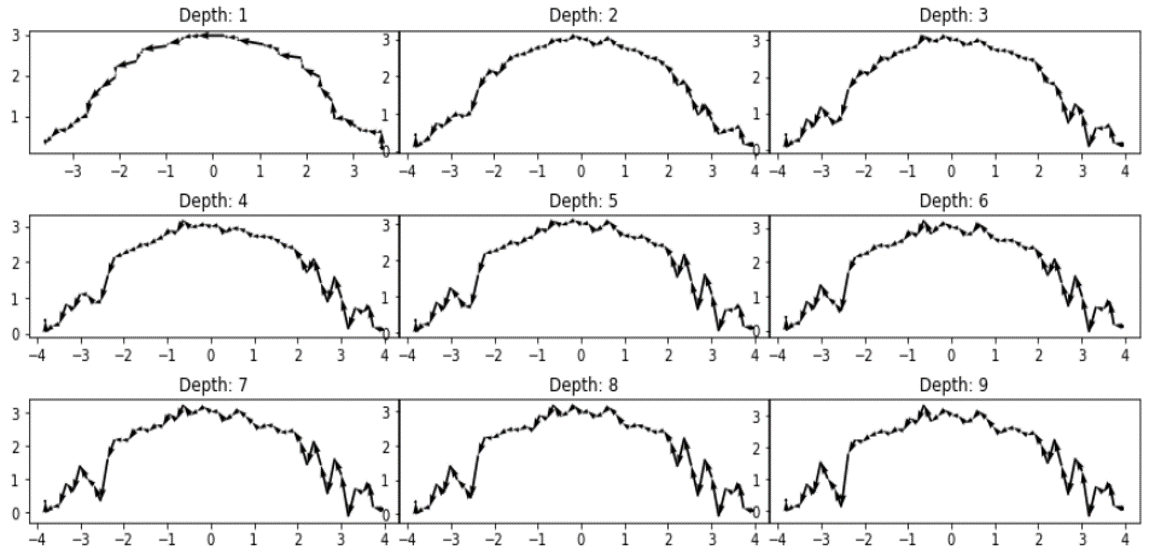}
    \end{minipage}
    \caption{Latent traversals for autoencoder (left) and XGboost (right) fixed range latent traversals. The autoencoder consists of various intermediate sizes $j\in\{1,2,4,8,16,32,64,128,256\}$, the pipeline is a 7 layer neural network $2\mapsto j\mapsto j\mapsto 1\mapsto j\mapsto j\mapsto 2$. The XGboost is constructed with 500 boosting rounds with $\text{depth}\in \{1,2,3,4,5,6,7,8,9\}$}
    \label{XGboost_nn_comp}
\end{figure*}

An illustration for the half-circle simulated setup is previously shown schematically (Figure \ref{half-circle-figure}). We can perform \textit{latent traversals} on the compressed one latent dimension as 1D real space is explicitly ordered. We present the latent traversal comparison between autoencoders and our approach (See Figure \ref{XGboost_nn_comp}). We observe that rule-based XGboost handles out of range latent variables better.
\begin{figure}[!htbp]
    \centering
        \includegraphics[width=5cm,height=3cm]{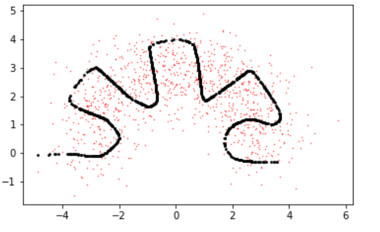}
    \caption{The red is the ground truth and the black scatters are the fitted reconstructions. We observe that the neural-network based autoencoder forms a dense fitting layer to “approximate” in the noise direction}
    \label{fig:nnoverfits}
\end{figure}

\begin{figure*}
\begin{minipage}[t]{.45\textwidth}
        \centering
        \includegraphics[width=\textwidth]{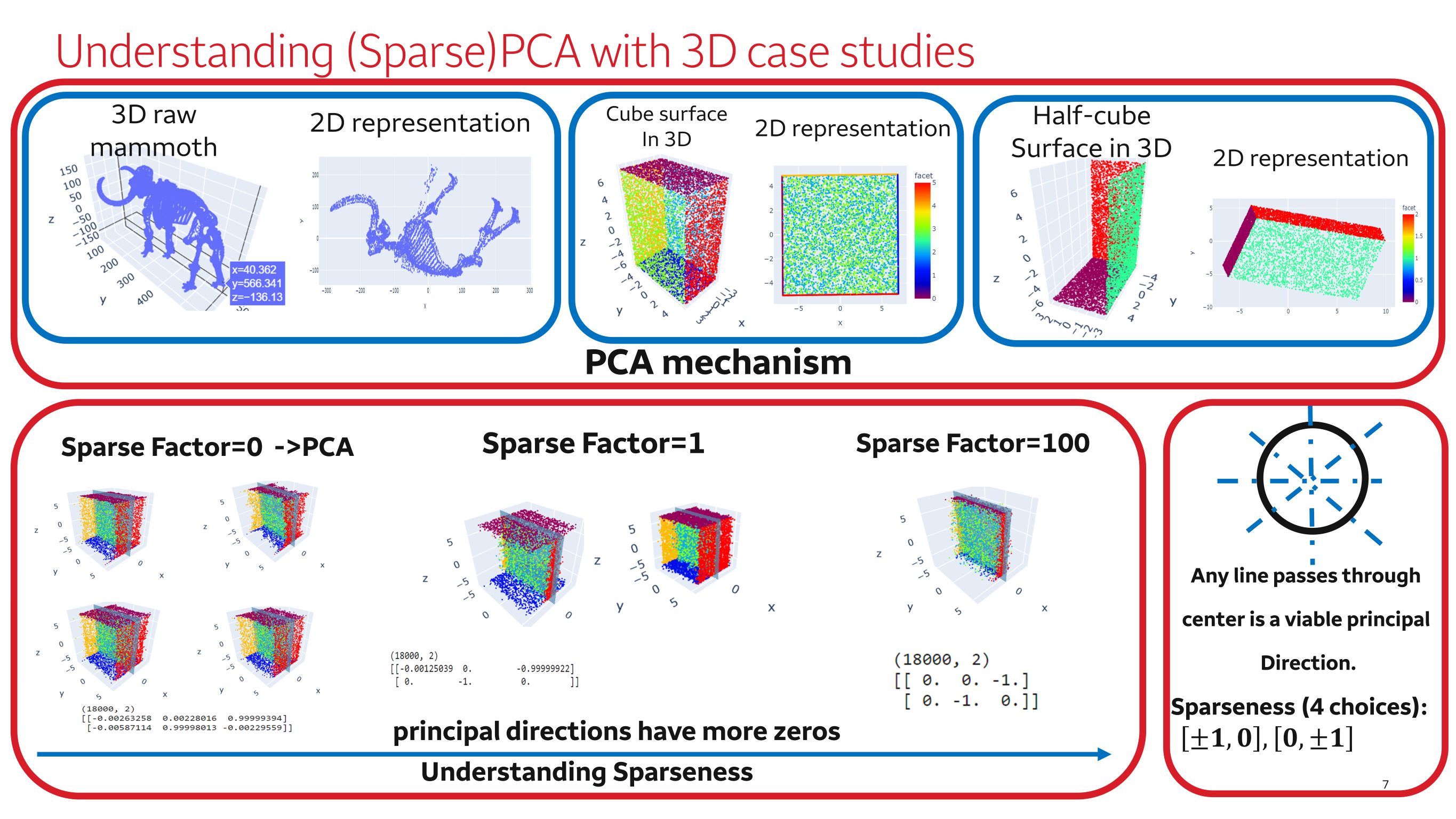}
    \end{minipage}
    \hfill
    \begin{minipage}[t]{.45\textwidth}
        \centering
        \includegraphics[width=\textwidth]{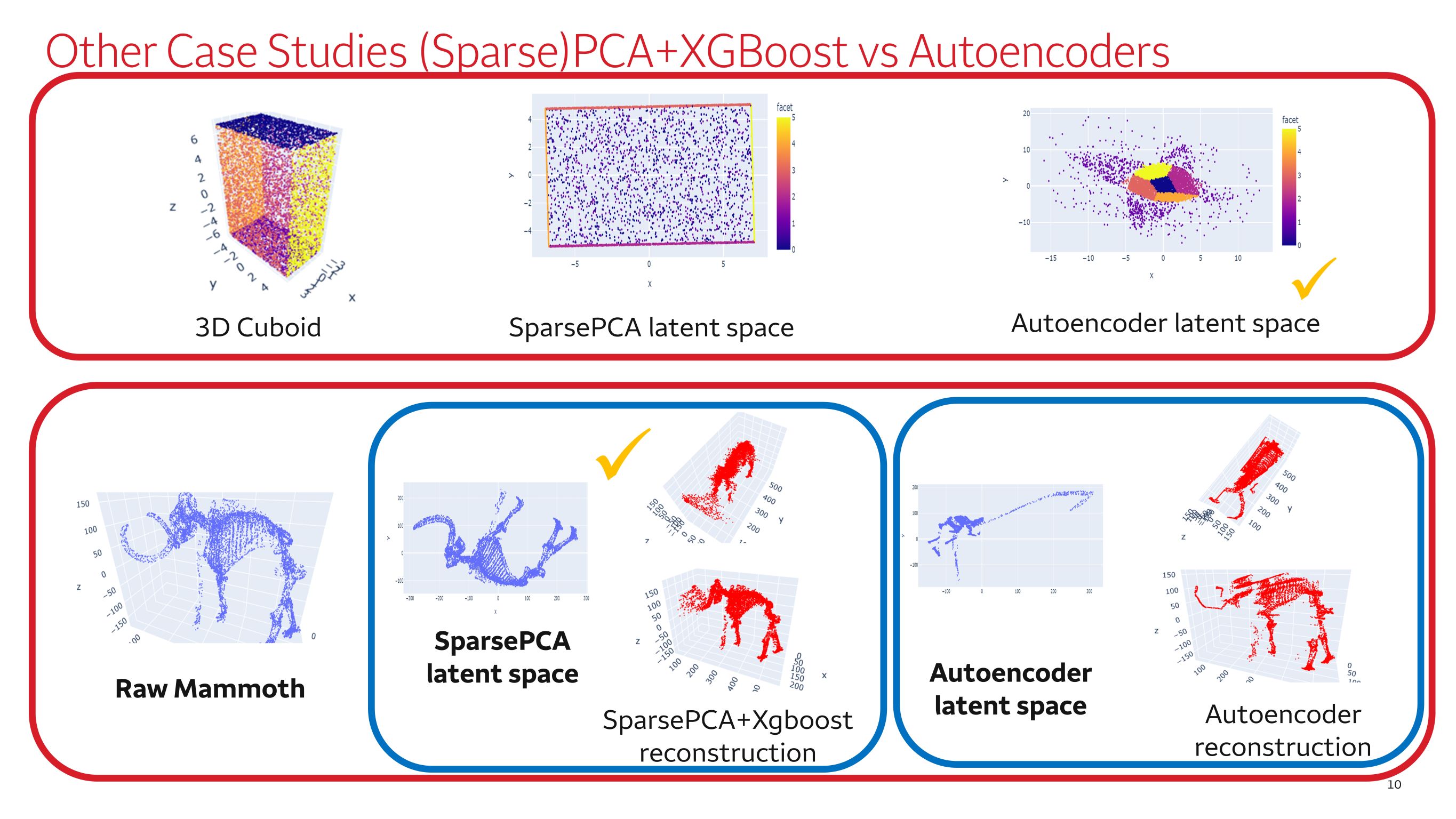}
        
    \end{minipage}
    \caption{(Left) Case studies to illustrate how does PCA projection work in practice. The key limitation for PCA is its symmetries. The bottom left illustrates increasing sparseness factor can be detrimental to the cube example for reconstruction. (Right) Reconstruction Case studies for 3D cuboid and mammoth. In the cuboid scenario our approach is limited because the upper and lower surface collapse together in latent space and is not separable. In the mammoth case, autoencoder preserved the bone structure but our approach is more solid and the overall shape is smoother and more realistic.}
    \label{fig:autoencoder_ma_cu}
\end{figure*}
\begin{figure*}[!htbp]
\begin{minipage}[t]{.5\textwidth}
        \centering
        \includegraphics[width=\textwidth]{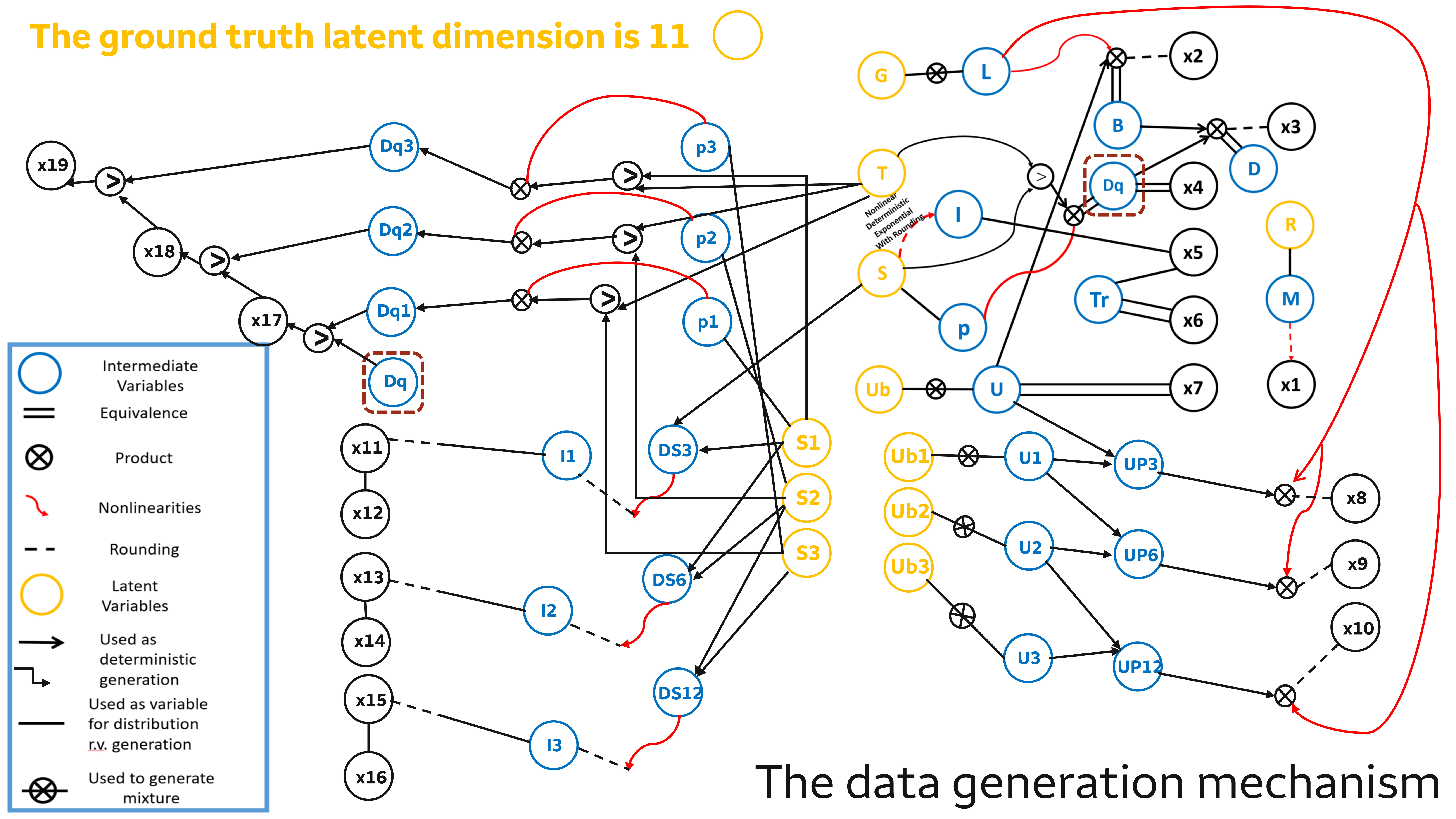}
    \end{minipage}
    \hfill
    \begin{minipage}[t]{.45\textwidth}
        \centering
        \includegraphics[width=\textwidth]{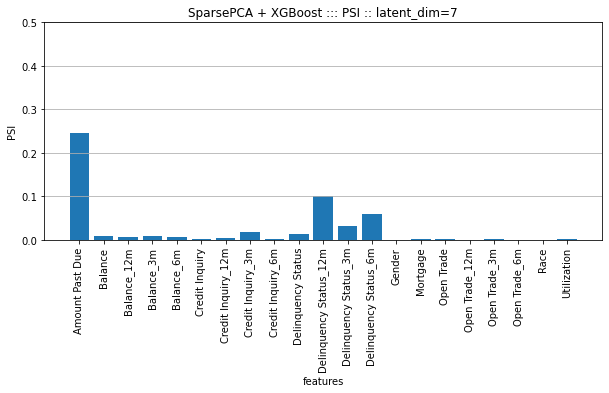}
    \end{minipage}
    \caption{(Left) The data generation mechanism for the simulated credit data. (Right) Quantitative Comparison for the 21 variables for PSI metric in Algorithm \ref{proposed-alg} (Step 4)}
    \label{fig:cred_gen1}
\end{figure*}

We also test explicitly on the performance for autoencoders under noise exposure. We set up a ground truth with a half circle with radius $3$
and $0.2$ standard deviation gaussian noise added on both $x$ and $y$ direction. 
We note that autoencoder can not model residual (error) information (nor could VAEs in this case) in one latent dimension. In this case, the autoencoder creates a sheet of layer mixing the circle law with the residual information if we minimize the reconstruction loss (See Figure \ref{fig:nnoverfits}). In this case the linear sparsePCA projection is expected to work better, as the sparsePCA encoder acts as an effective regularizer for the noise information.

\subsubsection{The 3D mammoth and the Cube}
We graphically illustrate our approach using two 3D point cloud objects (Cube and Mammoth). As expected, the 3D point clouds are linearly projected on a plane spanned by two largest variance direction and then XGboost recovers the nonlinearity. The qualitative result is reported in Figure \ref{fig:autoencoder_ma_cu}.

\subsection{Higher dimensional simulated credit data}

We simulated a credit dataset with a combination of linear and nonlinear groundtruth dynamics across a mixture of binary, categorical, and continuous variables. A schematic of the data generating mechanism for the credit data is given (See Fig \ref{fig:cred_gen1} (left)). The canonical synthetic generation (Step 4 in Algorithm \ref{proposed-alg}) approach with calculated PSI metric for the variables is given in Figure \ref{fig:cred_gen1} (right). In the result, some of the skewed variables including amount past due and delinquency status have higher PSI distance compared to other variables. 

Generating synthetic data from sampling a base distribution (Step 4 in Algorithm \ref{proposed-alg}) is limited in its applicability for robustness testing in a controllable setting. Therefore, we conduct a case study on simulated credit data in a controllable setting in the following section emphasizing on the robustness testing application focusing on Step \textcolor{blue}{$4^*$}.


%% file: case_study.tex
\section{Case Study with Application to Model Robustness Analysis}
 \input{robustness_v1.tex}
\input{baselines.tex}
\input{mortgage.tex}
\input{credit_i.tex}
\input{apd.tex}
\begin{figure*}[!htbp]
    \centering
    \includegraphics[width=\textwidth,height=10cm]{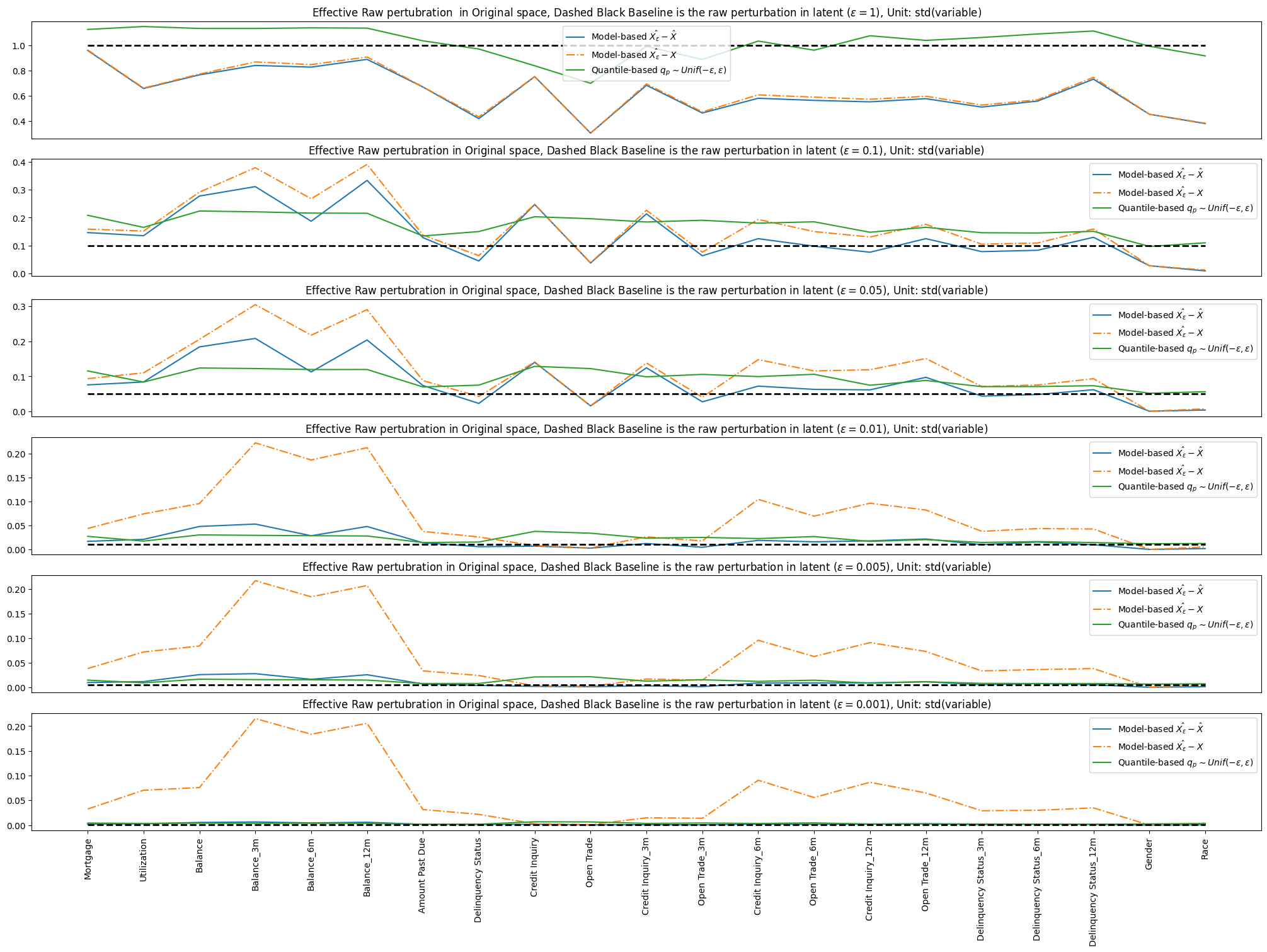}
    \caption{The y axis data are calculated from the perturbation sizes divided by the standard deviation of each feature. We observe that the model based approach offers an alternative to the raw perturbation and quantile perturbation which adds noise in a model based manner. Specifically, we note that when $\epsilon=1$ the model-based is small suggesting model information is retained. The blue lines is non-constant across variables compared to the dashed black lines (baseline $\epsilon$ is the perturbation size in latent space). Compared to quantile perturbation (green curves) the model-based method has more variability in most $\epsilon=0.1,0.01$ scenarios. The latent space size is 10, boosting round is 200 and depth of tree is 2.}
    \label{fig:effective_perturb}
\end{figure*}
\input{correlation_matrix.tex}
\input{r_q_model_based.tex}
21 predictors are selected (See Figure  \ref{fig:cred_gen1} (left)), these include $x_1,...,x_{19}, G,R$, where $R$, $G$ are race and gender variables which also contribute to generating other relevant variables. We first log transform the skewed data variables so that PCA can operate linearly. The underlying designed latent dimension is 11 (but the sources of randomness is higher than 11). Recall instead of sampling from the fitted uniform, we straightforwardly perturb the latents and obtain synthetic data outputs $\hat{X}_{\epsilon}$ and $X+\hat{X}_{\epsilon}-\hat{X}$(See Figure \ref{robust_alg}). 

There are several kinds of perturbation generating synthetic data for robustness testing: raw perturbation, quantile perturbation and model-based perturbation. For the raw perturbation, the  perturbation size does not depend on the density of the region. For the quantile perturbation, the perturbation size will depend on the density of the region. The model-based perturbation is done in the latent space while for raw and quantile perturbation, these are performed in the original space.

To summarize, within each variable, in the quantile scale, the quantile perturbation size is constant while the raw perturbation is not. In the raw scale, the perturbation is constant for raw but not for quantile. \textit{We show that the model-based approach is a reasonable alternative of these two (See Figure  \ref{fig:effective_perturb}}).

Across different variables, raw perturbation and quantile perturbation are based on strong assumptions - for raw perturbation the perturbation size is the same proportion of the standard deviation of a feature. For quantile perturbation the perturbation result have the same quantile size. The proposed method is a relaxation for both of these assumptions.

We focus on the perturbation size difference in different regions of the variables. We visualize the different regions of predictors for model based methods. Then we present some comparison of the perturbation size across different variables, we conclude model-based approach provides more flexibility. Specifically, we look at

\begin{itemize}
    \item Perturbation size within each variable
    \item Perturbation size across different variables
\end{itemize}
We present visualizations in terms of what these perturbation would look like in the model-based protocol (PCA+XGboost).
\subsection{Raw Visualization after Model-based perturbations}
\subsubsection{Visualizing the Model-based raw $\hat{X_{\epsilon}}$, $X-\hat{X}+\hat{X_{\epsilon}}$, and $X-\hat{X}+\hat{X_{\epsilon}}$(truncated) across variables}
We add Gaussian noise on each latent $i$ with standard deviation \tcbox[on line, frame empty,boxsep=0pt,left=1pt,right=1pt,top=1pt,bottom=1pt,colback=gray!40]{$\epsilon*$std(latent[:,$i$])} and obtain the corresponding synthetic data $X+\hat{X}_{\epsilon}-\hat{X}$ for robustness testing. The XGboost generator has a depth 2 with 200 boosting rounds to limit three-way variable interaction and overfitting. For continuous variables the bin size is 40. The empirical distribution are plotted (See Figure \ref{robustness_vis}).

\subsection{Within variable visualization}

As we discussed earlier, raw perturbation and quantile perturbation have very different behavior in different regions within a variable.  We would like to see how the model based perturbation will look like within a variable. 

Since PCA is effectively a linear transform, we applied a log transform preprocessing to the skewed variables and the result is evaluated in the preprocessed domain. \subsubsection{Baselines of the corresponding variables }
We first visualize the corresponding ground truth value landscape (Figure \ref{fig:baselines}). 

We focus on the result of the perturbation for  \underline{Mortgage} $x_1$(Figure \ref{fig:region_mortgage}), \underline{Credit Inquiry 6 month} $x_{13}$(Figure \ref{fig:region_credit}), \underline{Amount Past Due} $x_3$(Figure \ref{fig:region_apd}). We bin the result into 3 bins: The values lower than 10\% percentile, values higher than 90\% percentile and the middle 10\%. The Mortgage variable is picked from the lower 10\% percentile ($x\leq q_{0.1}$), values higher than 90\% percentile ($x\geq q_{0.9}$) and the middle 10\% ($q_{0.45}\leq x\leq q_{0.55}$). The credit variable is a bit skewed so the middle 10\% quantiles are picked from ($q_{0.5}\leq x\leq q_{0.6}$) to contrast with the lower $x\leq q_{0.1}$ cohort and upper $x\geq q_{0.9}$ cohort. The amount past due variable is also skewed. To minimize the overlap, the middle 10\% is picked from ($q_{0.58}\leq x\leq q_{0.68}$) to contrast with the lower $x\leq q_{0.1}$ cohort. This is due to the sheer amount of data at $x=0$, thus we tweak the middle group for \textit{\textbf{Credit Inquiry 6 months}} and \textit{\textbf{Amount Past Due}}.

For the perturbation of the \underline{Mortgage}(Figure \ref{fig:region_mortgage}), the perturbation size remains consistent across different regions of the variable, similar to the raw perturbation. The mortgage variable has undergone a log transformation and roughly follows a Gaussian distribution. This consistency is partly due to the perturbation method applied in the latent space. By adding Gaussian noise in the latent space, we are effectively using raw perturbation. The model-based perturbation did not significantly alter the perturbation pattern across different regions. 

For the perturbation of the \underline{Credit Inquiry 6 month} (Figure \ref{fig:region_credit}), the conclusion remains consistent: the perturbation is more similar to raw perturbation than to quantile perturbation. However, we observed that the Credit Inquiry 6-month variable becomes negative for some observations after perturbation, which is illogical. A post hoc truncation may be necessary in this situation. This is also a limitation for the raw perturbation, which will also introduce illogical perturbations.  Quantile perturbation will not have this issue.     

For the perturbation of the \underline{Amount Past Due}(Figure \ref{fig:region_apd}), the conclusion is different.   This variable is highly skewed, with a high density near zero. In the lower 10\% region, the perturbation size is much smaller, even though the perturbation size in the latent space remains the same. The model significantly impacts the perturbation size in different regions in this highly skewed variable situation. 

These analyses suggest that the model-based perturbation method potentially offers a good compromise between raw perturbation and quantile perturbation. However, it may still produce illogical values after perturbation, necessitating post hoc adjustments. Additionally, the assumption of raw perturbation in the latent space requires further testing and validation.

\subsection{Noise Landscape and perturbation result}
We observe that the noise size $\hat{X}_{\epsilon}-\hat{X}$ converges to a delta distribution more consistently compared to $\hat{X}_{\epsilon}-X$, which is what is expected, suggesting the $\hat{X}_{\epsilon}-\hat{X}$ perturbation may be more applicable for practice. In practical setting, we propose using $\epsilon\sim 10^{-2}-10^{-1}$ (See Figure \ref{fig:effective_perturb}).
\subsection{Comparison between Raw and Model-based
and Quantile perturbation}
Notice the two protocol $\hat{X_{\epsilon}}$ and $X+\hat{X_{\epsilon}}-\hat{X}$ has effective perturbation $\hat{X_{\epsilon}}-X$ and $\hat{X_{\epsilon}}-\hat{X}$ on the base dataset $X$ respectively. We compare three protocols (See Figure \ref{fig:effective_perturb}):
\begin{itemize}
    \item Adding $\hat{X_{\epsilon}}-X$ perturbation
    \item Adding $\hat{X_{\epsilon}}-\hat{X}$ perturbation
    \item Adding quantile $q_{p}\sim Unif(-\epsilon,\epsilon)$ in the quantile space
\end{itemize}
Let's closely examine the cross-variable perturbation in Figure \ref{fig:effective_perturb}.  For perturbation, $\hat{X_{\epsilon}}-X$, there is some impact from the reconstruction error. Even when the perturbation size is very close to 0, $\hat{X_{\epsilon}}-X$ can be significantly different from 0.  However, as the perturbation size increases, the relative difference between $\hat{X_{\epsilon}}-X$ and $\hat{X_{\epsilon}}-\hat{X}$ becomes smaller. At a perturbation size of 0.1, the relative difference between the two perturbations is no longer significant for most variables. Observe when perturbation size $\epsilon =1$, the model-based method consistently has a smaller perturbation size, suggesting that the model-based method retains model information when noise is large.

We are mostly interested in the perturbation size in the range of $\epsilon\sim 10^{-2}-10^{-1}$.  Overall, the perturbations across different variables from the model-based methods and quantile methods are generally similar, with exceptions. 


One exception is the Balance variables. The abnormality may be caused by the correlations and insufficient latent variables\footnote{Empirically, we observe that lowering the boosting rounds can shrink down the perturbation sizes $\hat{X}_{\epsilon}-\hat{X}$ significantly. 
}. The bottom rows of Figure \ref{fig:effective_perturb} also suggests that the balance variables have high perturbation size when the $\epsilon$ is small suggesting underfitting.

Another exception is the credit Inquiry variables. The credit variables perturbation sizes are slightly higher than quantile perturbation but not significant. This phenomena is also observed in Figure \ref{robustness_vis} for $\hat{X}_{0.1}$. The observation could be attributed to the XGboost classification tree for these variables are less robust. 

For the amount past due variable, the model-based noise coincide with quantile based method for $\epsilon=0.05,0.1$ also validating our previous claim that it is closer to quantile perturbation.

The cross-variable perturbation results for the model-based methods appear reasonable overall. Further investigation and post hoc adjustments might be needed for variables with unusual perturbation sizes.

\subsection{Robustness Testing Comparison}
We can use the generated perturbations to test robustness of model predictions to slightly perturbed dataset. To do this we can obtain the model performance on the perturbed datasets. It is expected that as we increase the perturbations the model performance will deteriorate. We built 3 models, ‘xgb1’ with depth 1 and 100 estimators, ‘xgb2’ with depth 2 and 100 estimators and ‘xgb5’ with depth 5 and 300 estimators with all other hyper-parameters at default settings. The performance can be observed in Table \ref{tab:rob_table}.
\begin{table}[]
    \centering
    \begin{tabular}{|c|c|c|c|c|}
    \hline
         &train\_auc&test\_auc &train\_ll& test\_ll  \\
         \hline
         xgb1&0.781&0.767&0.563&0.574\\
         xgb2&0.805&0.772&0.534&0.566\\
         xbg5&0.979&0.744&0.277&0.623\\
         \hline
    \end{tabular}
    \caption{Various benchmark XGboost models for robustness testing.}
    \label{tab:rob_table}
\end{table}
We can see that ‘xgb2’ has the best performance closely followed by ‘xgb1’, and ‘xgb5’ has clearly overfit based on the AUC performance in test data and the train test gap. The ``ll" stands for the logloss, which is the negative loglikelihood. The logloss statistics also supports that shallower trees achieves a better performance. We now conduct the robustness test for these models using 3 types of perturbation strategy, the ‘spca’ strategy described in this paper and two other strategies ‘raw’ perturbations and ‘q’ perturbations. The ‘raw’ perturbations generate noise from a Gaussian distribution with same correlation as the original data. The perturbations are scaled to the standard deviation of the individual variables. The ‘q’ perturbations transform the data into quantile space, adds noise in the quantile space and reverts the perturbed data into original data space. For the ‘raw’ and ‘q’ and  ‘spca’ perturbations we generate 10 replicates of perturbed test data, compute the AUC of each perturbed replicate, and get the average of the AUC across the replications as the final metric.  We plot the metrics (Figure \ref{fig:robust}) for increasing noise in the perturbations specified by the ‘budget’ argument which controls the amount of perturbation in each method. Note that same budget may produce differing amount of perturbation for different strategies.

The results from different perturbation strategies show slight variations. The robustness test results for `xgb1' and `xgb2' do not differ significantly under raw perturbation. However, the quantile perturbation and model-based perturbation methods can better distinguish between the two models.  The `xgb2' model performs worse than `xgb1' as the perturbation sizes increase, indicating that `xgb1' is more robust in such scenarios. Given that the performance of `xgb1' is very close to that of `xgb2', it may be preferable to choose the `xgb1' model in this case.  The model-based perturbation is most sensitive in the area of small perturbation sizes, which is of primary interest. The impact of perturbation on `xgb5' is similar; however, `xgb5' is clearly overfitting and consistently performs worse at each perturbation size.

Overall, the tests show consistent results across the strategies, and we can see that the data generated using the method described in this paper is suitable for conducting downstream robustness analysis.

%% file: robustness_v1.tex
\begin{figure*}[!htbp] \centering
    \makebox[0.01\textwidth]{}
    \makebox[0.3\textwidth]{\Large $\hat{X_{\epsilon}}$}
    \makebox[0.3\textwidth]{\Large $X+\hat{X_{\epsilon}}-\hat{X}$}
    \makebox[0.3\textwidth]{\Large \quad$X+\hat{X_{\epsilon}}-\hat{X}$(truncated)}
    \\
    \vspace{0.3cm}
    \raisebox{0.1\height}{\makebox[0.01\textwidth]{\rotatebox{90}{\makecell[c]{\small{ \quad\quad\quad$\epsilon=0.1$} }}}}
    \includegraphics[width=0.3\textwidth]{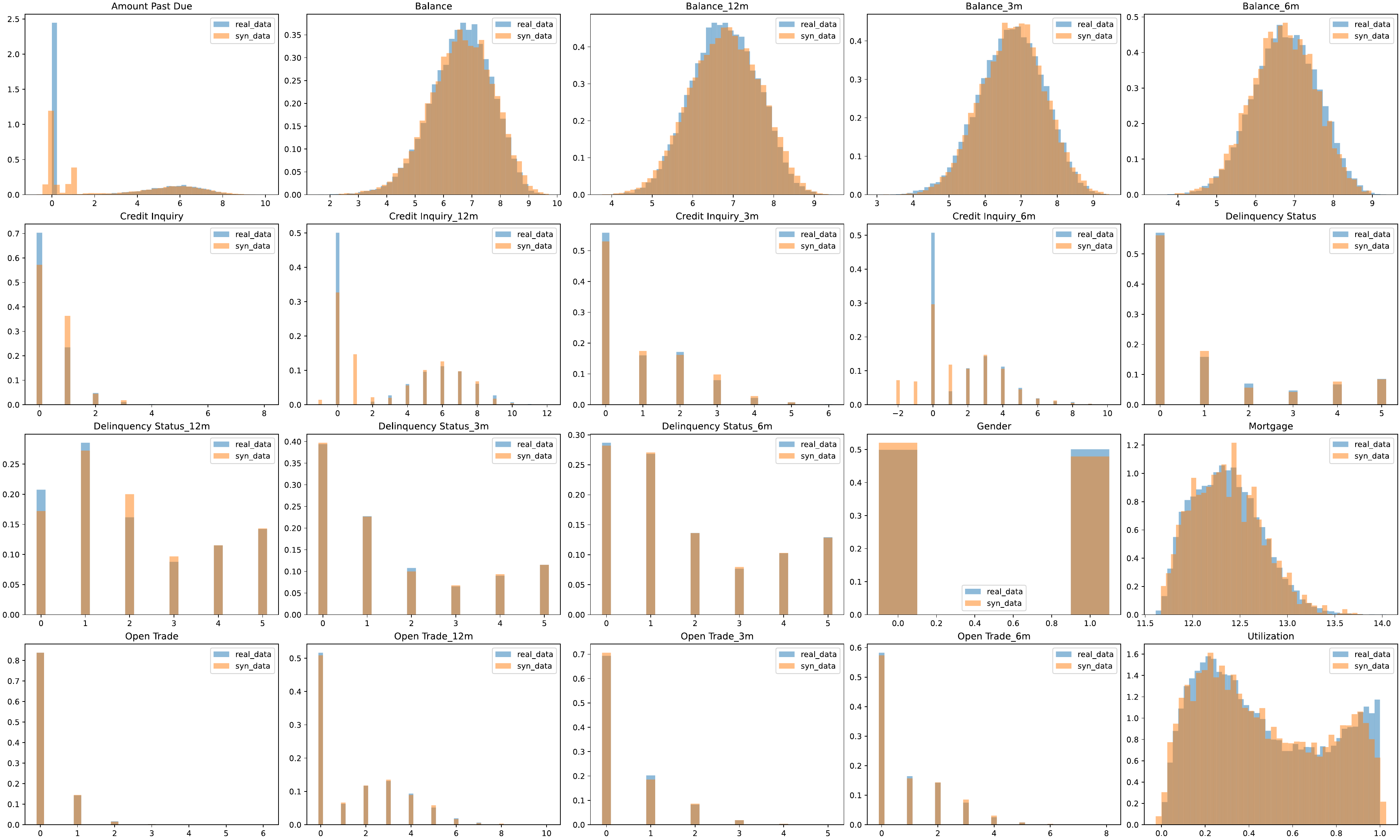}
    \hspace{0.2cm}\includegraphics[width=0.3\textwidth]{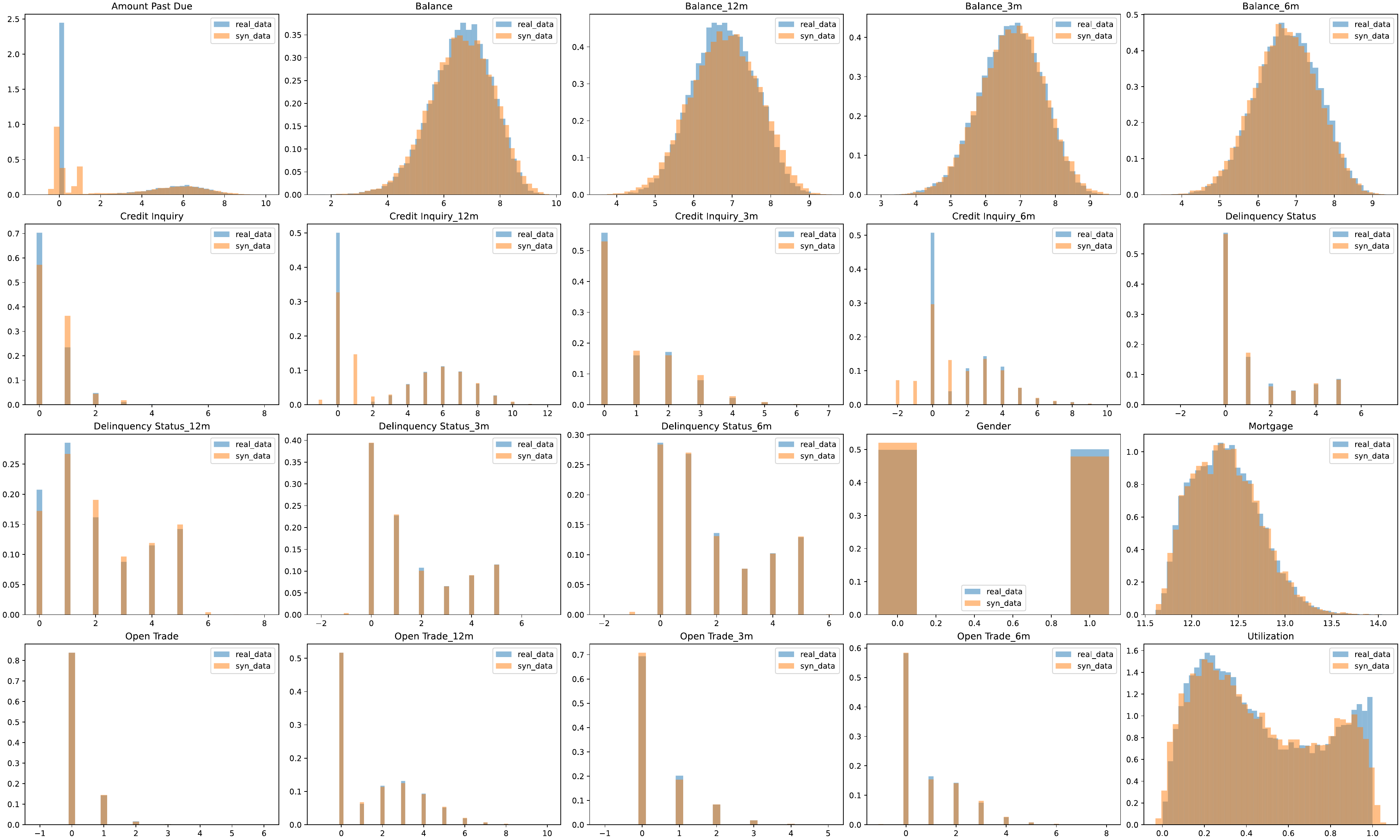}
    \hspace{0.2cm}\includegraphics[width=0.3\textwidth]{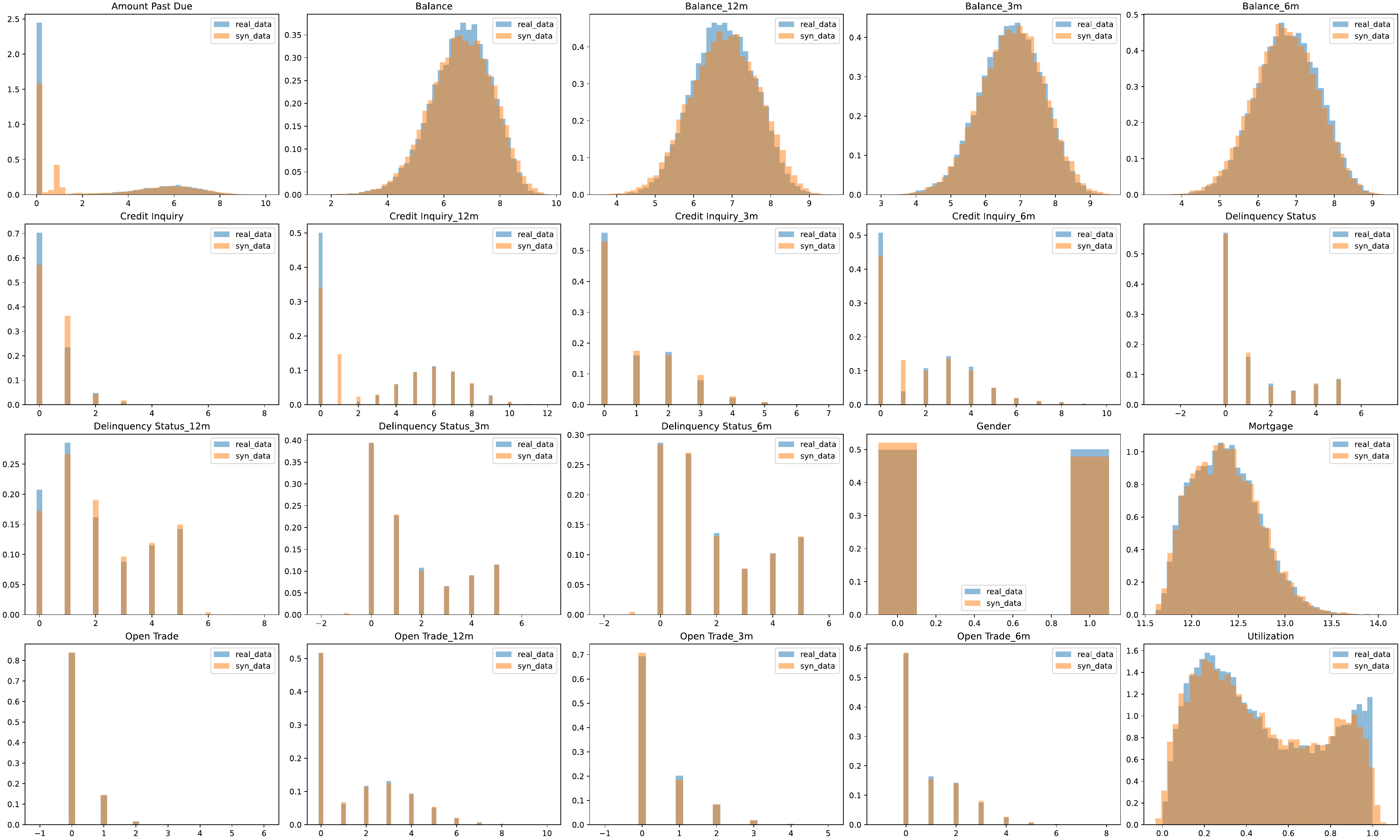}
     \\
    \vspace{0.3cm}
    \raisebox{0.1\height}{\makebox[0.01\textwidth]{\rotatebox{90}{\makecell[c]{\small{ \quad\quad$\epsilon=0.001$} }}}}
    \includegraphics[width=0.3\textwidth]{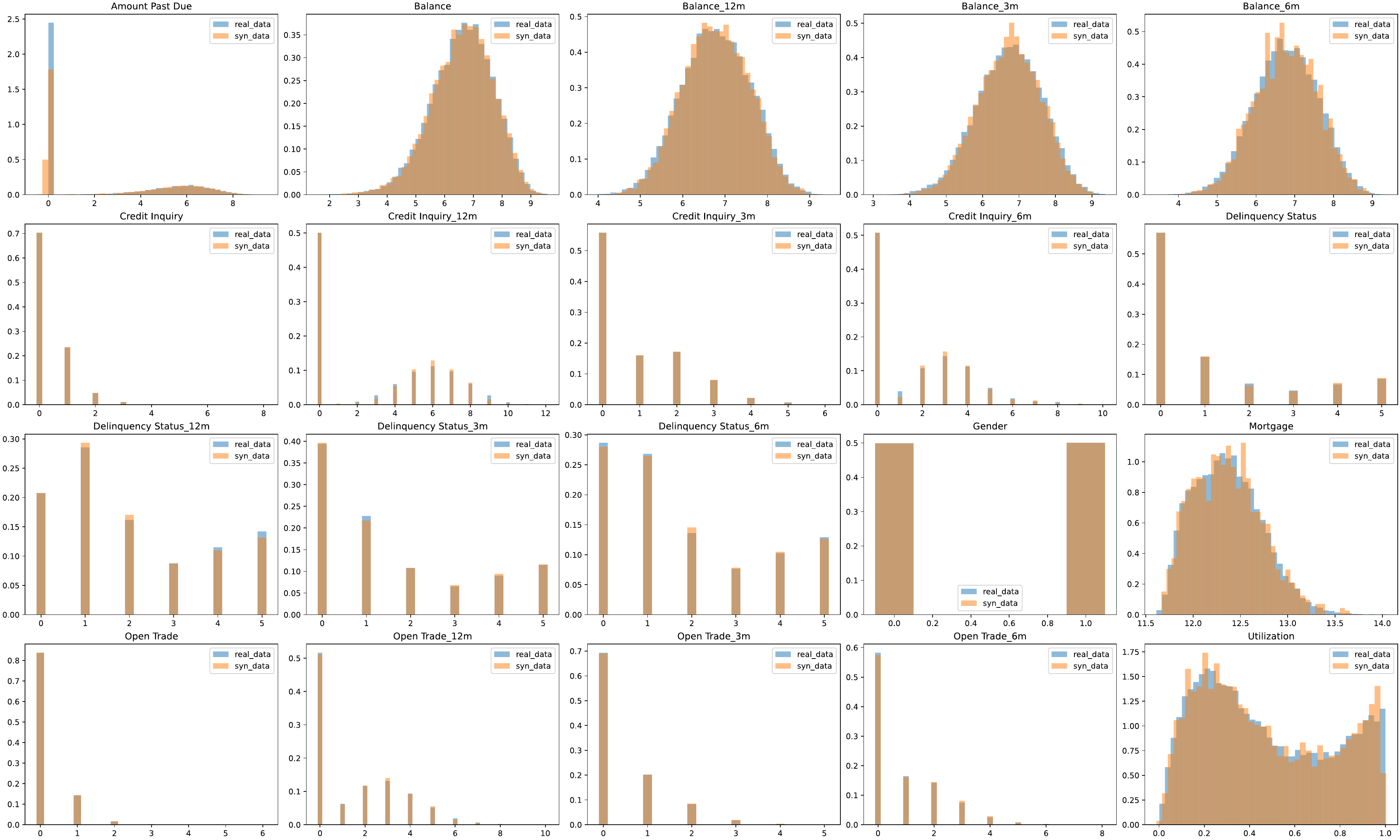}
    \hspace{0.2cm}\includegraphics[width=0.3\textwidth]{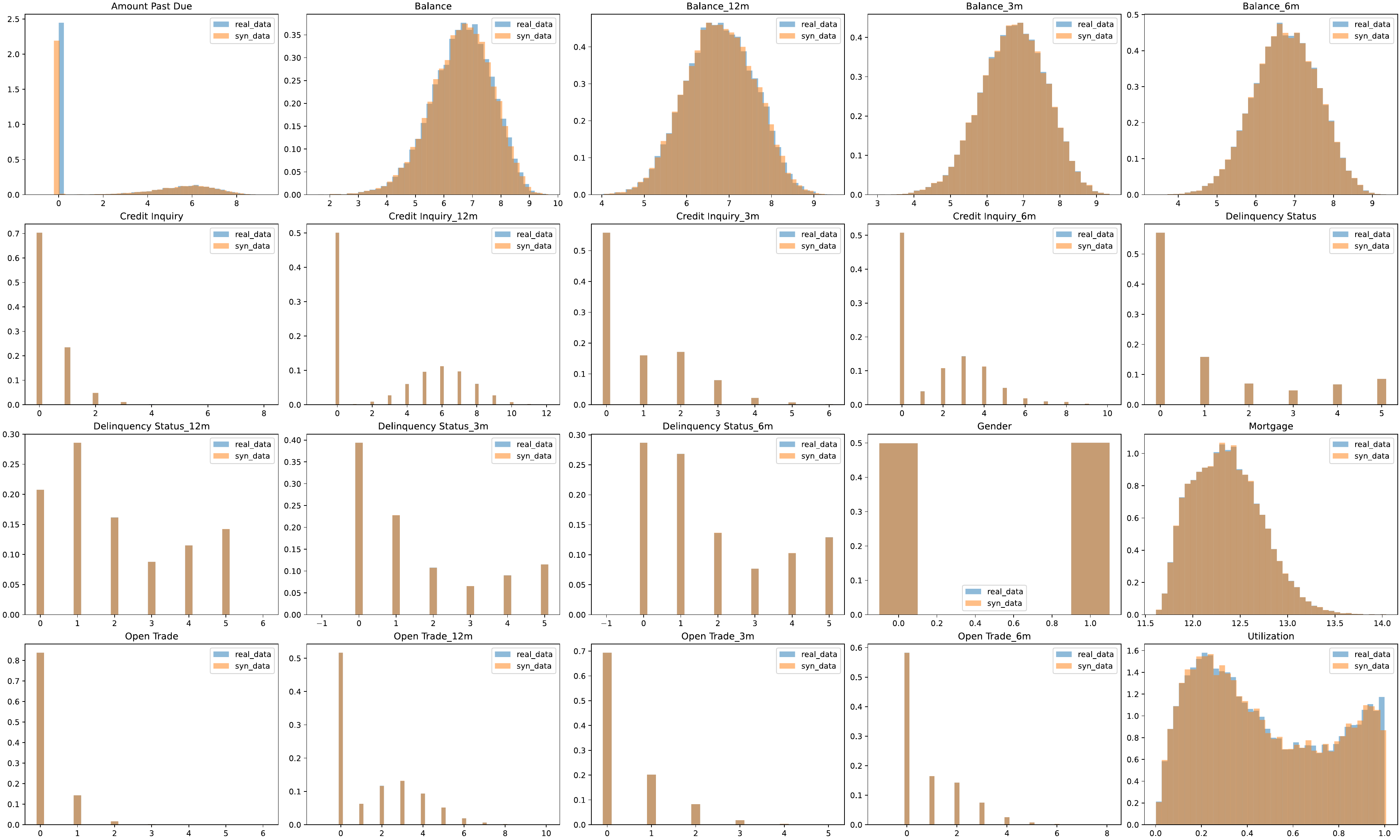}
    \hspace{0.2cm}\includegraphics[width=0.3\textwidth]{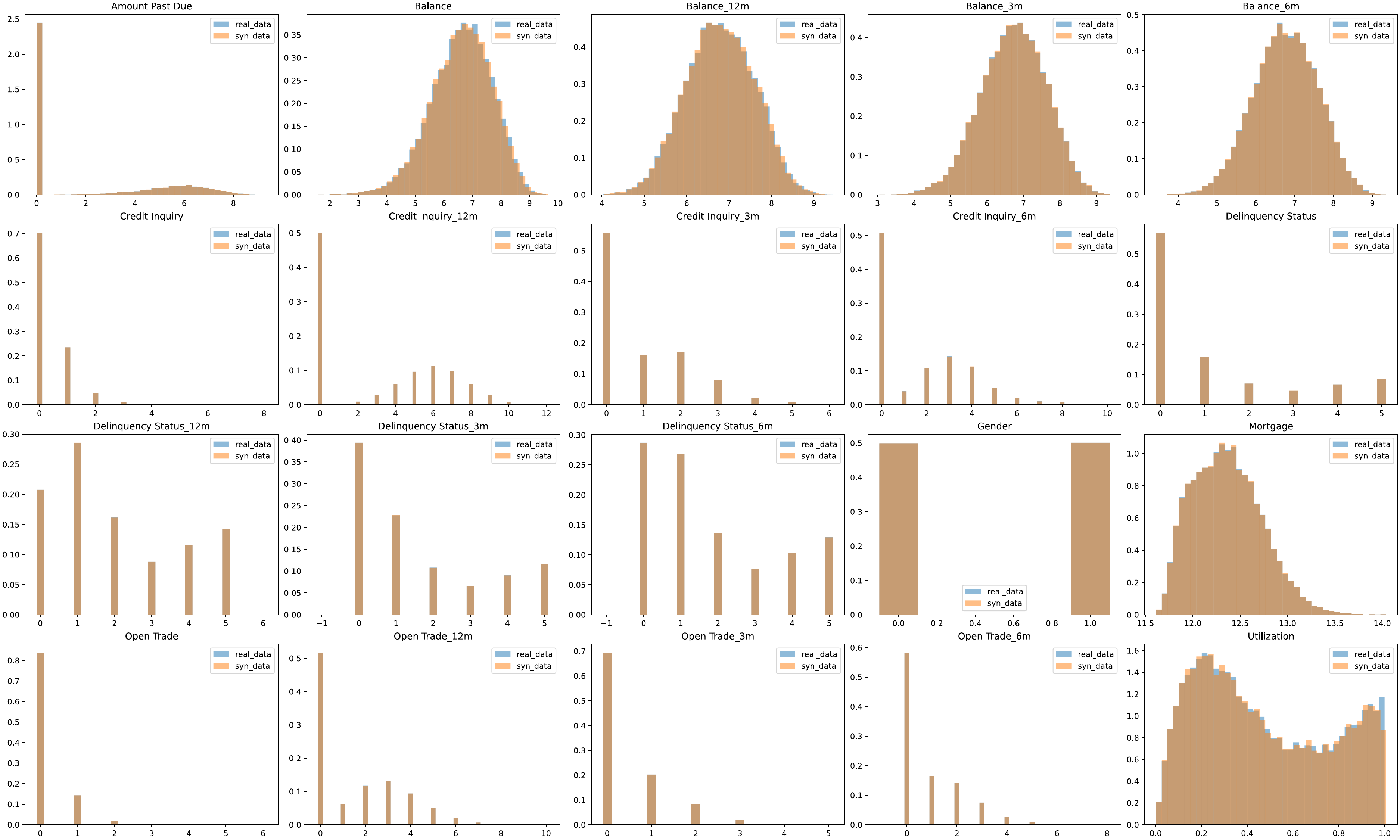}
    
    \caption{A table with tabularized model-based perturbation hyperparameters $\epsilon$ and different estimation strategy. $\hat{X_{\epsilon}}$ represents the synthetic output for trained PCA+XGboost model with perturbation $\epsilon$ on the pca latents. $X+\hat{X_{\epsilon}}-\hat{X}$ stands for the model based perturbation with the adjustment for possible model misspecification $\hat{X} = \hat{X_0}$. The (truncated) group applies postprocessing to simulated data to satisfy reality constraints (for example Credit inquiry and amount past due features cannot be negative). The PCA latents are chosen to be 10 and XGboost has maximum depth 2 and boosting round 200.}  
    \label{robustness_vis}
\end{figure*}

%% file: baselines.tex
\begin{figure*}[!htbp] \centering
    \makebox[0.01\textwidth]{}
    \makebox[0.3\textwidth]{\Large Mortgage}
    \makebox[0.3\textwidth]{\Large Credit Inquiry}
    \makebox[0.3\textwidth]{\Large Amount Past Due}
   
    \raisebox{0.1\height}{\makebox[0.01\textwidth]{\rotatebox{90}{\makecell[c]{\small{ \quad\quad} }}}}
    \includegraphics[width=0.3\textwidth]{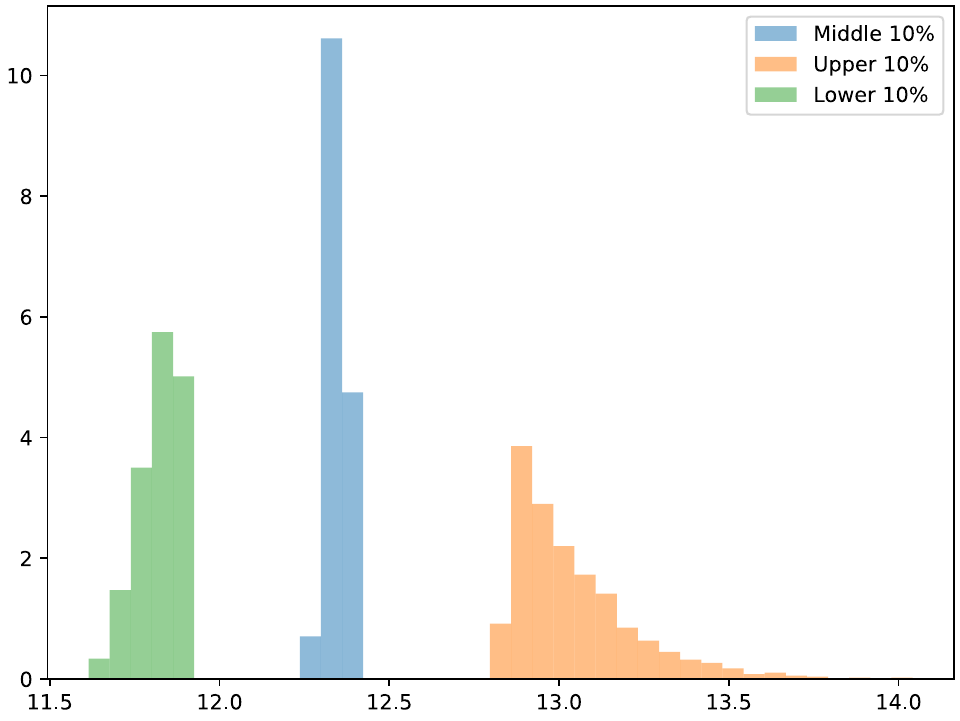}
    \hspace{0.2cm}\includegraphics[width=0.3\textwidth]{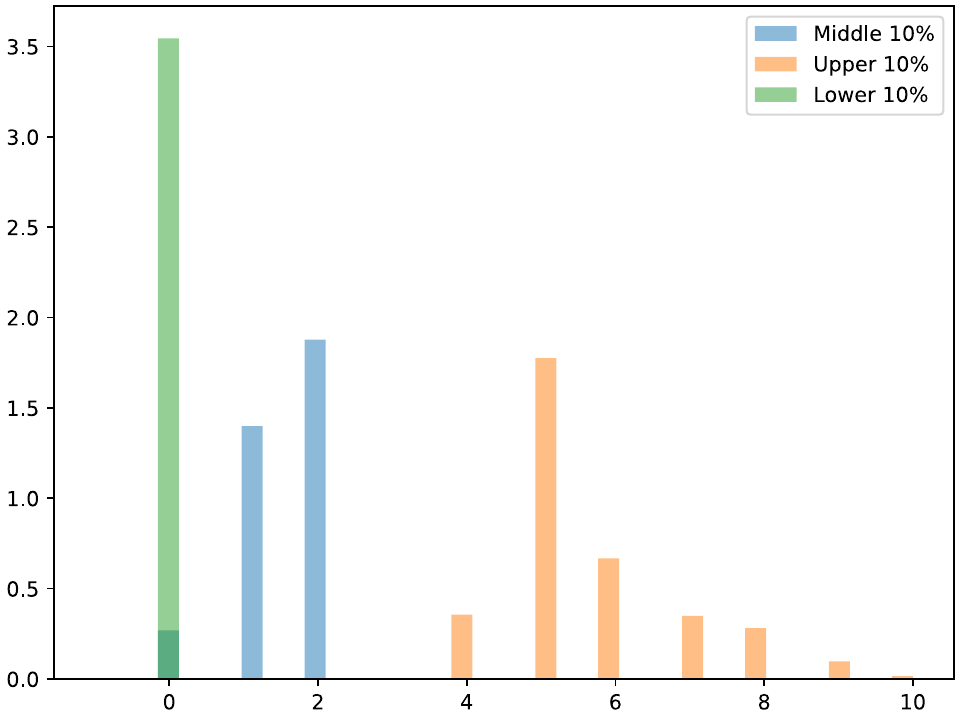}
    \hspace{0.2cm}\includegraphics[width=0.3\textwidth]{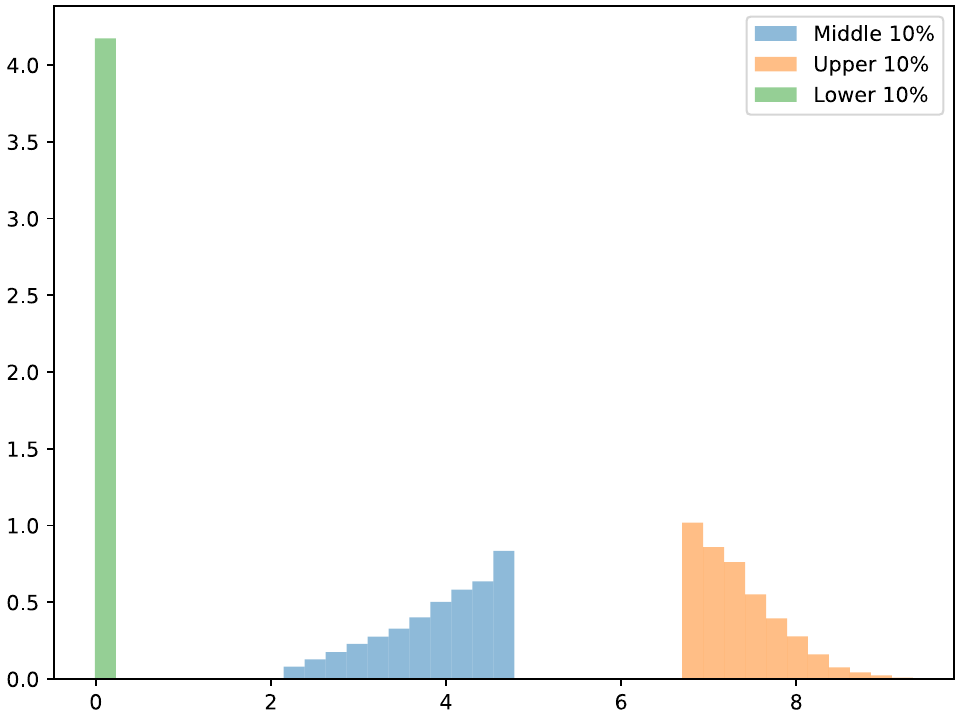}
    
    \caption{The baselines selected distribution landscape for different variables across selected ``Lower", ``Middle", ``Upper" group. The Credit Inquiry variable in the middle in credit inquiry 6m.}  
    \label{fig:baselines}
\end{figure*}

%% file: mortgage.tex
\begin{figure*}[!htbp] \centering
    \makebox[0.01\textwidth]{}
    \makebox[0.14\textwidth]{\small $\epsilon=1$}
    \makebox[0.14\textwidth]{\small $\epsilon=0.1$}
    \makebox[0.14\textwidth]{\small \quad$\epsilon=0.05$}
    \makebox[0.14\textwidth]{\small \quad$\epsilon=0.01$}
    \makebox[0.14\textwidth]{\small \quad$\epsilon=0.005$}
    \makebox[0.14\textwidth]{\small \quad$\epsilon=0.001$}
    \\
    \raisebox{0.1\height}{\makebox[0.01\textwidth]{\rotatebox{90}{\makecell[c]{\scriptsize{ \quad\quad$\hat{X_{\epsilon}}$} }}}}
    \includegraphics[width=0.14\textwidth]{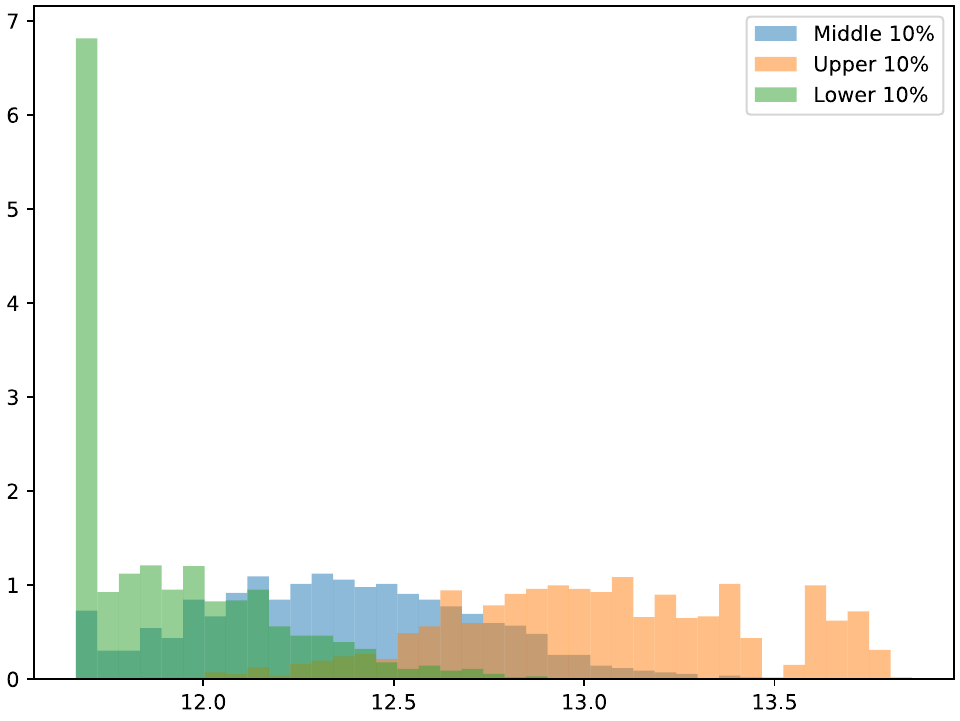}
    \includegraphics[width=0.14\textwidth]{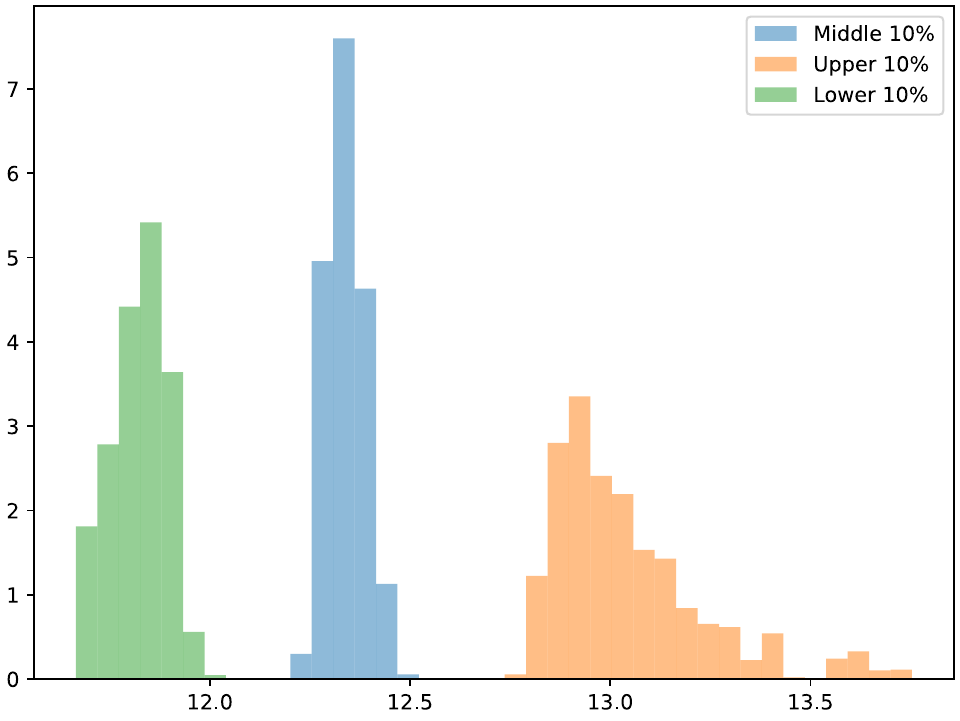}
    \includegraphics[width=0.14\textwidth]{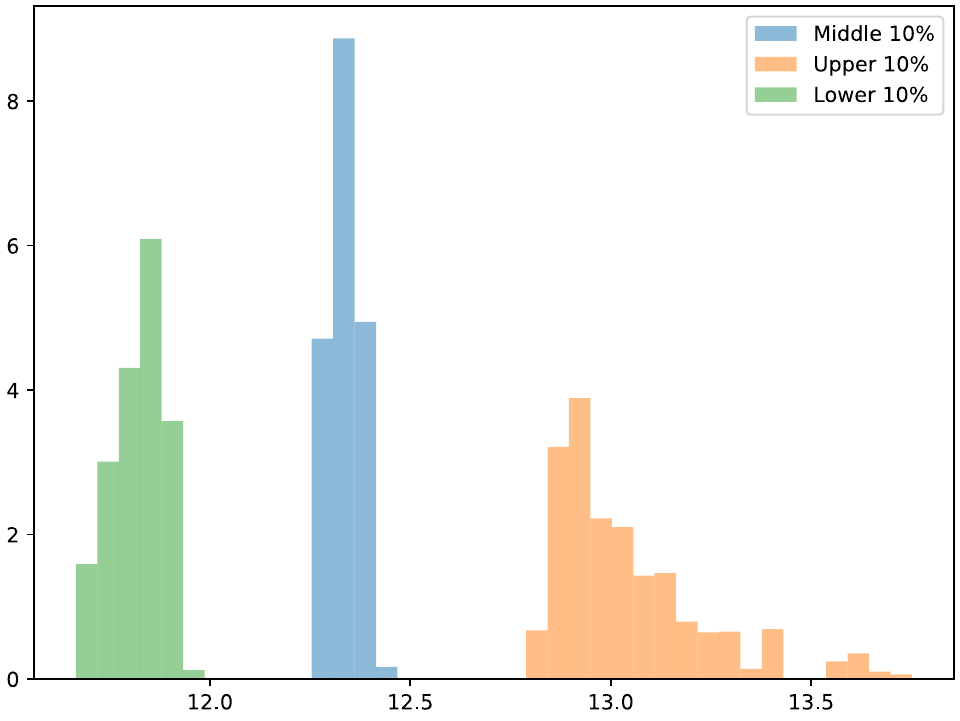}
    \includegraphics[width=0.14\textwidth]{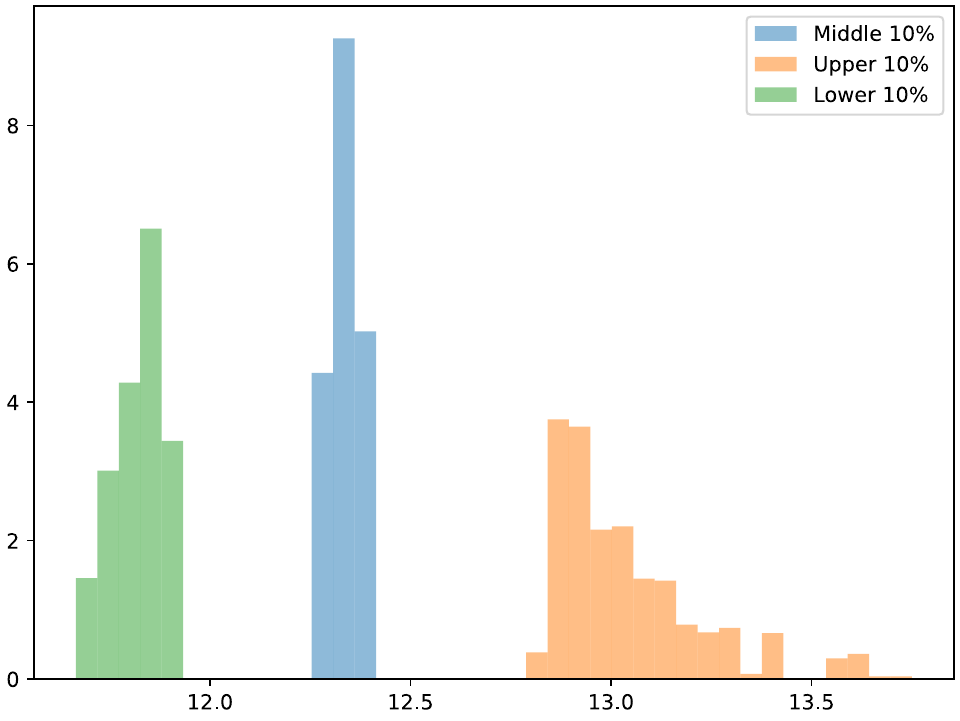}
    \includegraphics[width=0.14\textwidth]{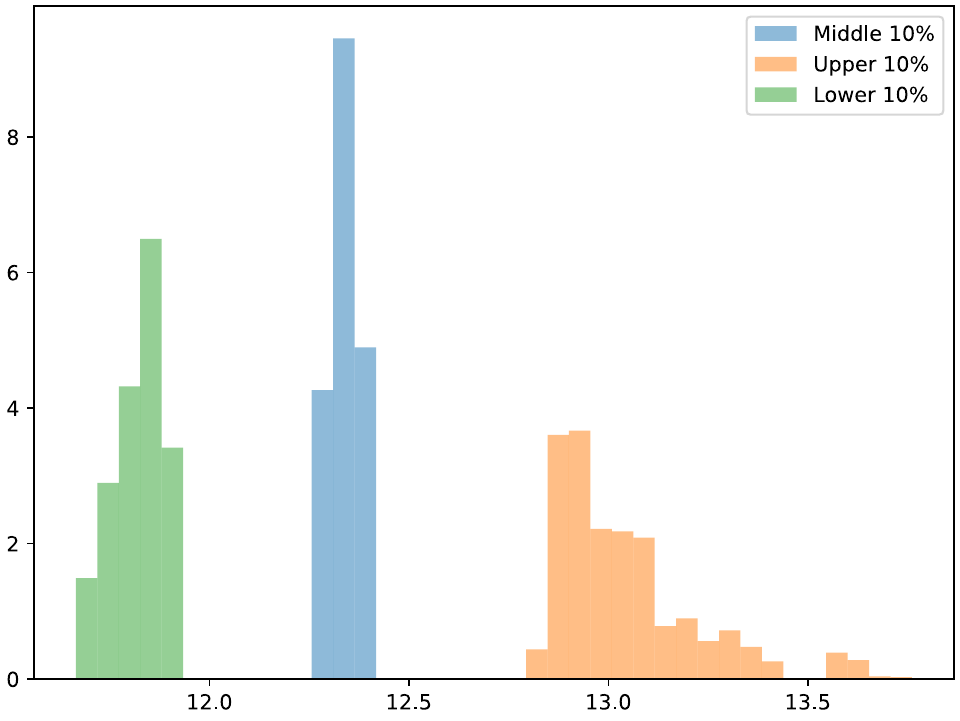}
    \includegraphics[width=0.14\textwidth]{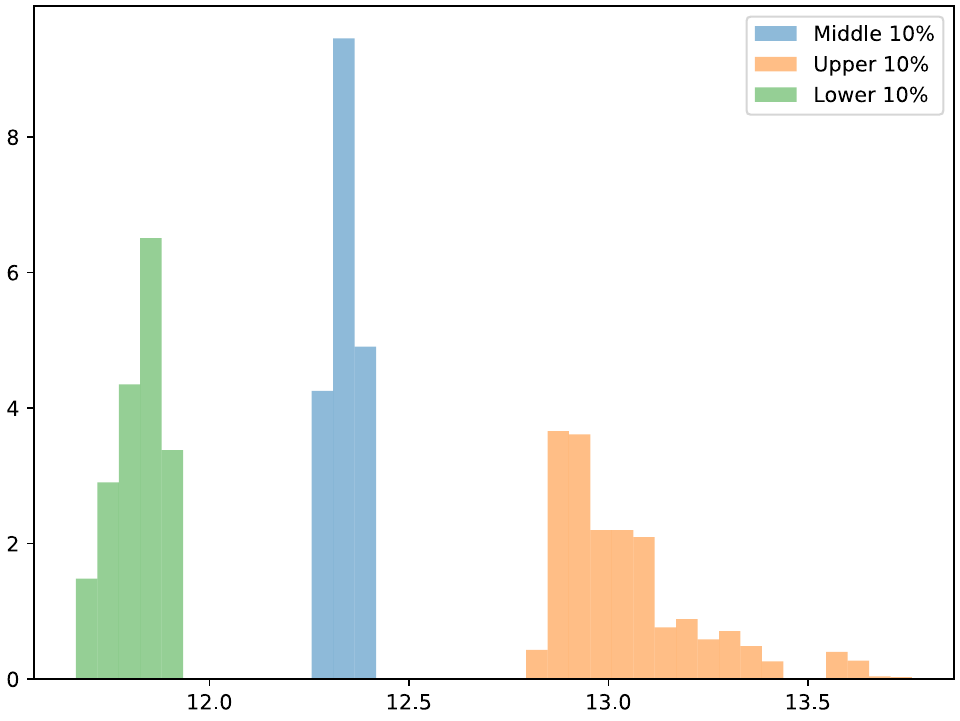}
    
    \raisebox{0.1\height}{\makebox[0.01\textwidth]{\rotatebox{90}{\makecell[c]{\scriptsize{ \quad$\hat{X_{\epsilon}}-X$} }}}}
    \includegraphics[width=0.14\textwidth]{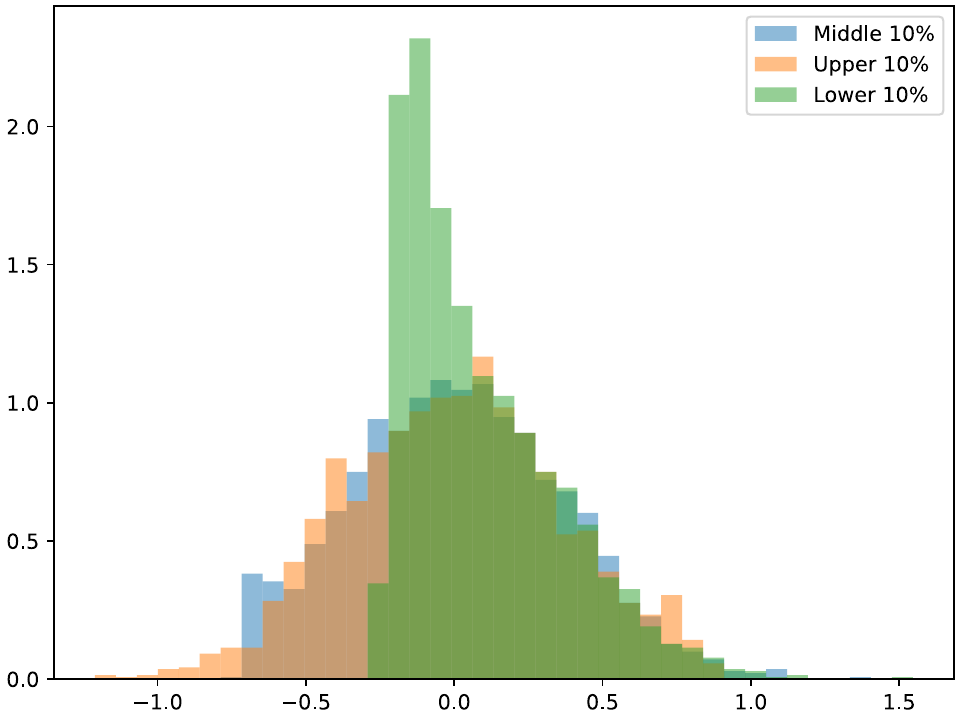}
    \includegraphics[width=0.14\textwidth]{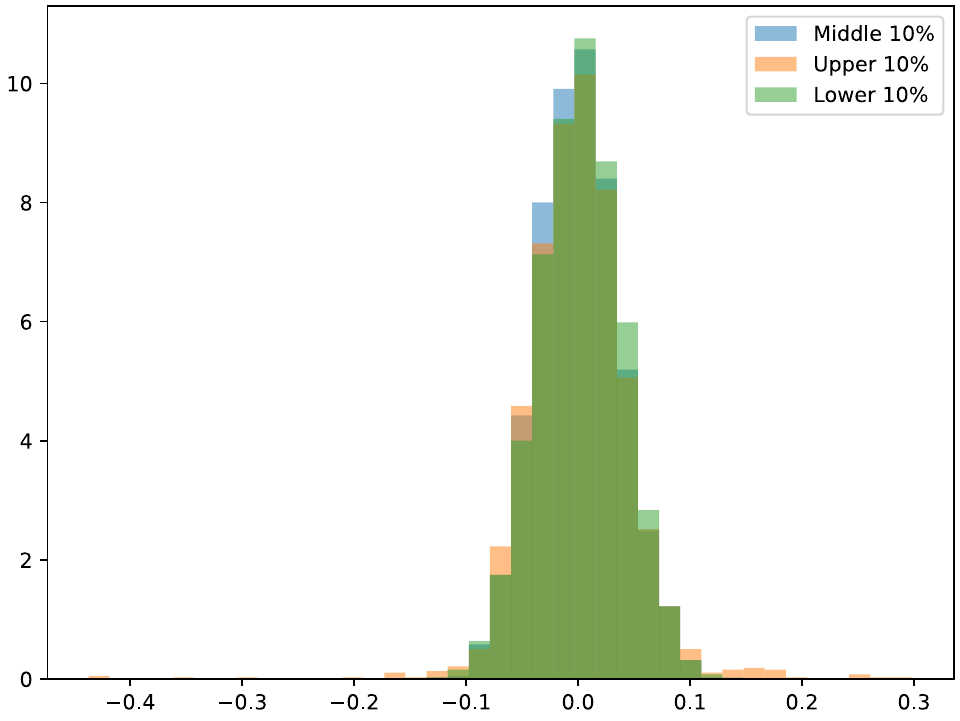}
    \includegraphics[width=0.14\textwidth]{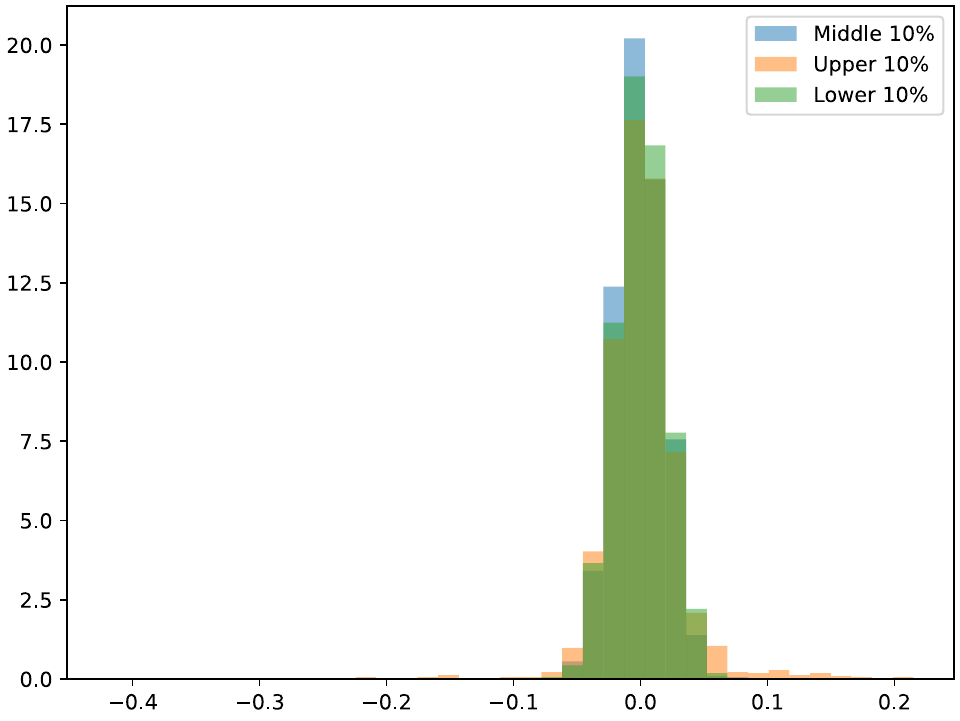}
    \includegraphics[width=0.14\textwidth]{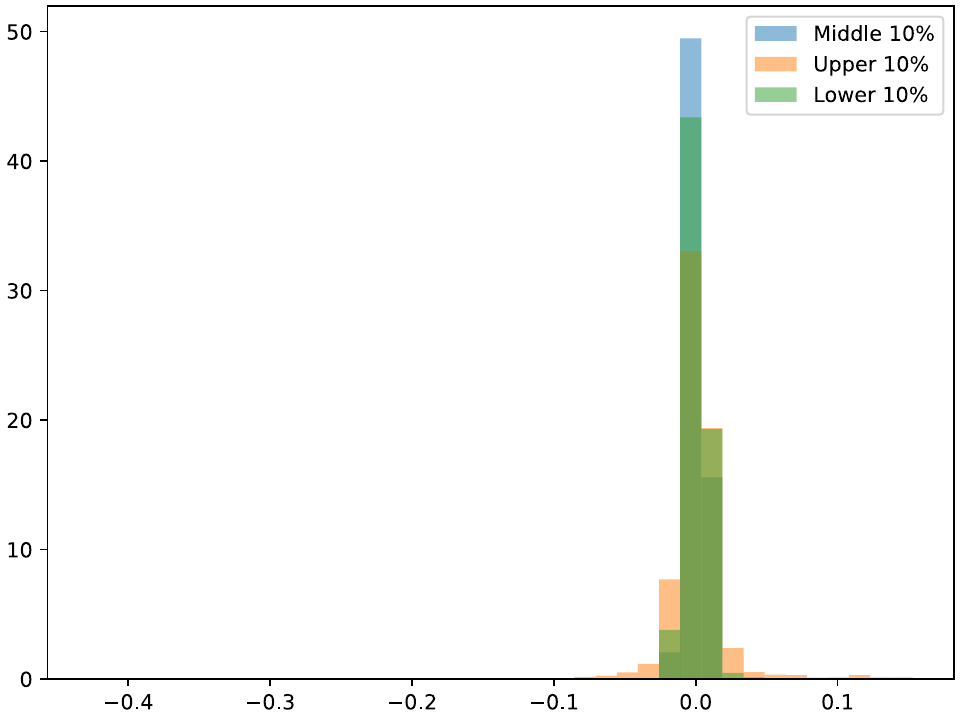}
    \includegraphics[width=0.14\textwidth]{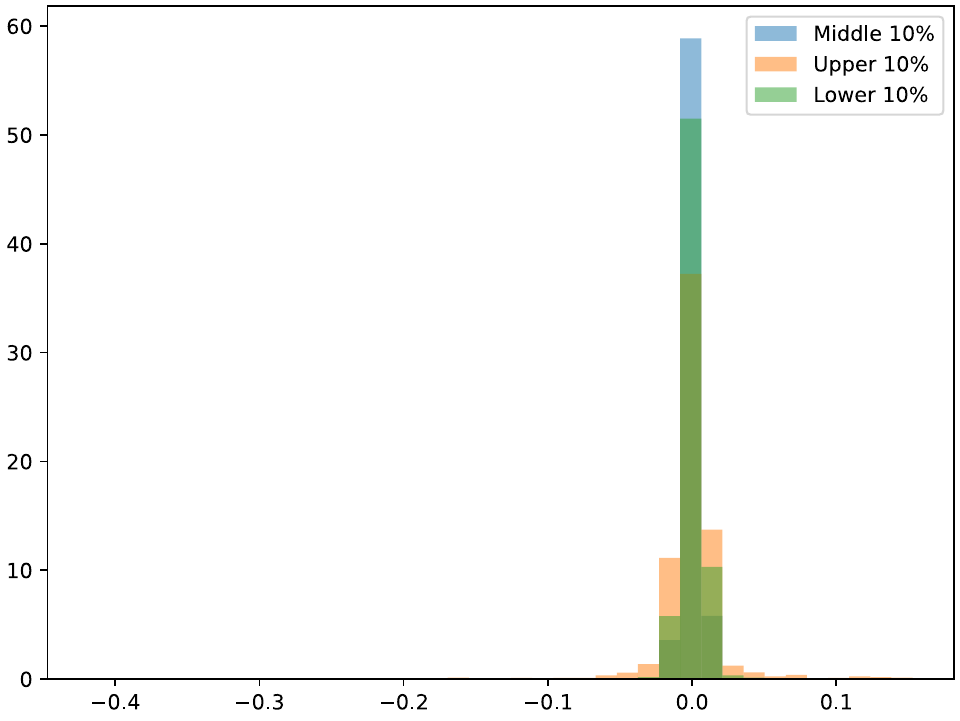}
    \includegraphics[width=0.14\textwidth]{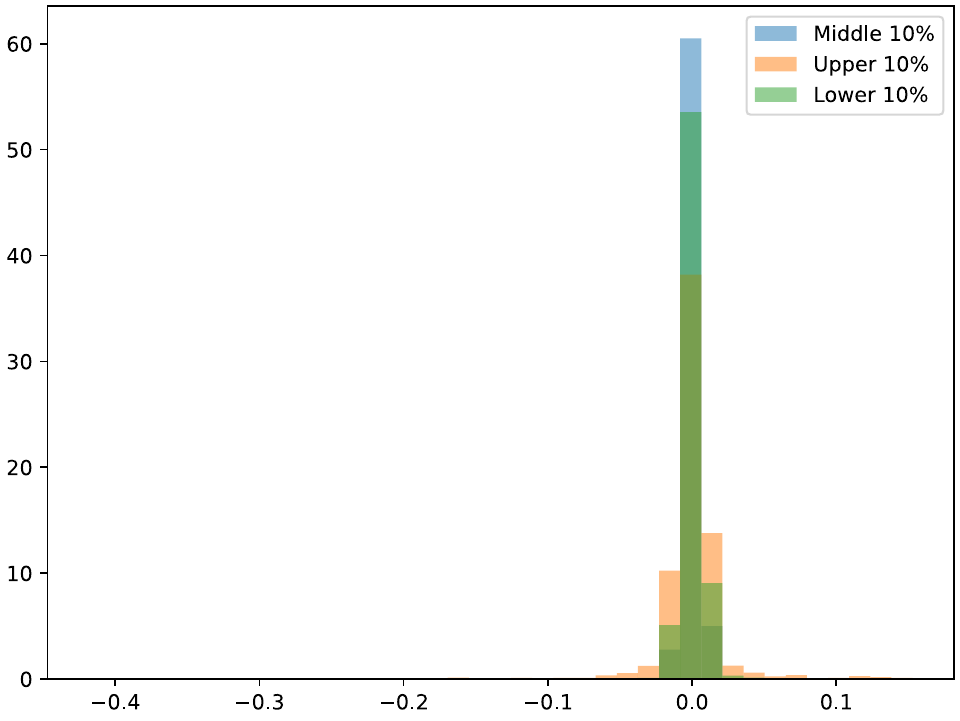}
    


    \raisebox{0.1\height}{\makebox[0.01\textwidth]{\rotatebox{90}{\makecell[c]{\scriptsize{ $X+\hat{X_{\epsilon}}-\hat{X}$} }}}}
    \includegraphics[width=0.14\textwidth]{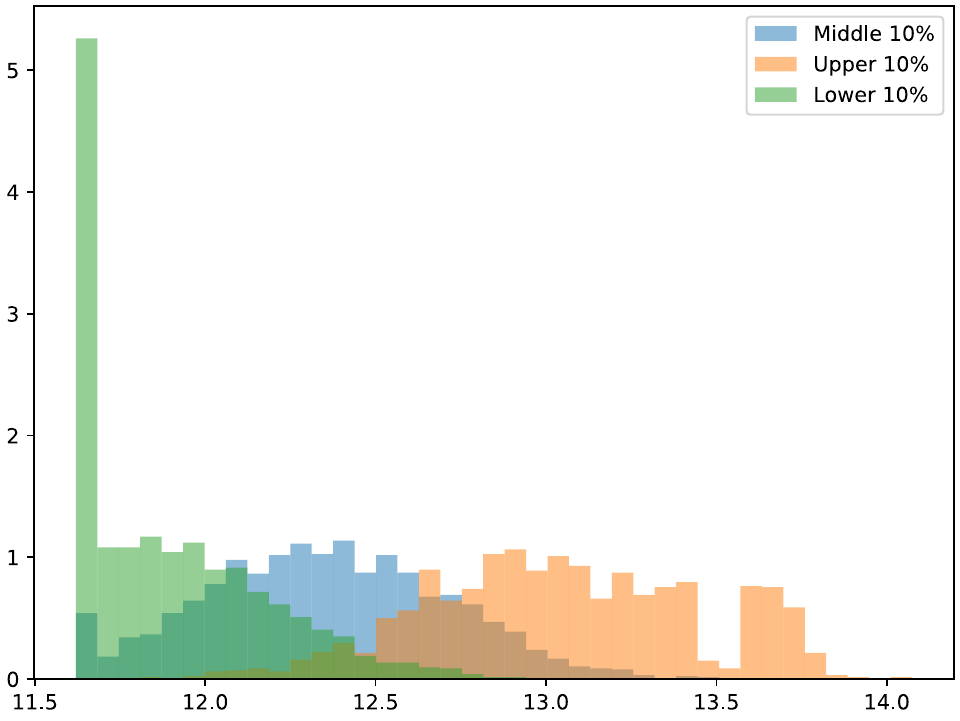}
    \includegraphics[width=0.14\textwidth]{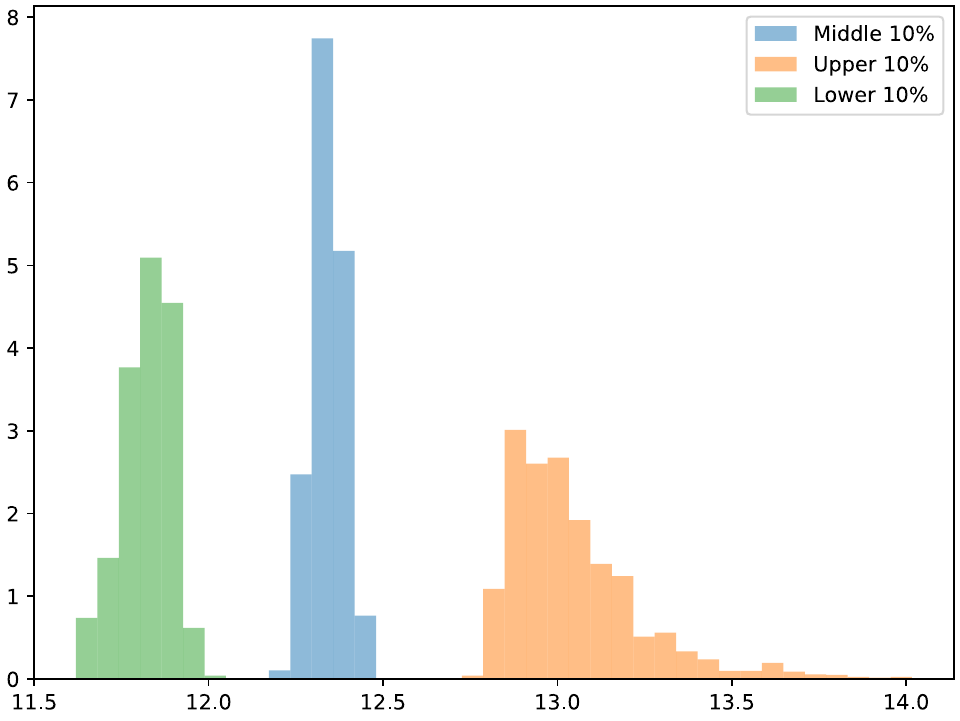}
    \includegraphics[width=0.14\textwidth]{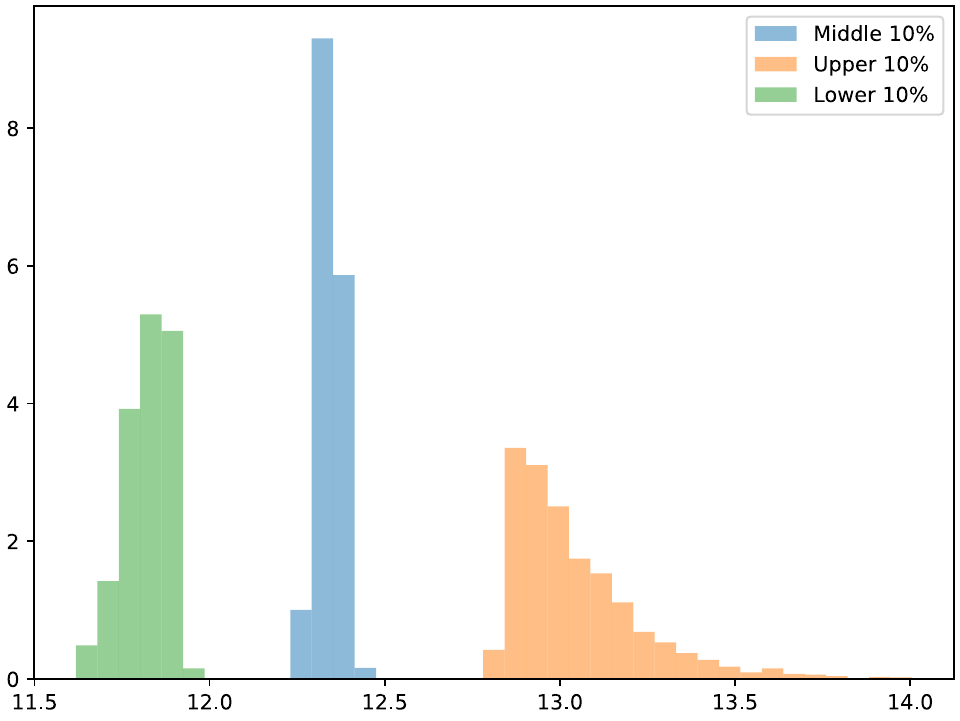}
    \includegraphics[width=0.14\textwidth]{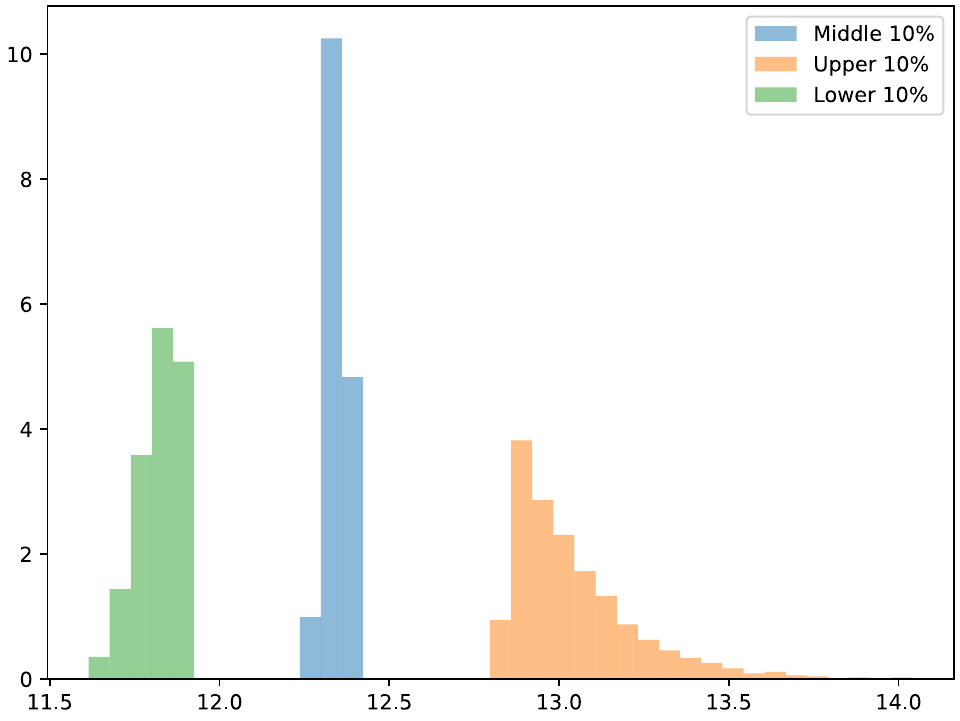}
    \includegraphics[width=0.14\textwidth]{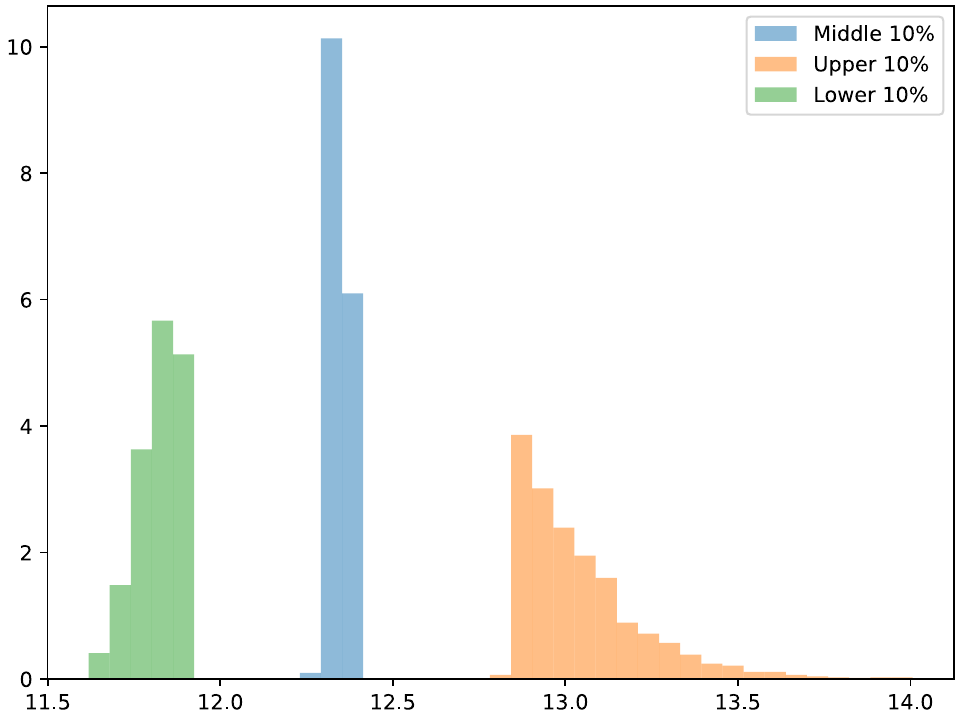}
    \includegraphics[width=0.14\textwidth]{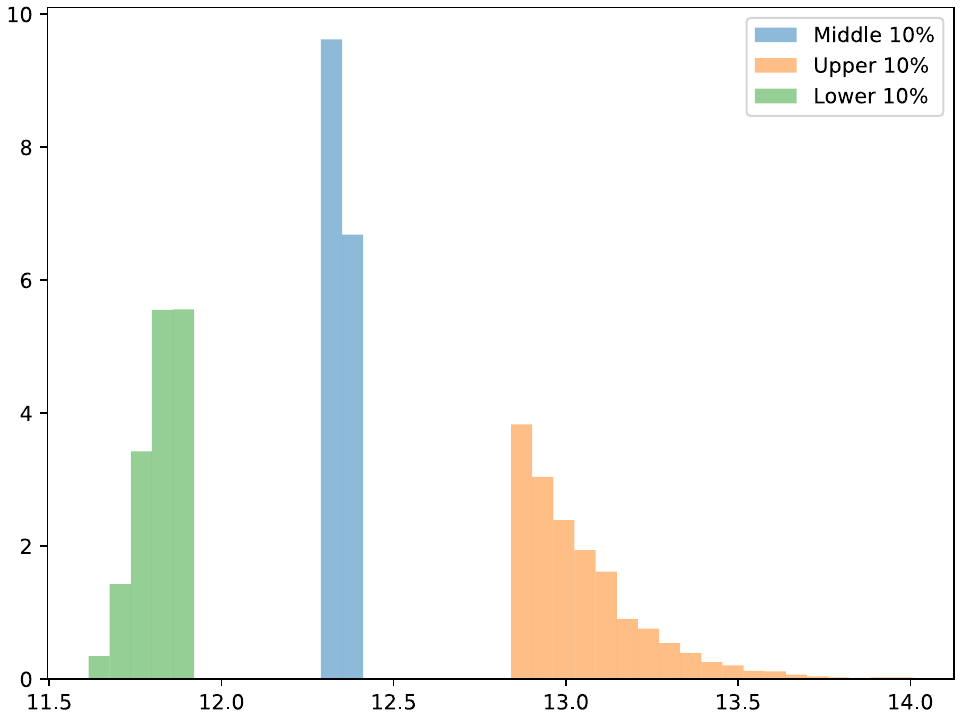}

    \raisebox{0.1\height}{\makebox[0.01\textwidth]{\rotatebox{90}{\makecell[c]{\scriptsize{ \quad$\hat{X_{\epsilon}}-\hat{X}$} }}}}
    \includegraphics[width=0.14\textwidth]{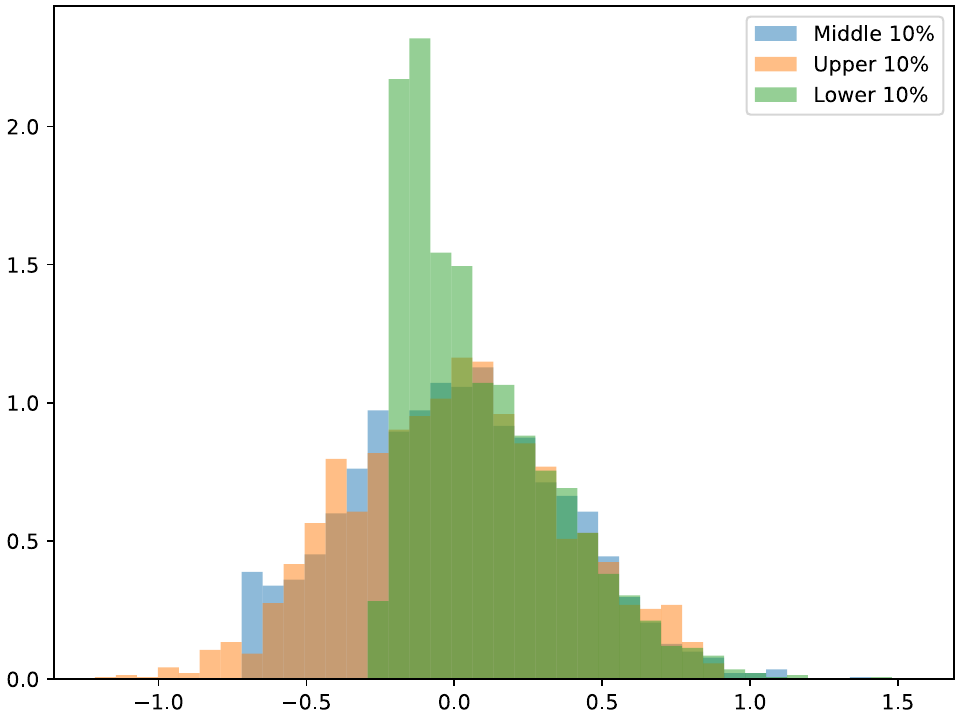}
    \includegraphics[width=0.14\textwidth]{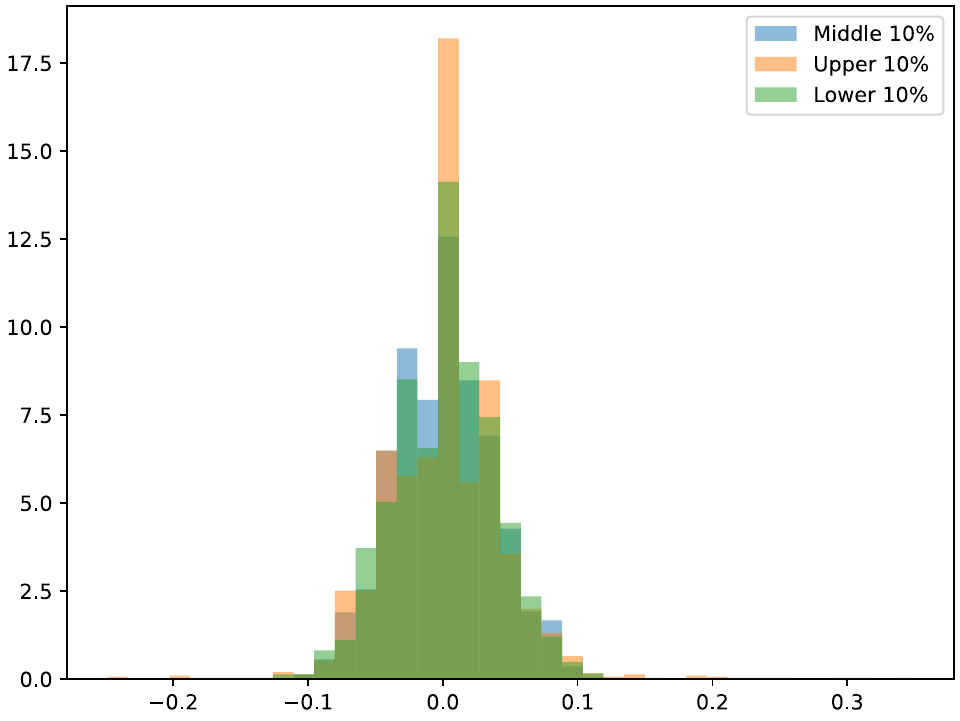}
    \includegraphics[width=0.14\textwidth]{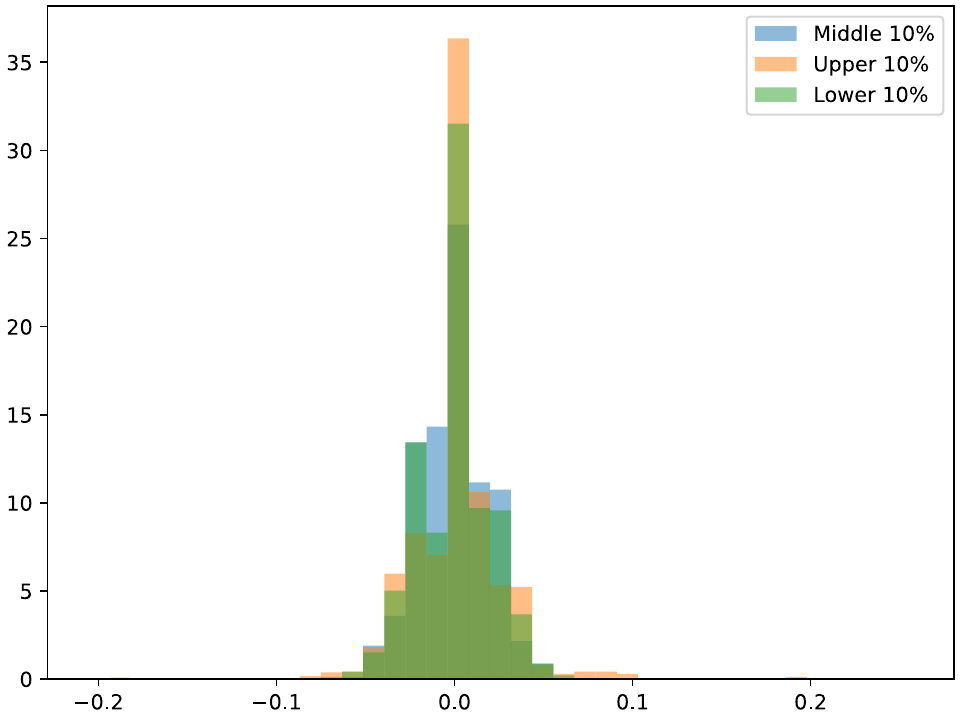}
    \includegraphics[width=0.14\textwidth]{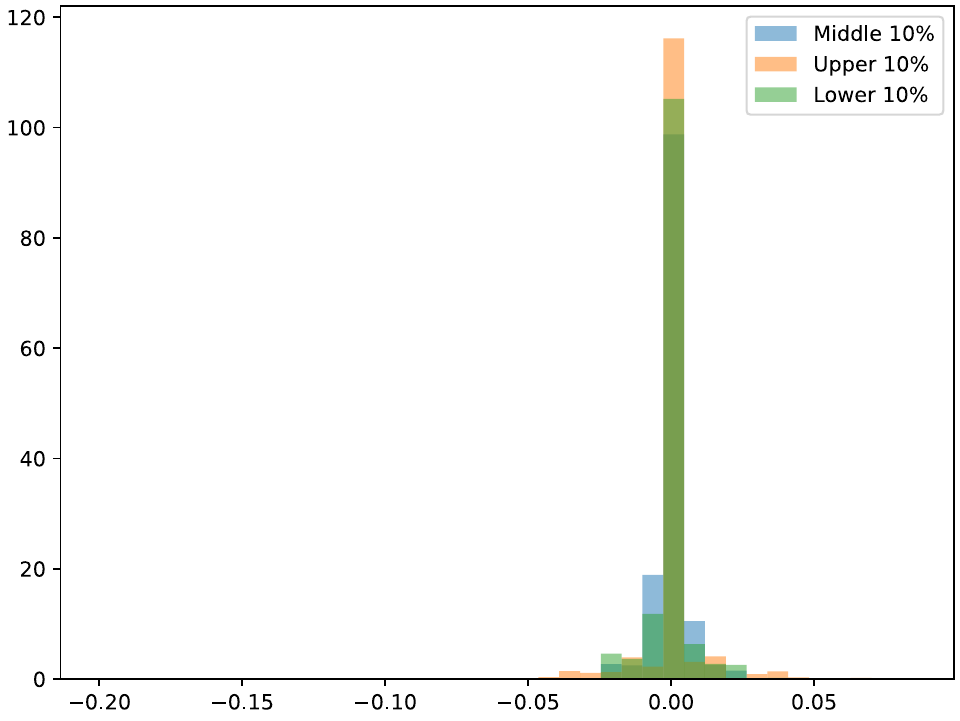}
    \includegraphics[width=0.14\textwidth]{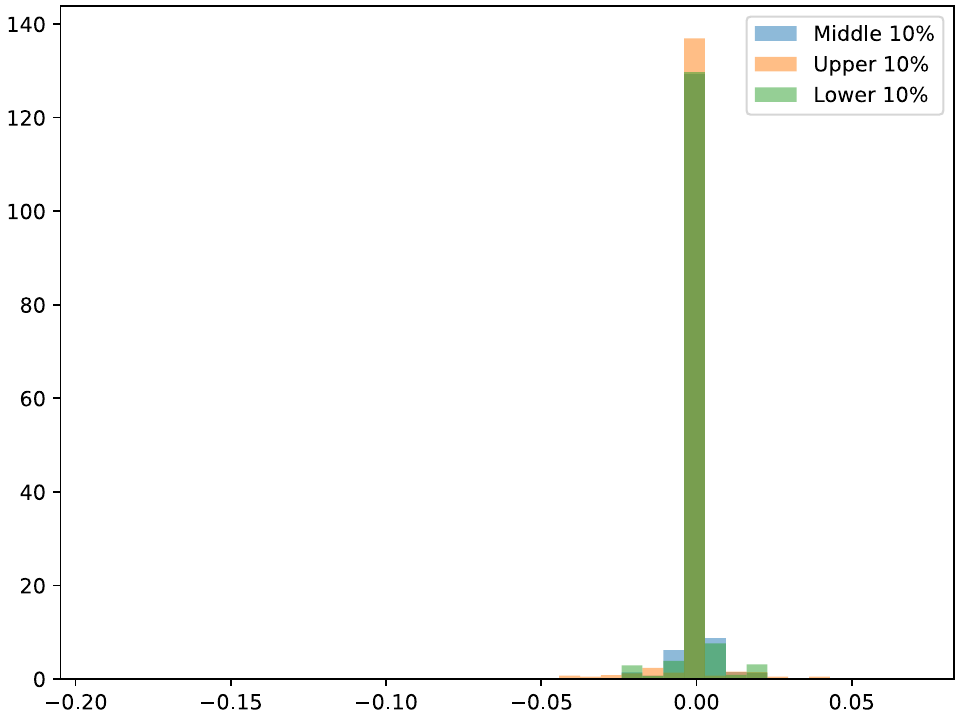}
    \includegraphics[width=0.14\textwidth]{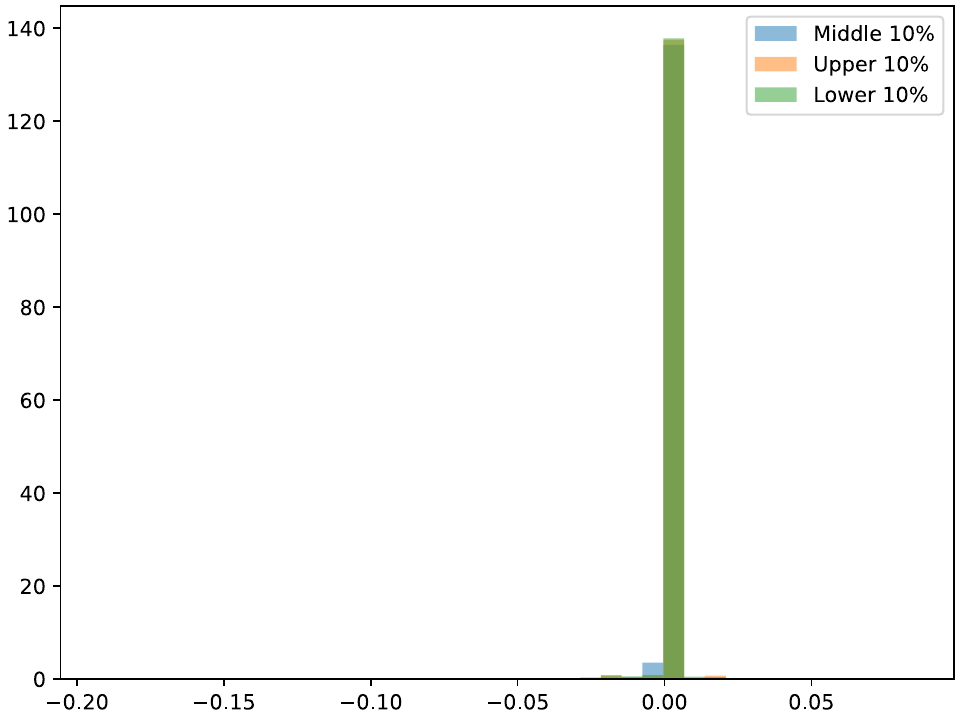}

    \caption{A comparison table for different strategy corresponding to the perturbation result for the \textbf{Mortgage} variable in different region.}  
    \label{fig:region_mortgage}
\end{figure*}

%% file: credit_i.tex
\begin{figure*}[!htbp] \centering
    \makebox[0.01\textwidth]{}
    \makebox[0.14\textwidth]{\small $\epsilon=1$}
    \makebox[0.14\textwidth]{\small $\epsilon=0.1$}
    \makebox[0.14\textwidth]{\small \quad$\epsilon=0.05$}
    \makebox[0.14\textwidth]{\small \quad$\epsilon=0.01$}
    \makebox[0.14\textwidth]{\small \quad$\epsilon=0.005$}
    \makebox[0.14\textwidth]{\small \quad$\epsilon=0.001$}
    \\
    \raisebox{0.1\height}{\makebox[0.01\textwidth]{\rotatebox{90}{\makecell[c]{\scriptsize{ \quad\quad$\hat{X_{\epsilon}}$} }}}}
    \includegraphics[width=0.14\textwidth]{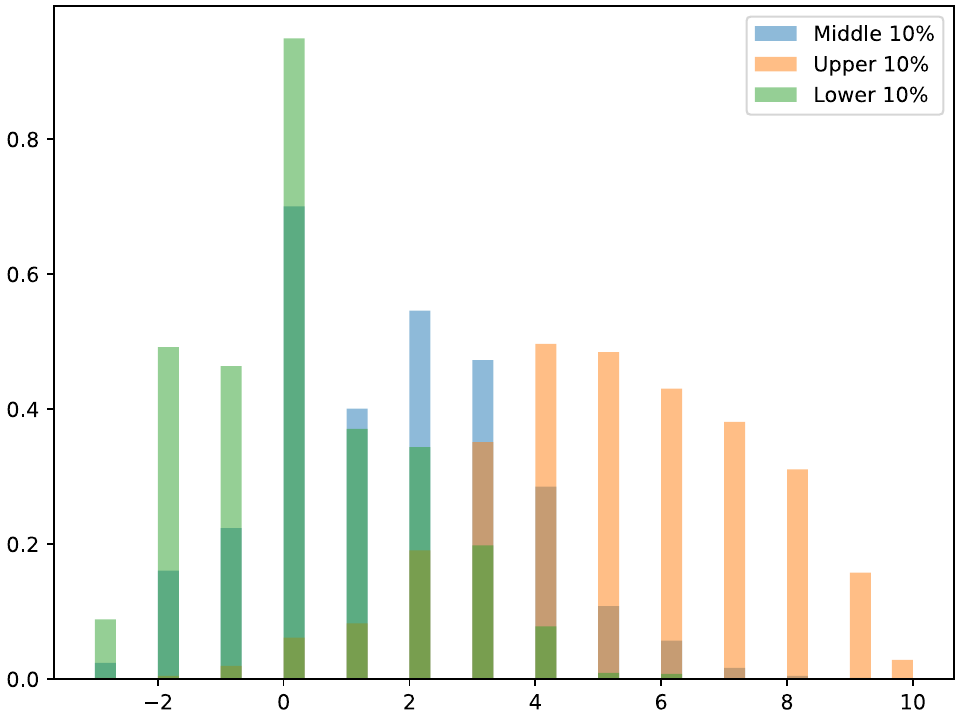}
    \includegraphics[width=0.14\textwidth]{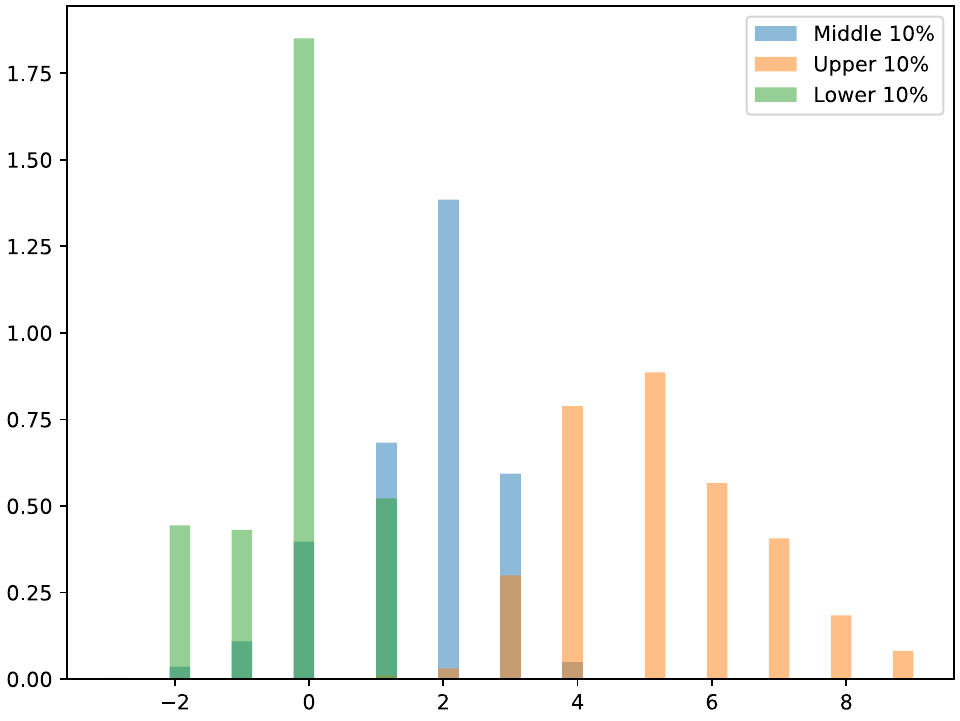}
    \includegraphics[width=0.14\textwidth]{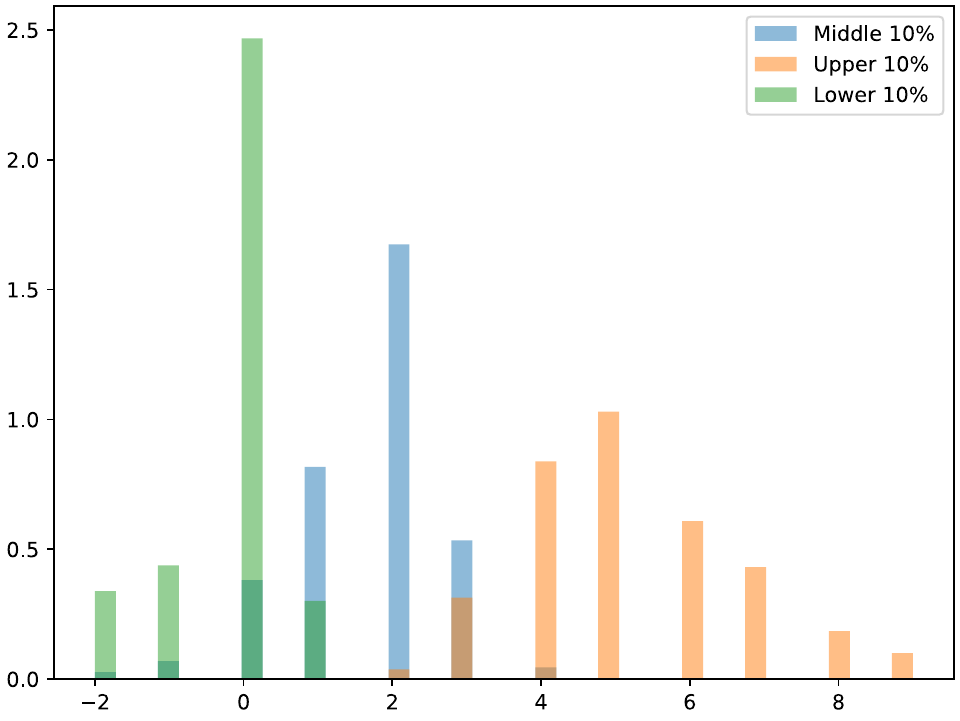}
    \includegraphics[width=0.14\textwidth]{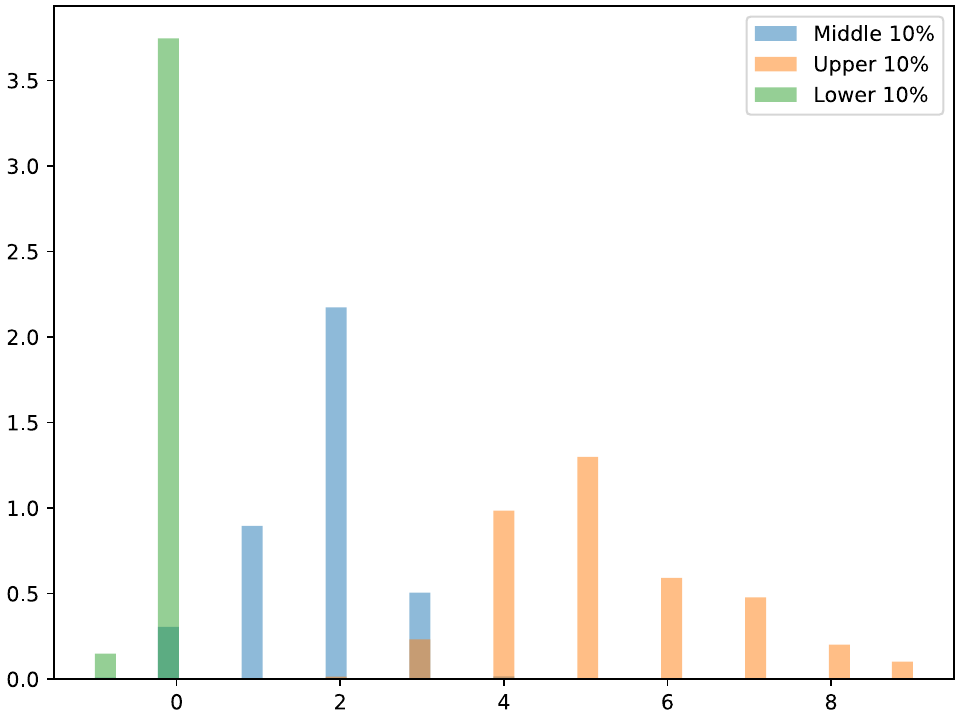}
    \includegraphics[width=0.14\textwidth]{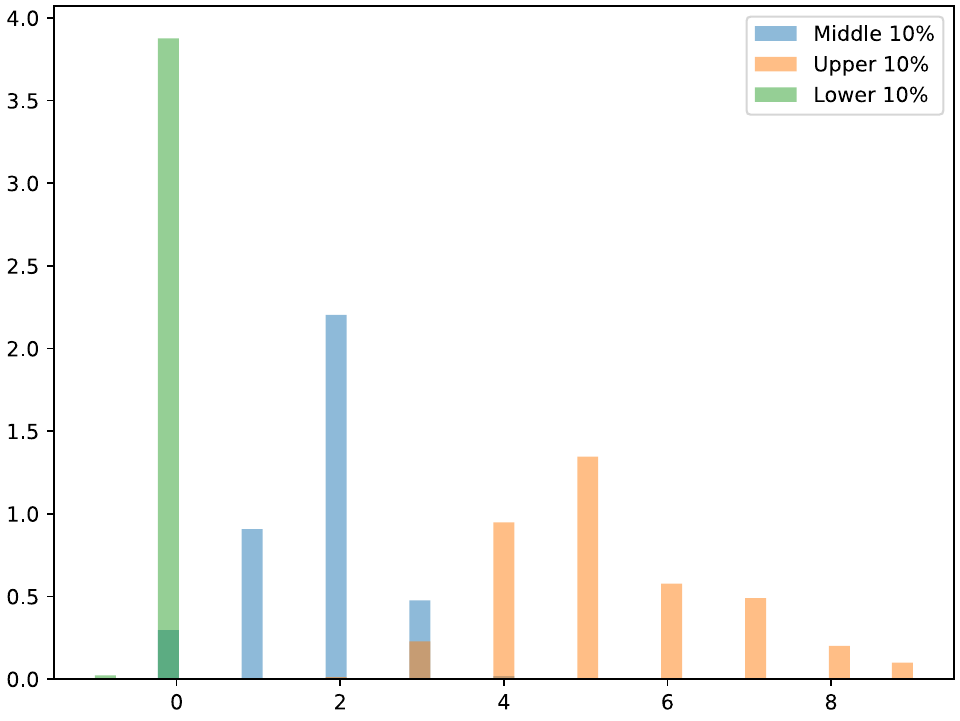}
    \includegraphics[width=0.14\textwidth]{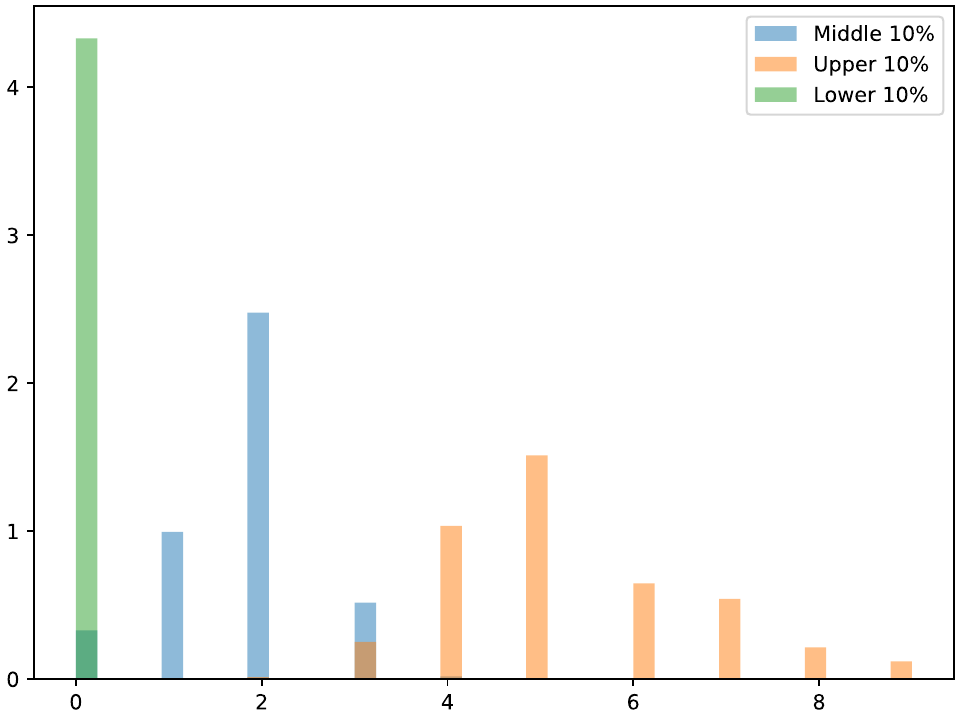}
    
    \raisebox{0.1\height}{\makebox[0.01\textwidth]{\rotatebox{90}{\makecell[c]{\scriptsize{ \quad$\hat{X_{\epsilon}}-X$} }}}}
    \includegraphics[width=0.14\textwidth]{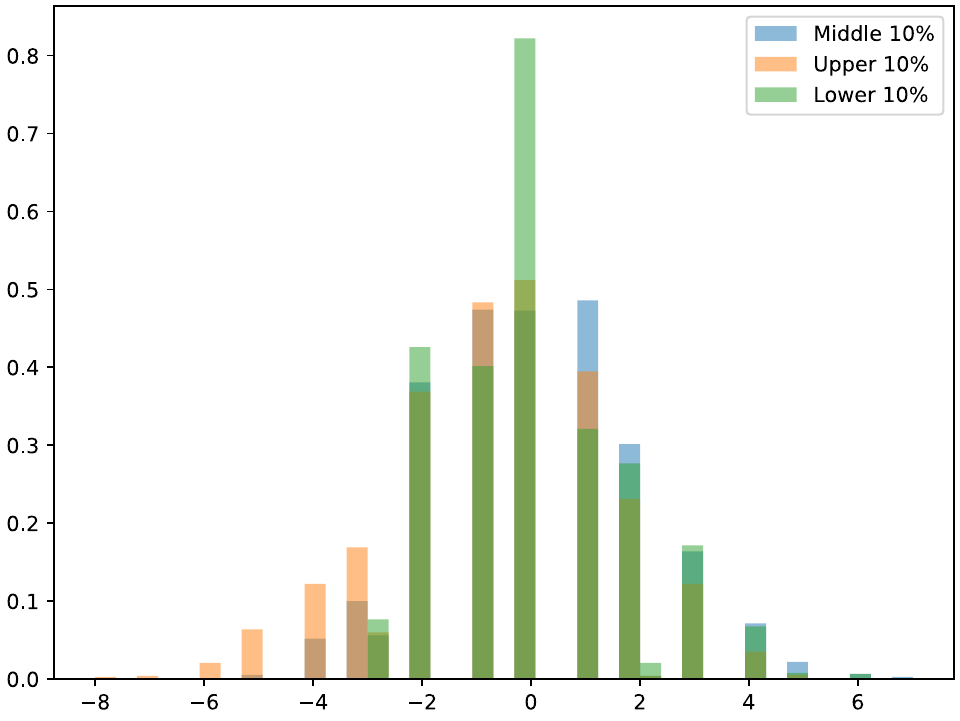}
    \includegraphics[width=0.14\textwidth]{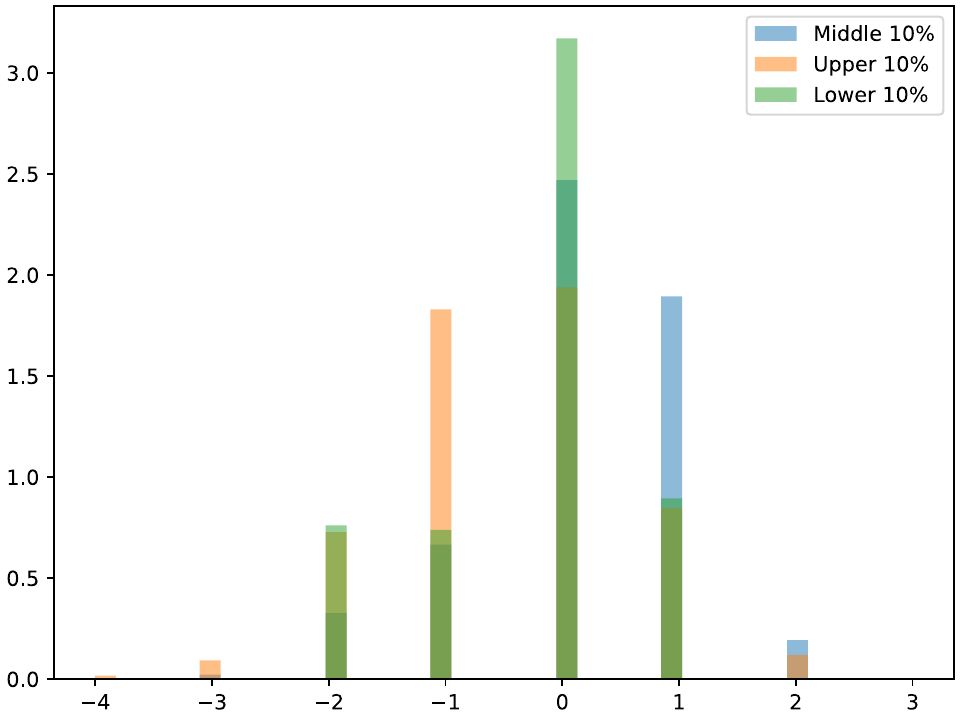}
    \includegraphics[width=0.14\textwidth]{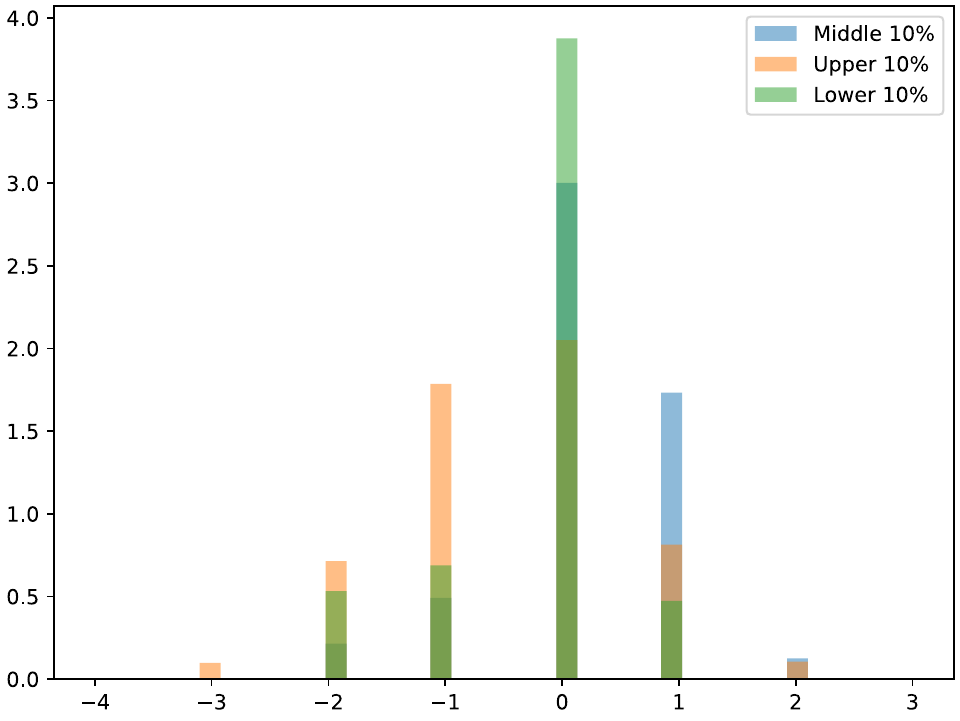}
    \includegraphics[width=0.14\textwidth]{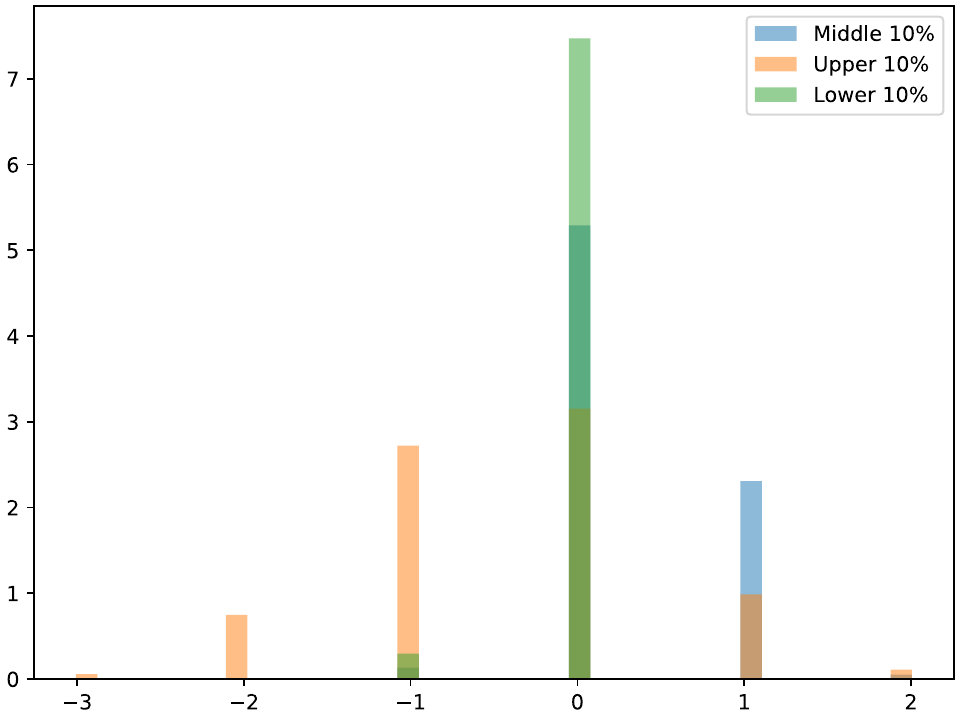}
    \includegraphics[width=0.14\textwidth]{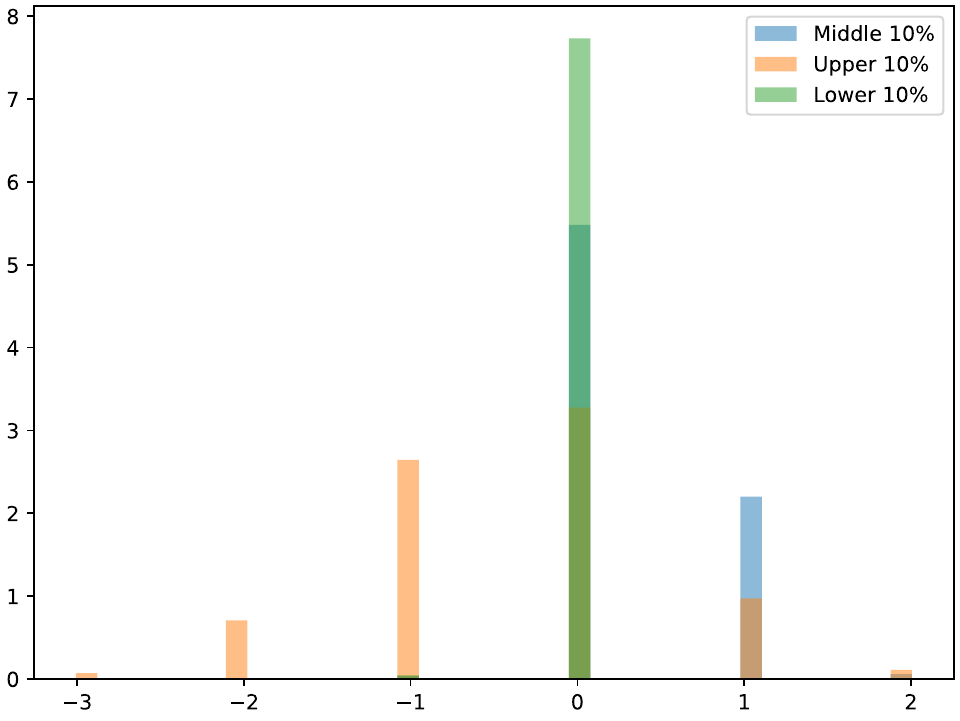}
    \includegraphics[width=0.14\textwidth]{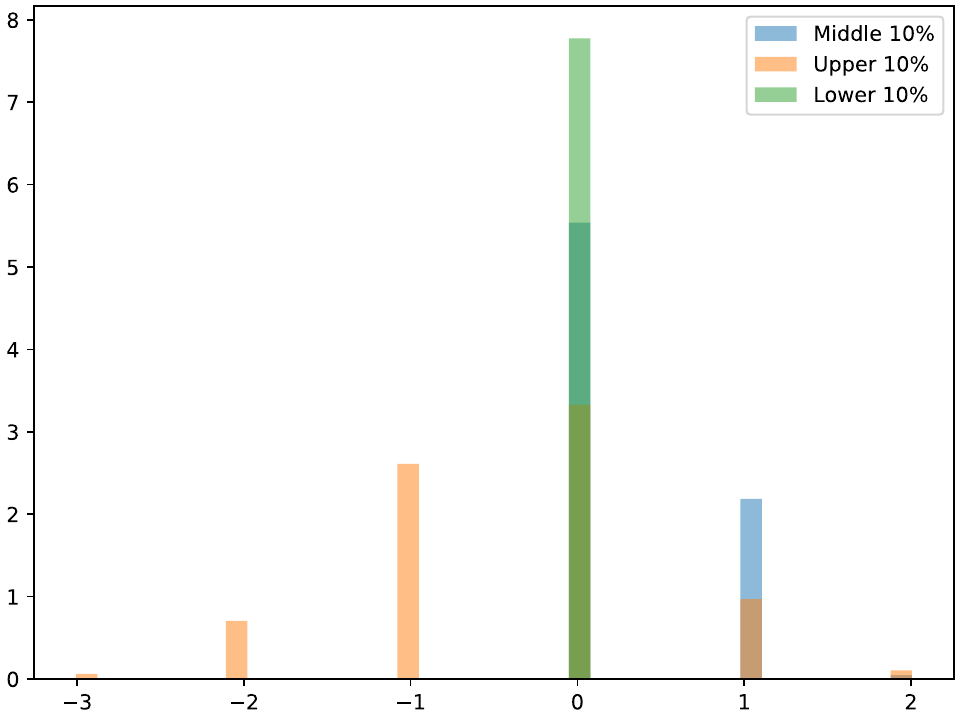}
    


    \raisebox{0.1\height}{\makebox[0.01\textwidth]{\rotatebox{90}{\makecell[c]{\scriptsize{ $X+\hat{X_{\epsilon}}-\hat{X}$} }}}}
    \includegraphics[width=0.14\textwidth]{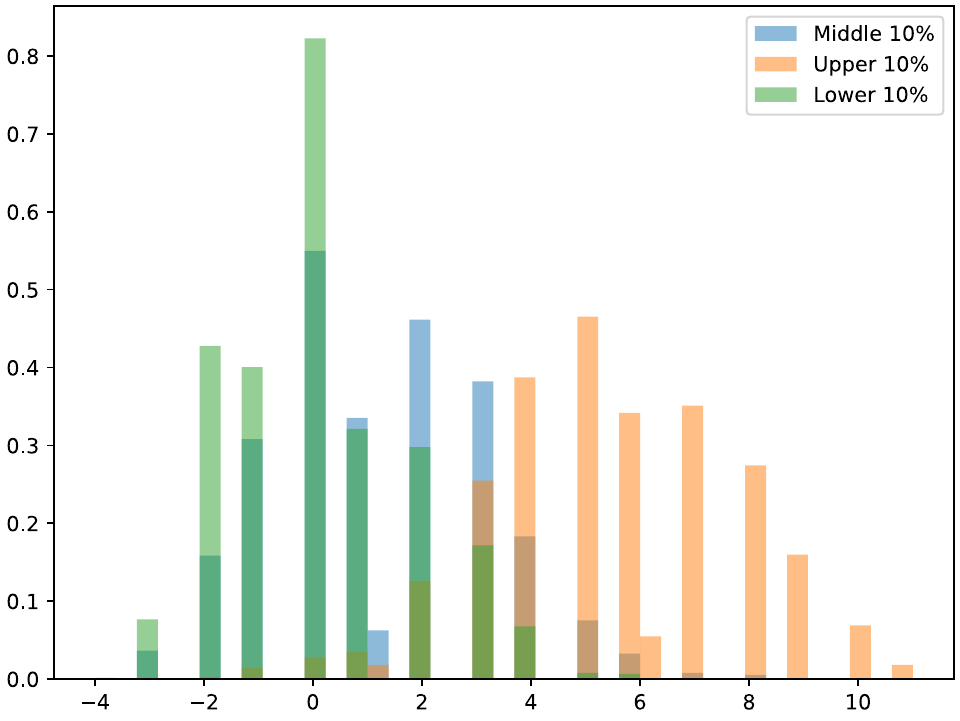}
    \includegraphics[width=0.14\textwidth]{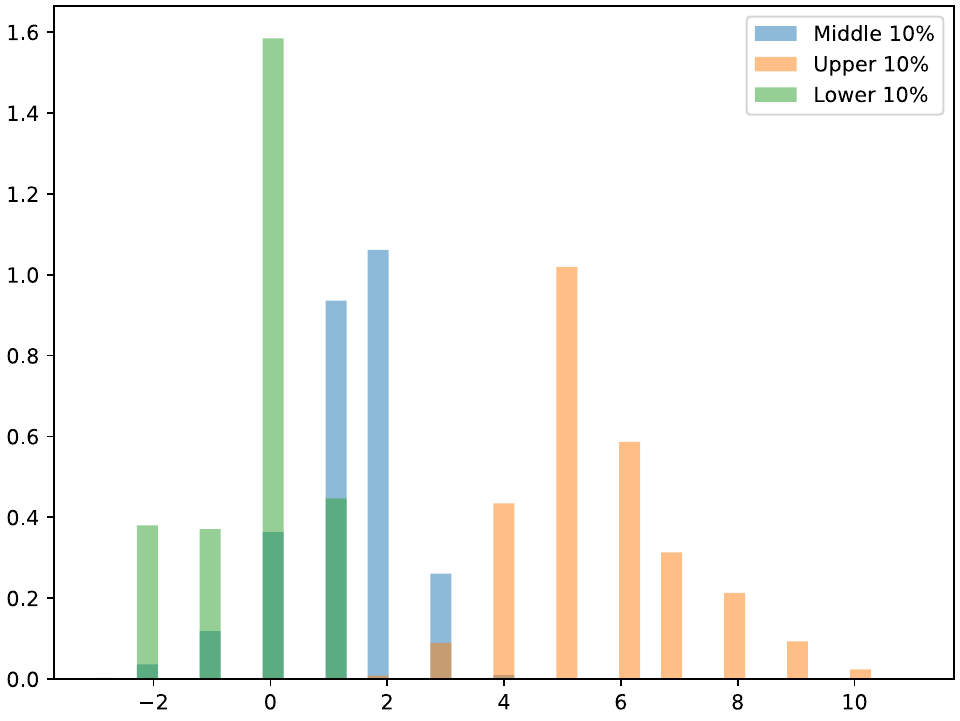}
    \includegraphics[width=0.14\textwidth]{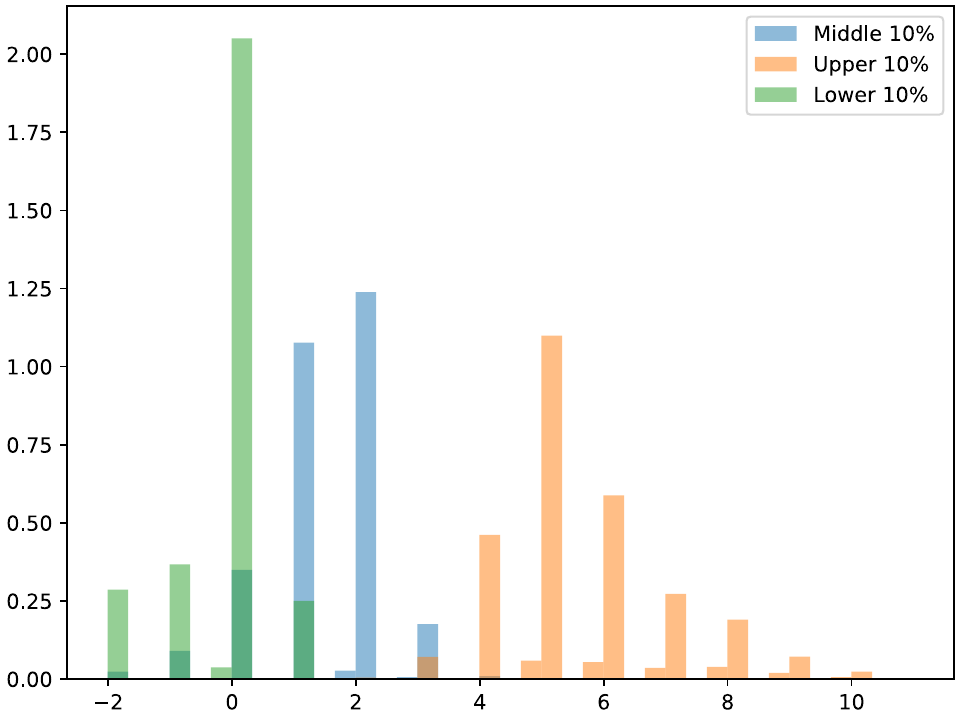}
    \includegraphics[width=0.14\textwidth]{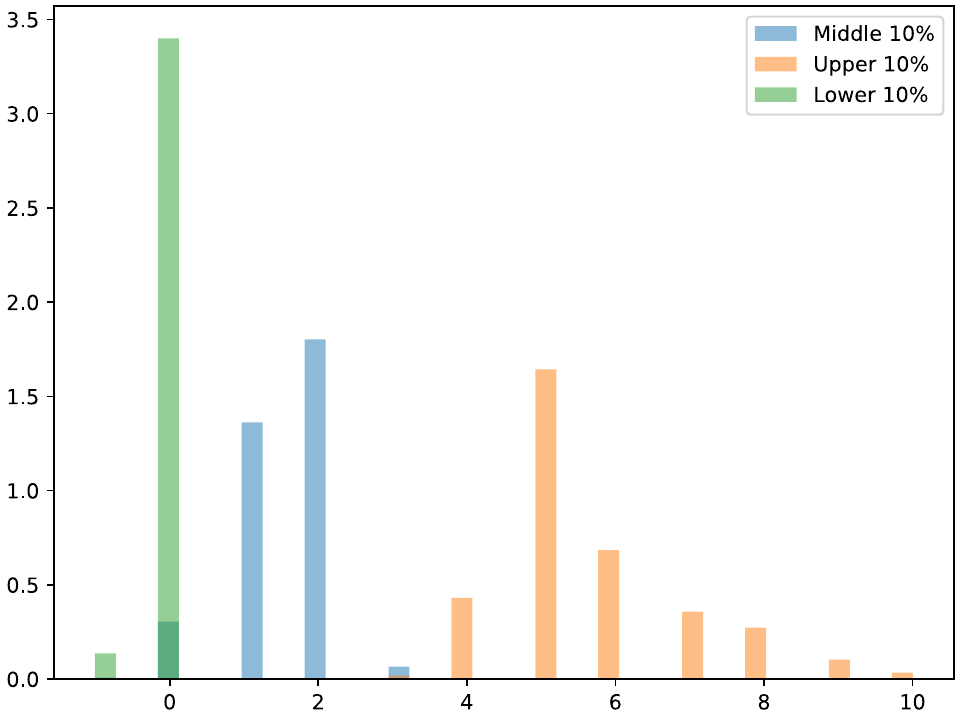}
    \includegraphics[width=0.14\textwidth]{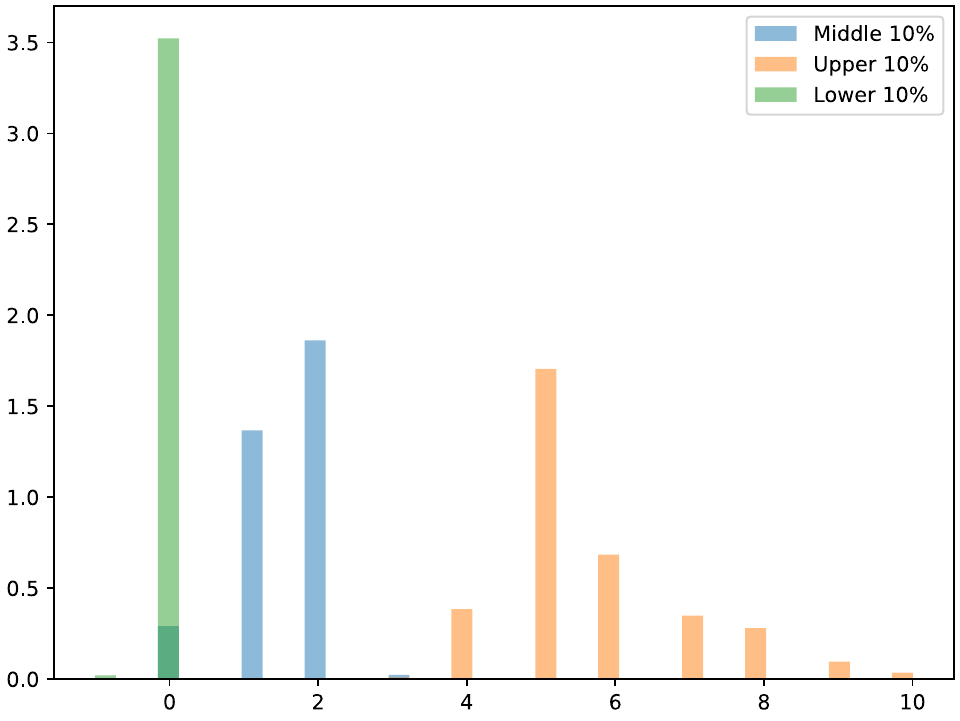}
    \includegraphics[width=0.14\textwidth]{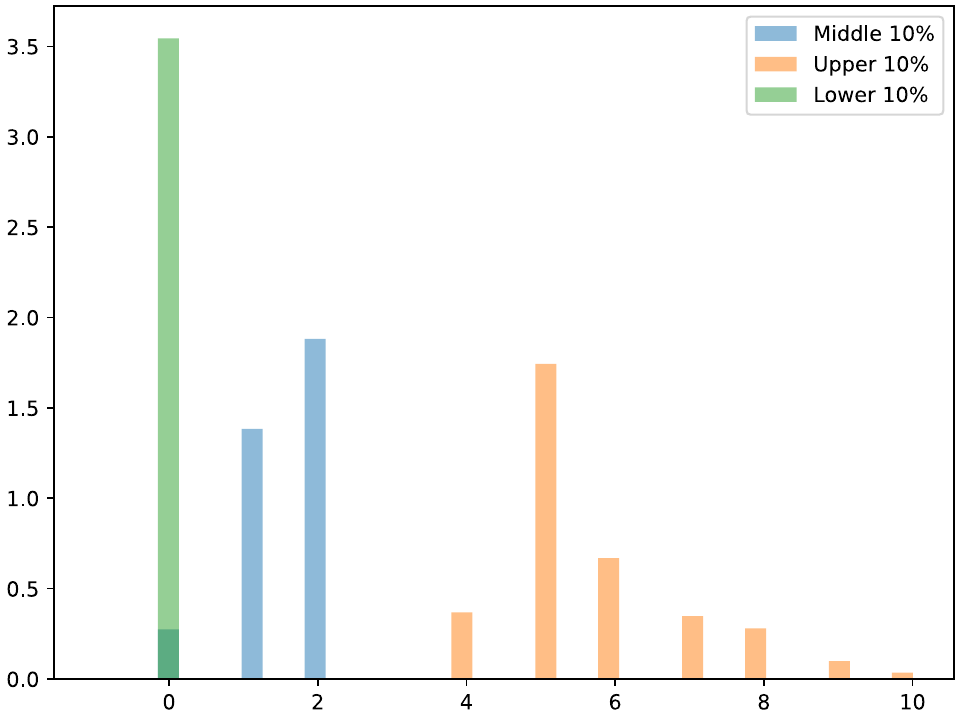}

    \raisebox{0.1\height}{\makebox[0.01\textwidth]{\rotatebox{90}{\makecell[c]{\scriptsize{ \quad$\hat{X_{\epsilon}}-\hat{X}$} }}}}
    \includegraphics[width=0.14\textwidth]{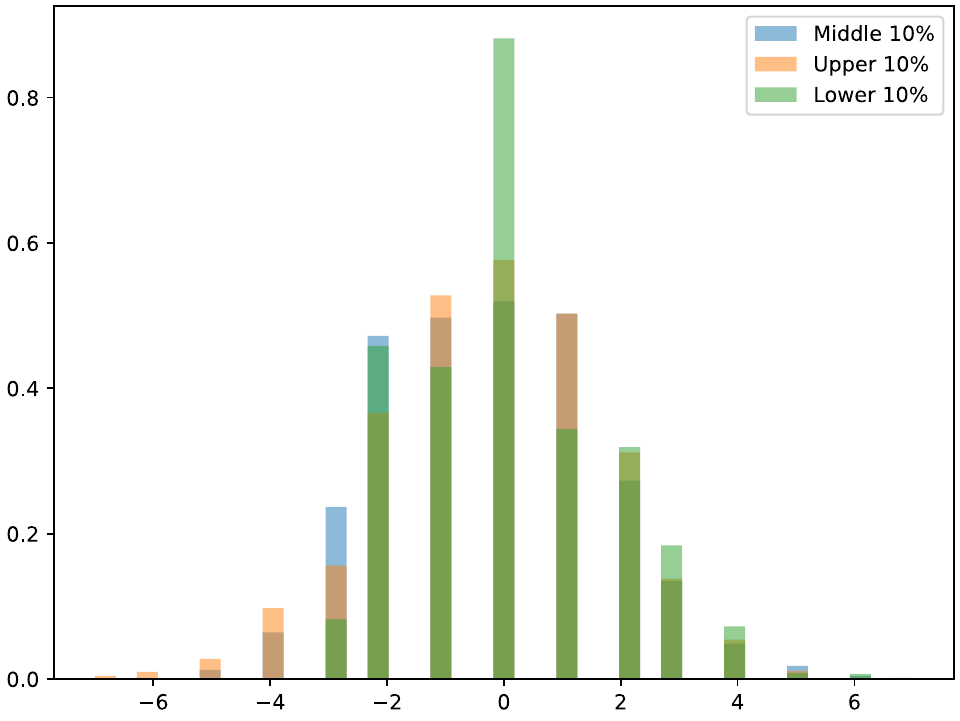}
    \includegraphics[width=0.14\textwidth]{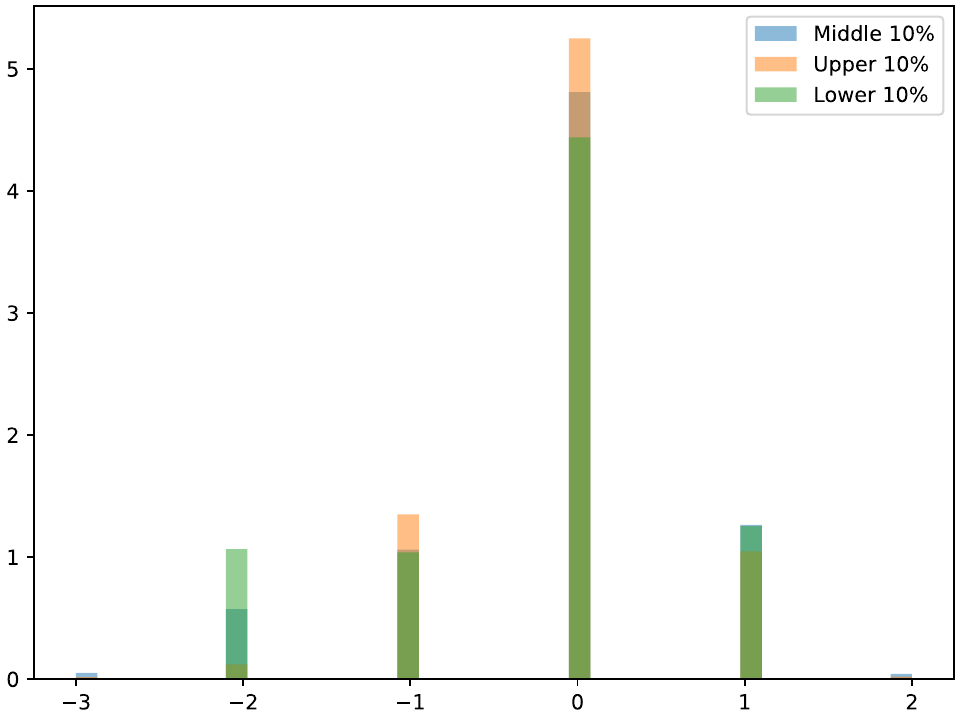}
    \includegraphics[width=0.14\textwidth]{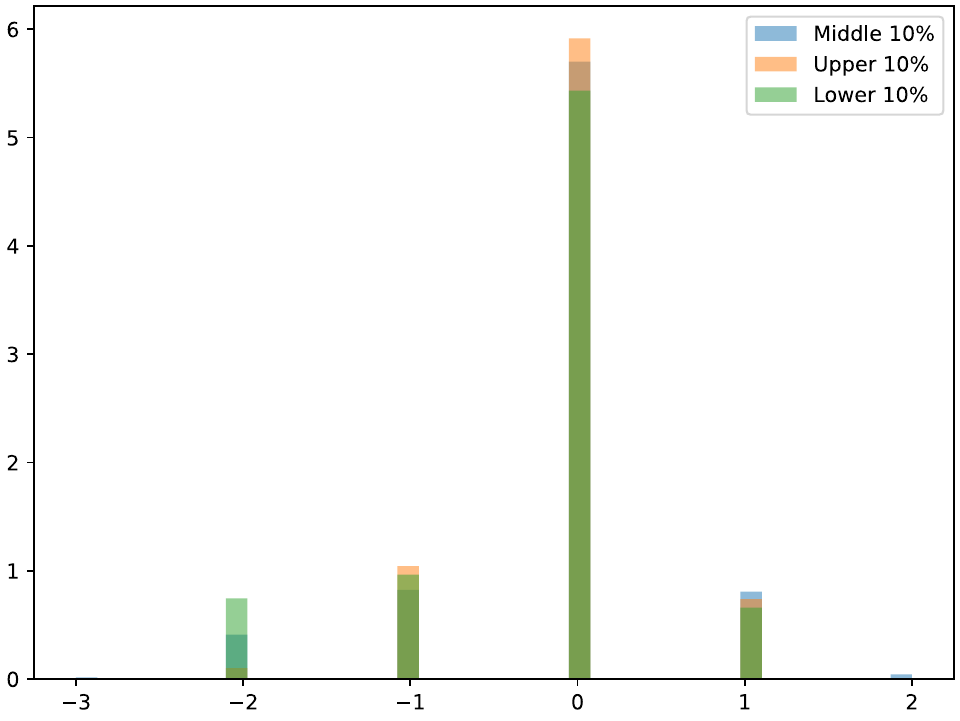}
    \includegraphics[width=0.14\textwidth]{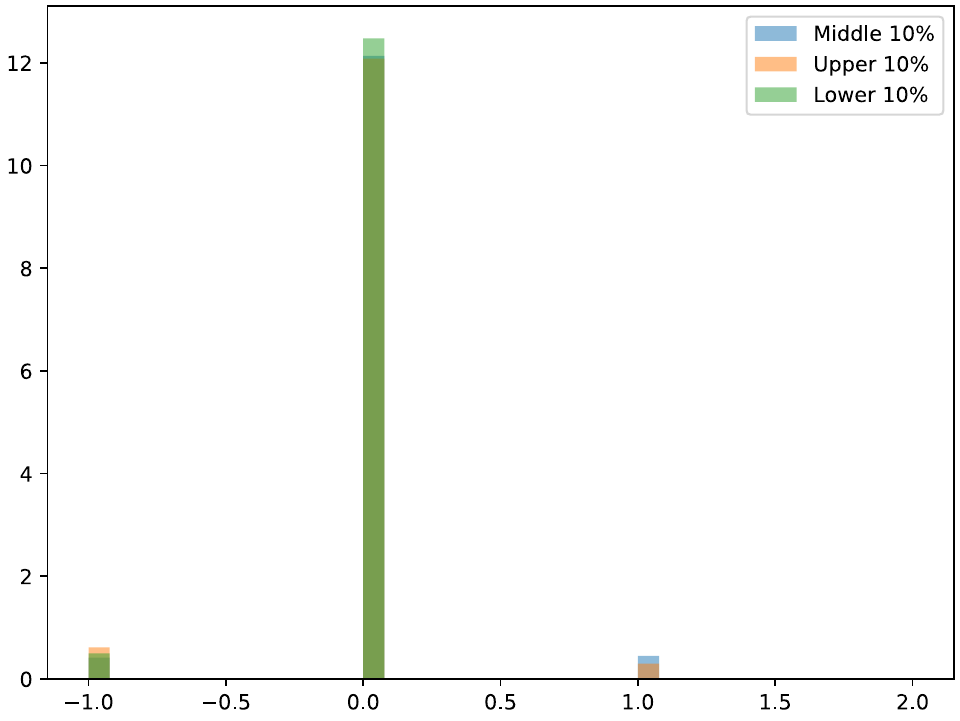}
    \includegraphics[width=0.14\textwidth]{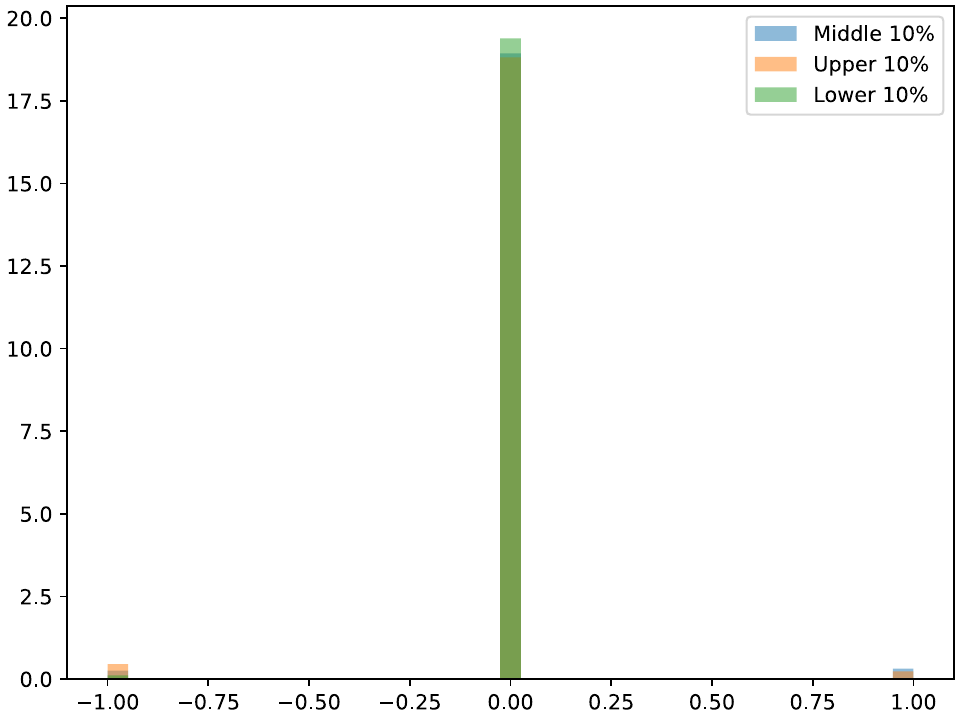}
    \includegraphics[width=0.14\textwidth]{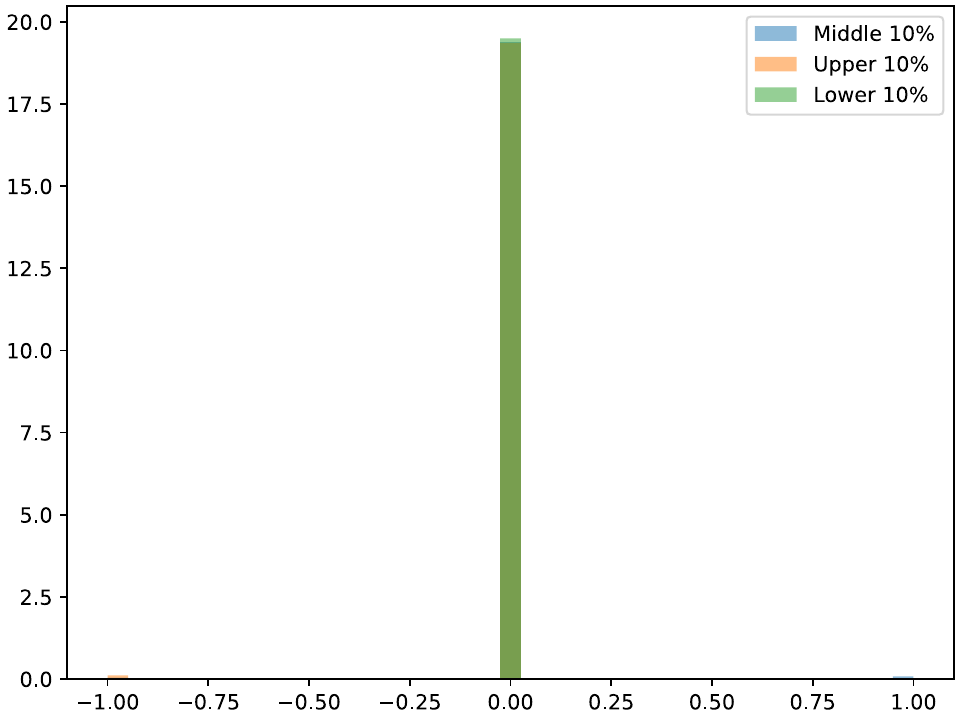}
    
    \caption{A comparison table for different strategy corresponding to the perturbation result for the \textbf{Credit Inquiry 6m} variable in different region.}  
    \label{fig:region_credit}
\end{figure*}

%% file: apd.tex
\begin{figure*}[!htbp] \centering
   \makebox[0.01\textwidth]{}
    \makebox[0.14\textwidth]{\small $\epsilon=1$}
    \makebox[0.14\textwidth]{\small $\epsilon=0.1$}
    \makebox[0.14\textwidth]{\small \quad$\epsilon=0.05$}
    \makebox[0.14\textwidth]{\small \quad$\epsilon=0.01$}
    \makebox[0.14\textwidth]{\small \quad$\epsilon=0.005$}
    \makebox[0.14\textwidth]{\small \quad$\epsilon=0.001$}
    \\
    \raisebox{0.1\height}{\makebox[0.01\textwidth]{\rotatebox{90}{\makecell[c]{\scriptsize{ \quad\quad$\hat{X_{\epsilon}}$} }}}}
    \includegraphics[width=0.14\textwidth]{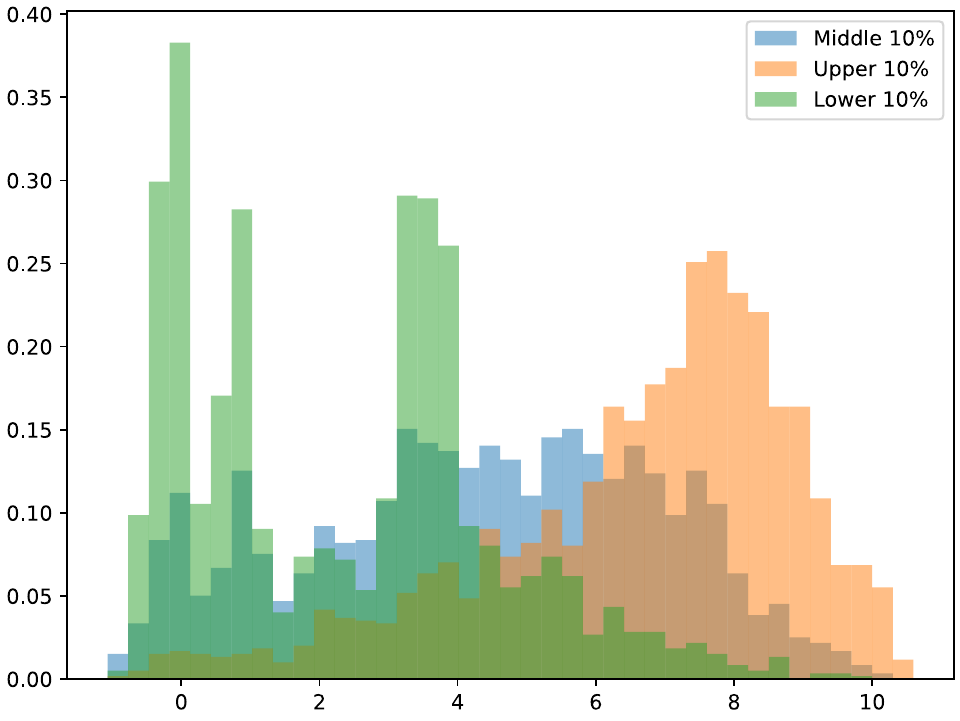}
    \includegraphics[width=0.14\textwidth]{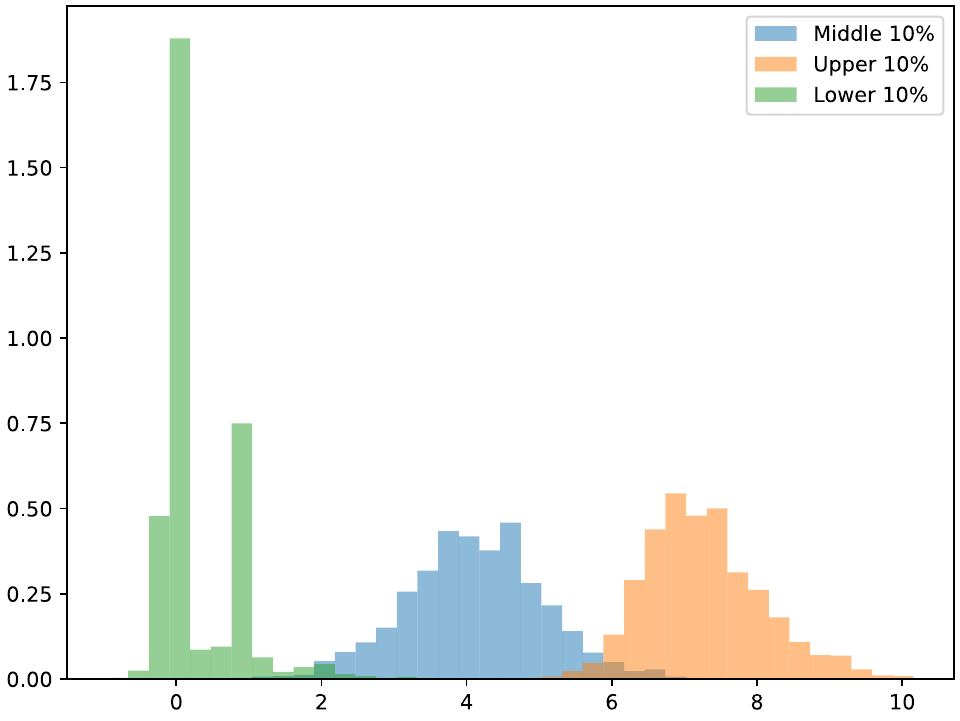}
    \includegraphics[width=0.14\textwidth]{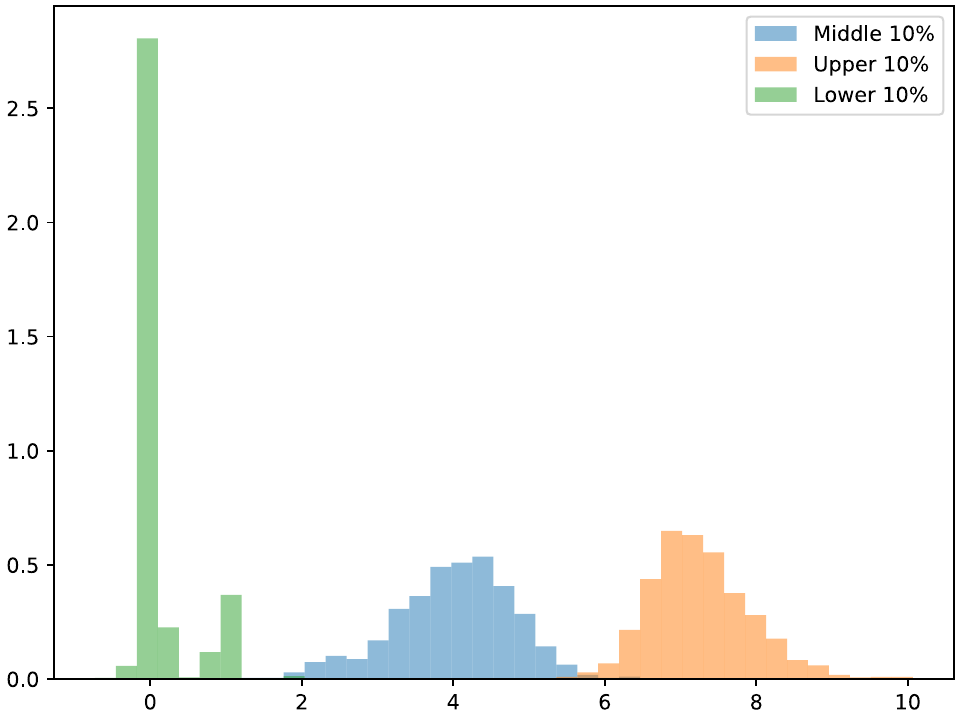}
    \includegraphics[width=0.14\textwidth]{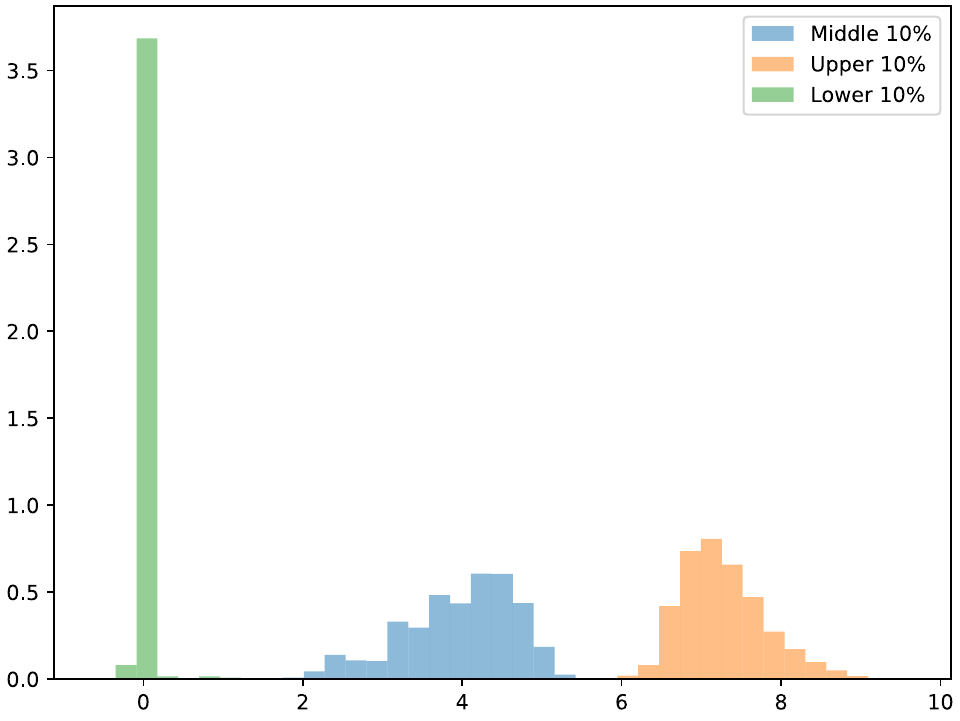}
    \includegraphics[width=0.14\textwidth]{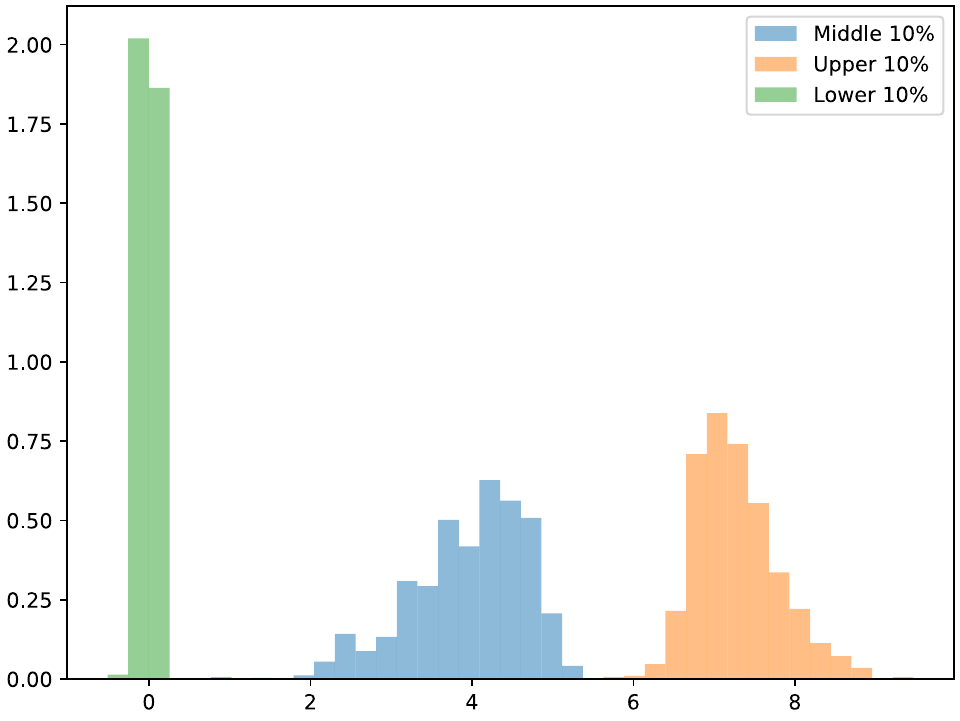}
    \includegraphics[width=0.14\textwidth]{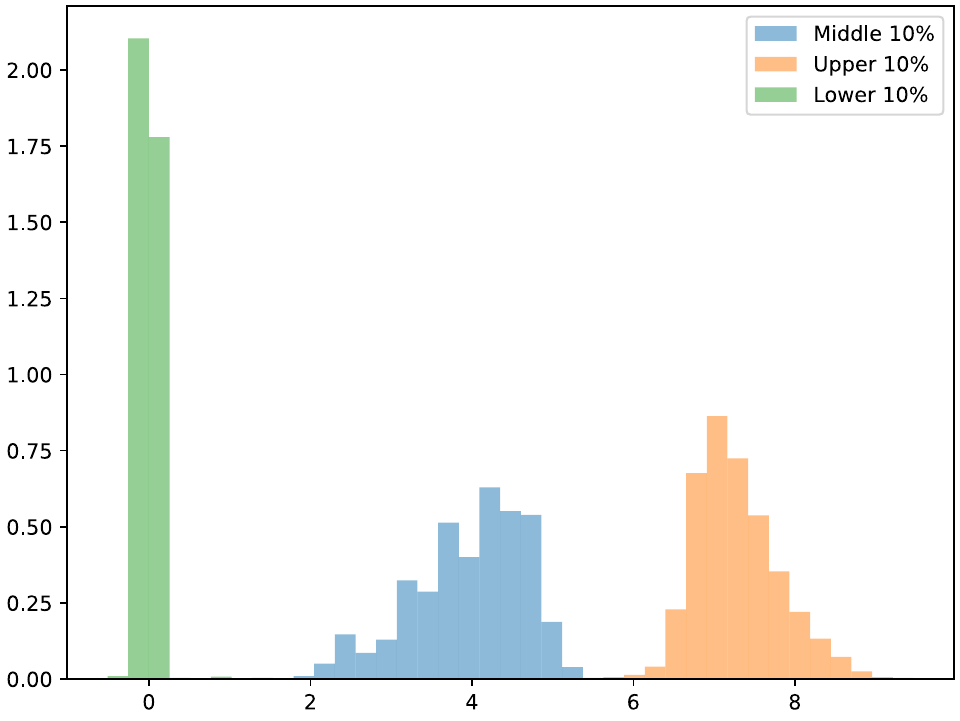}

    
    \raisebox{0.1\height}{\makebox[0.01\textwidth]{\rotatebox{90}{\makecell[c]{\scriptsize{ \quad$\hat{X_{\epsilon}}-X$} }}}}
    \includegraphics[width=0.14\textwidth]{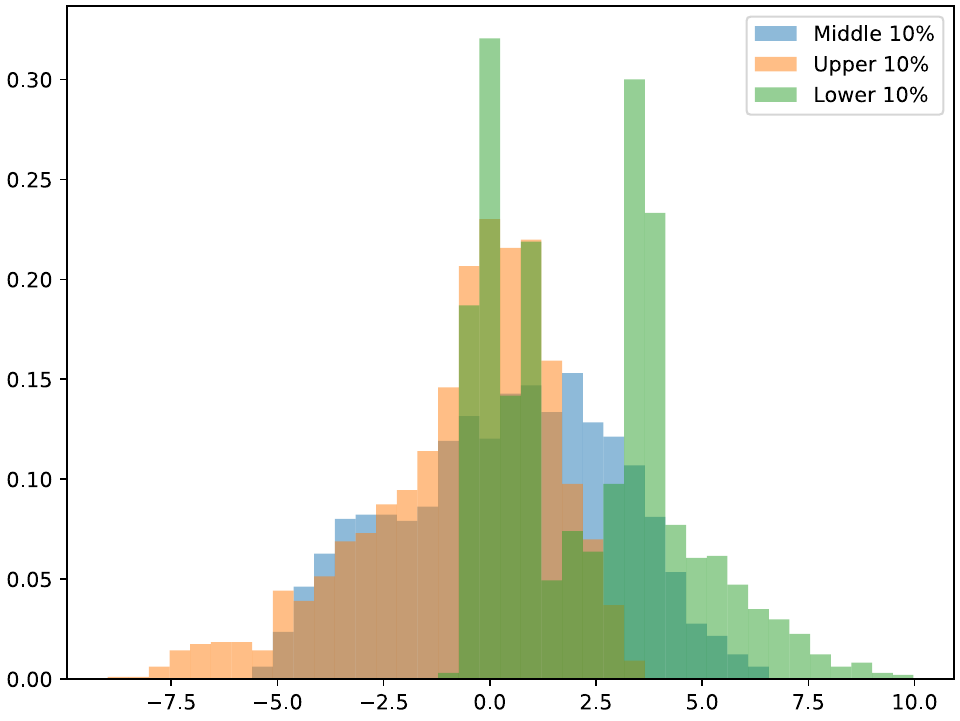}
    \includegraphics[width=0.14\textwidth]{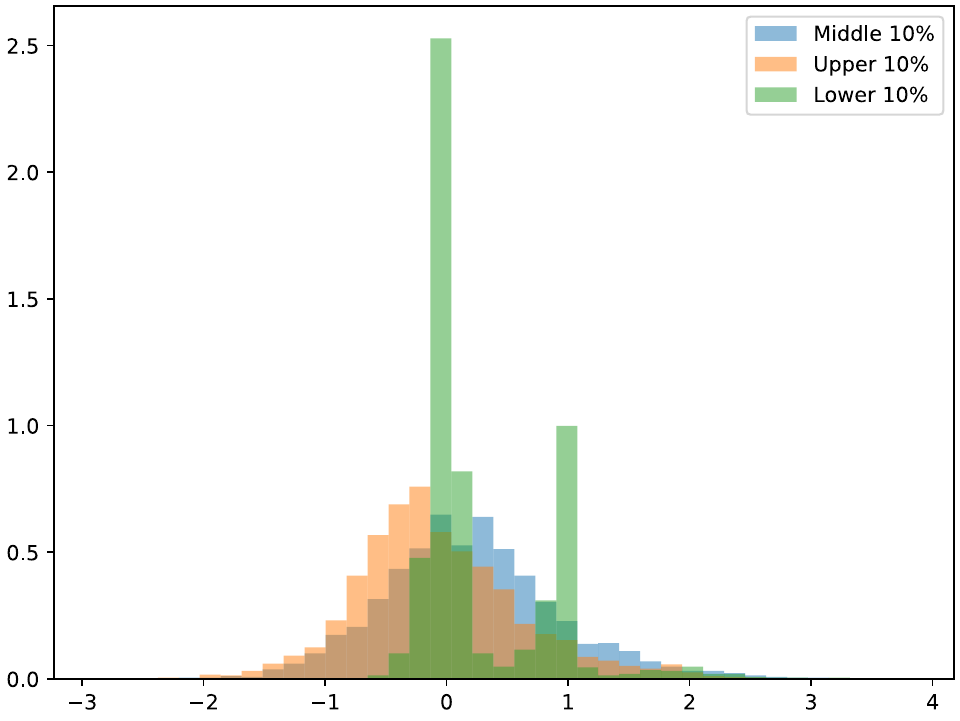}
    \includegraphics[width=0.14\textwidth]{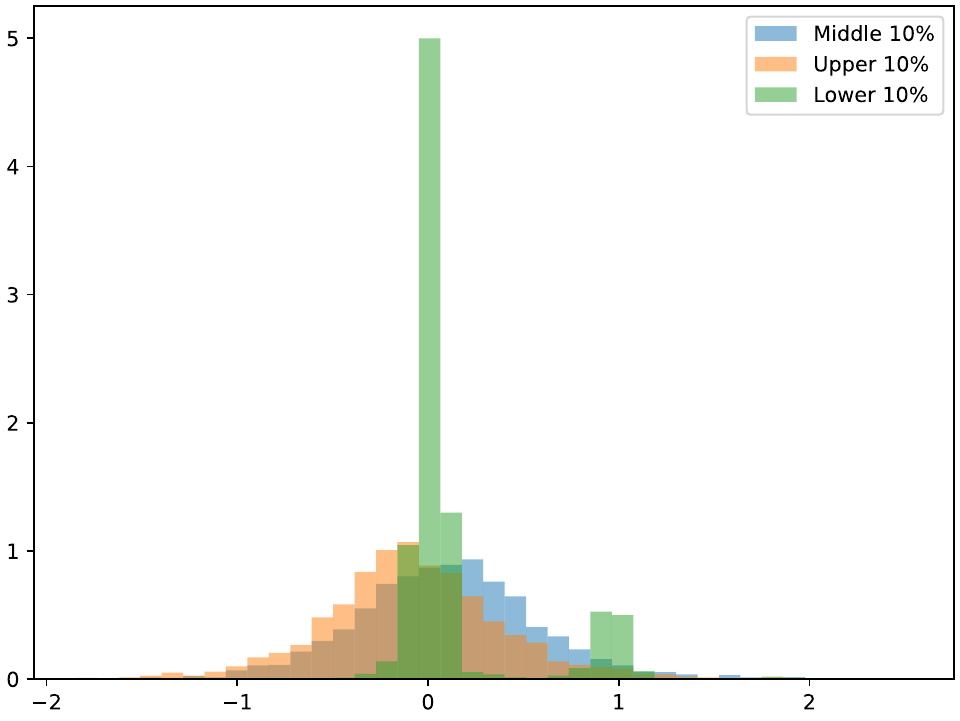}
    \includegraphics[width=0.14\textwidth]{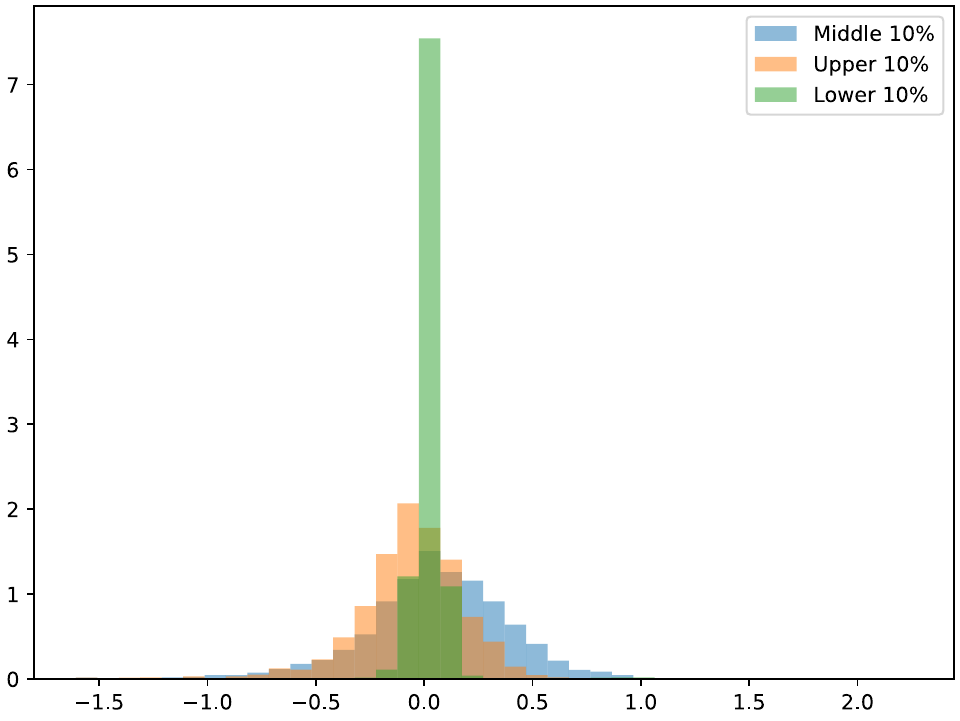}
    \includegraphics[width=0.14\textwidth]{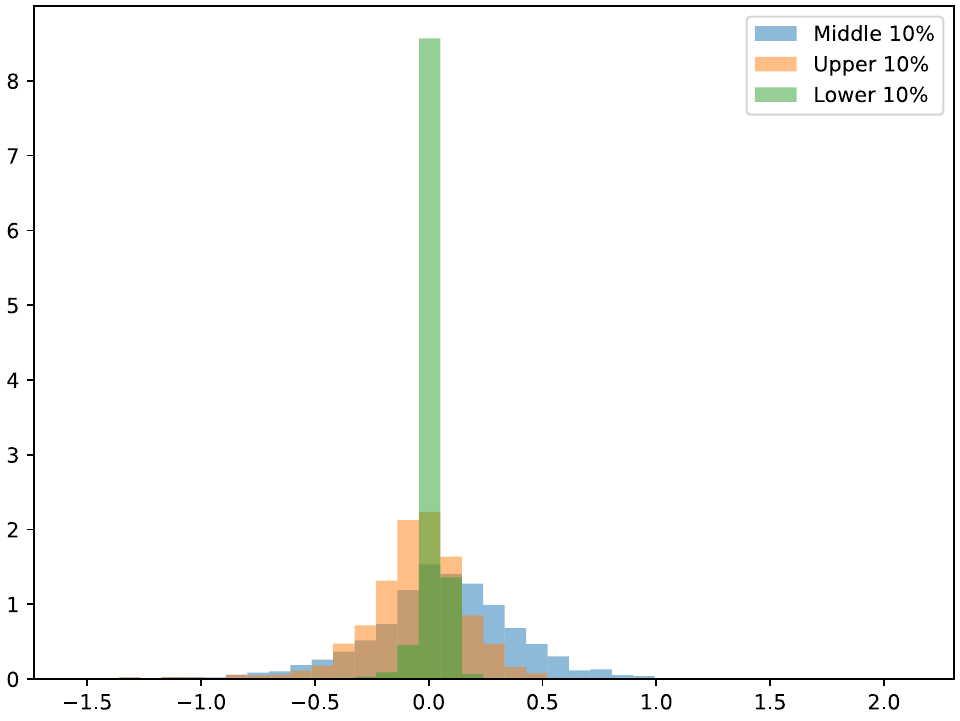}
    \includegraphics[width=0.14\textwidth]{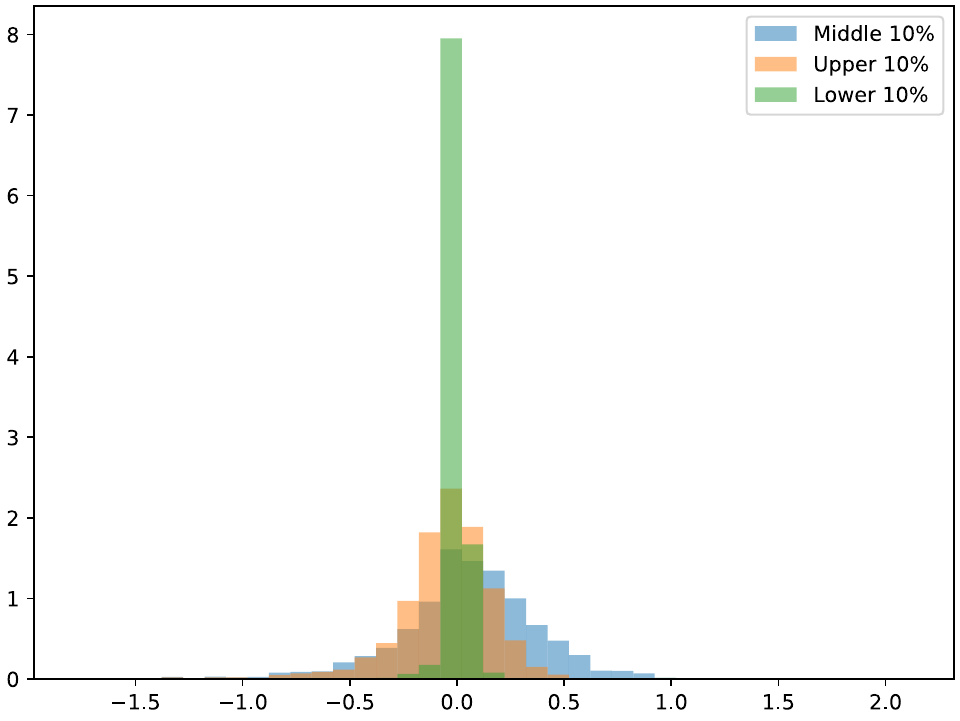}



    \raisebox{0.1\height}{\makebox[0.01\textwidth]{\rotatebox{90}{\makecell[c]{\scriptsize{ $X+\hat{X_{\epsilon}}-\hat{X}$} }}}}
    \includegraphics[width=0.14\textwidth]{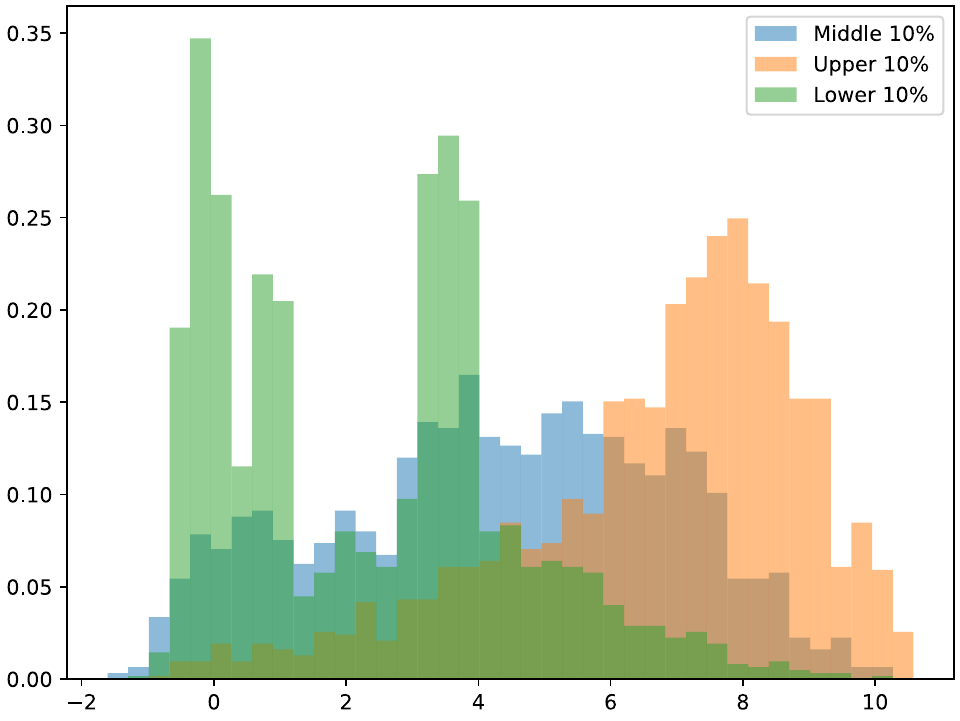}
    \includegraphics[width=0.14\textwidth]{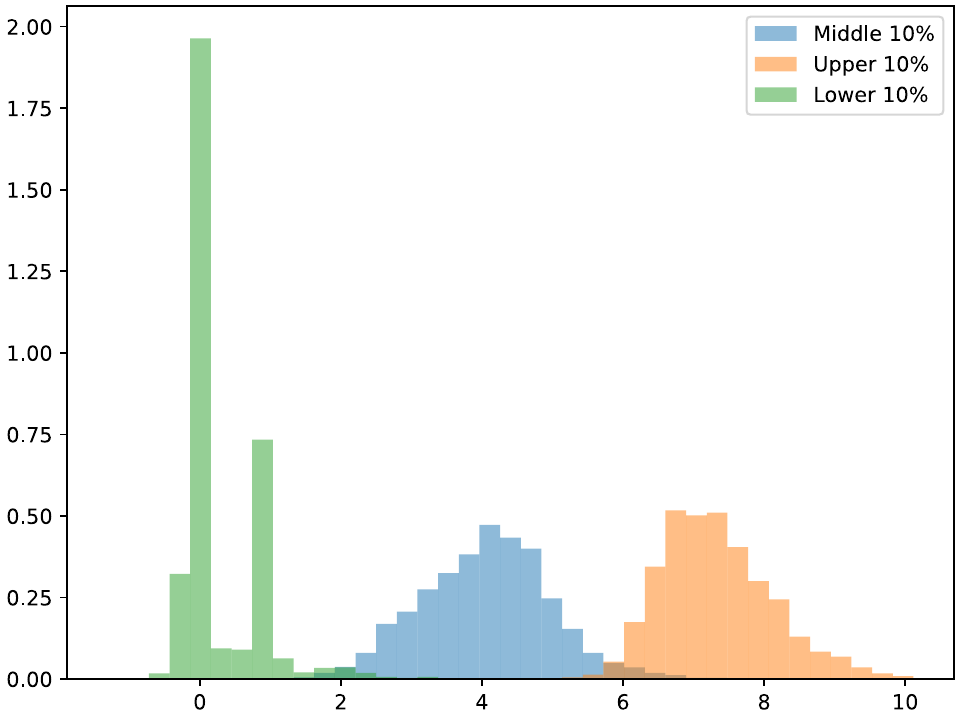}
    \includegraphics[width=0.14\textwidth]{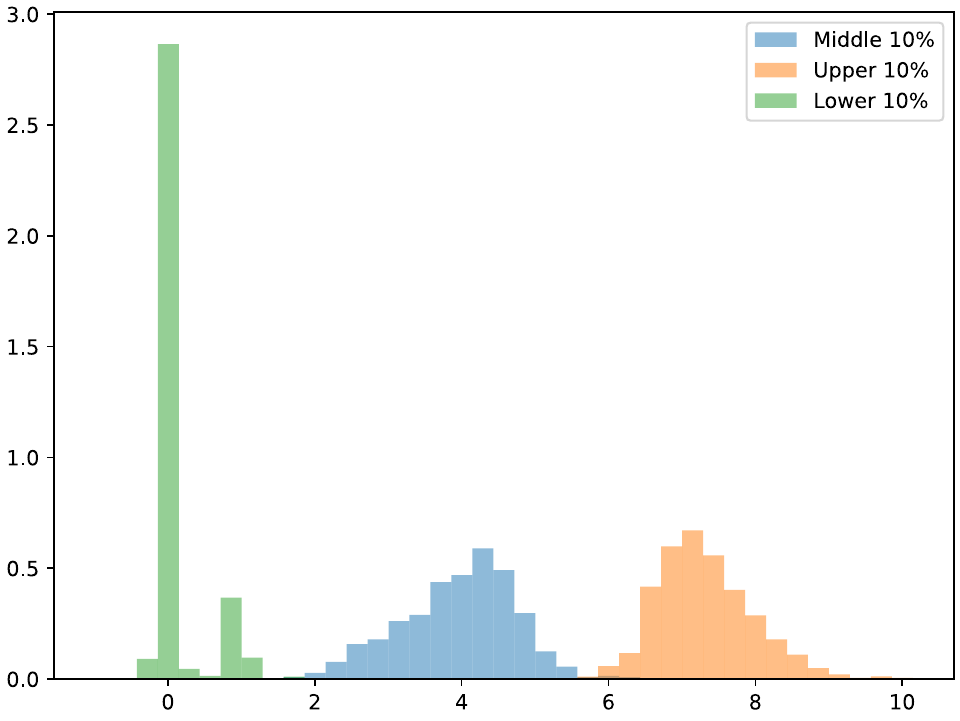}
    \includegraphics[width=0.14\textwidth]{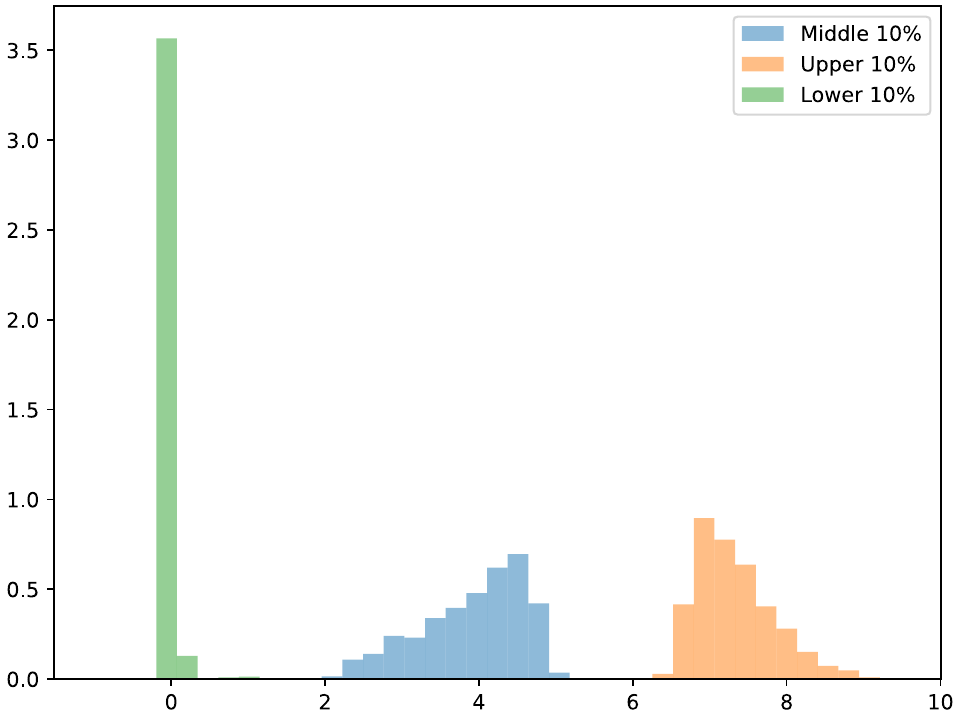}
    \includegraphics[width=0.14\textwidth]{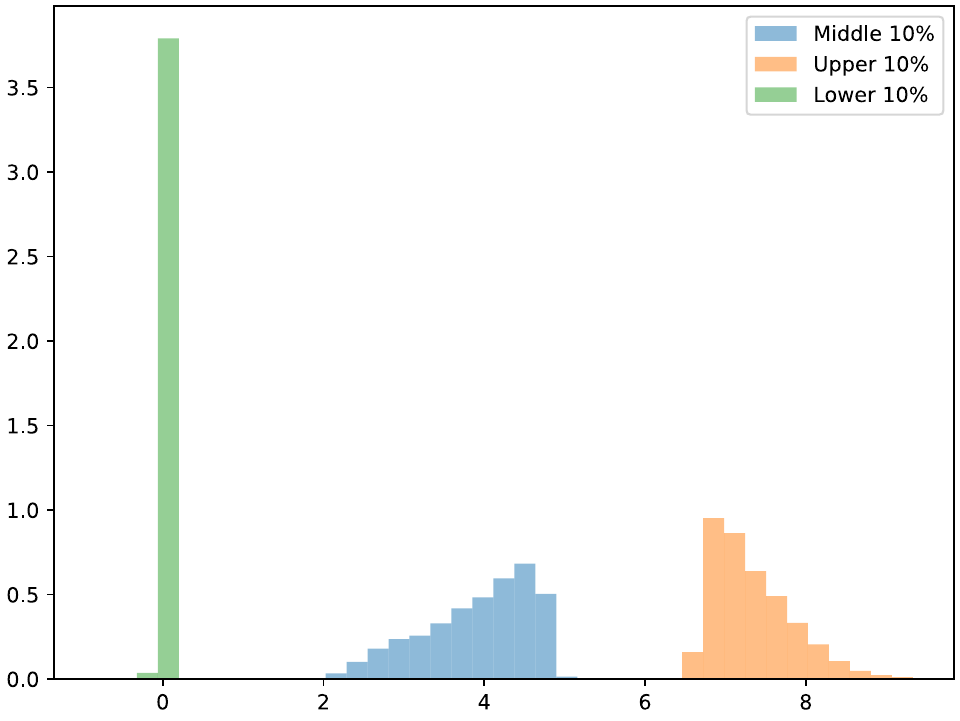}
    \includegraphics[width=0.14\textwidth]{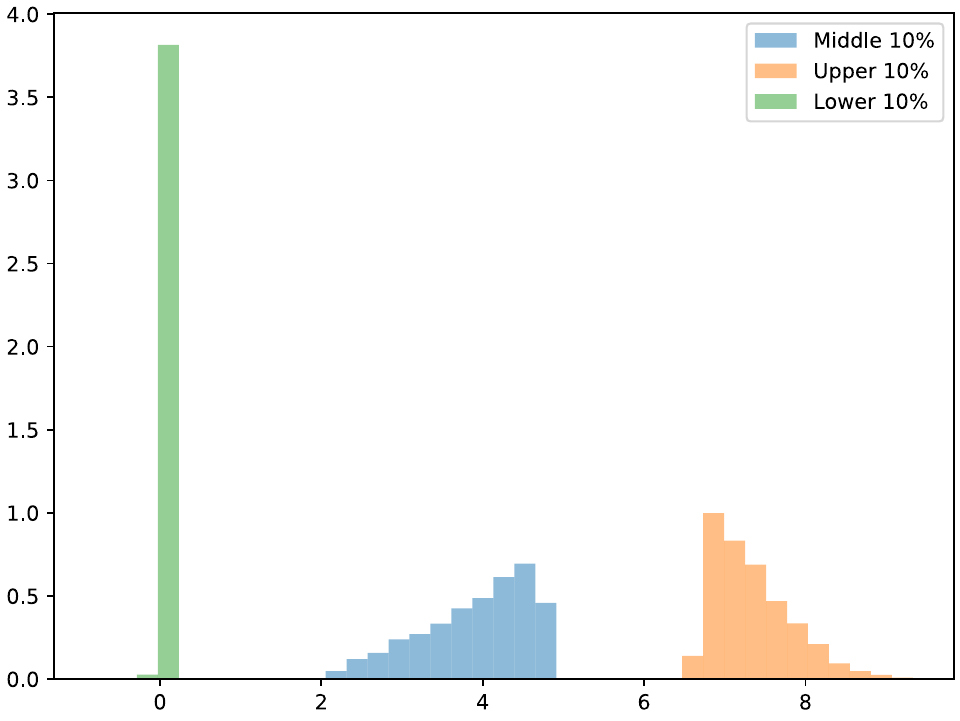}
    
    \raisebox{0.1\height}{\makebox[0.01\textwidth]{\rotatebox{90}{\makecell[c]{\scriptsize{ \quad$\hat{X_{\epsilon}}-\hat{X}$} }}}}
    \includegraphics[width=0.14\textwidth]{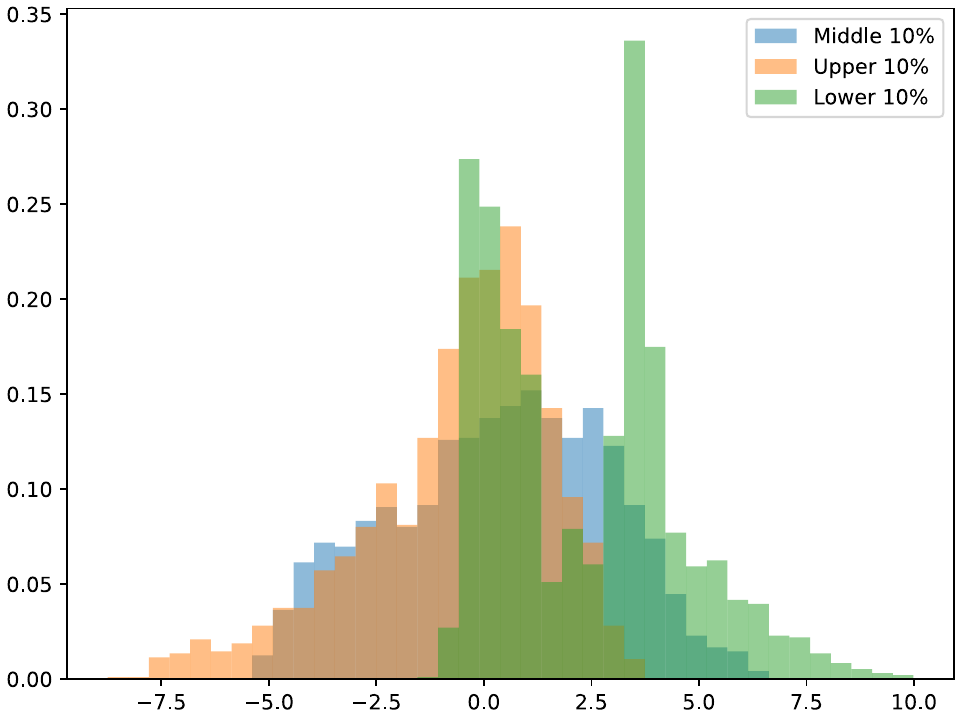}
    \includegraphics[width=0.14\textwidth]{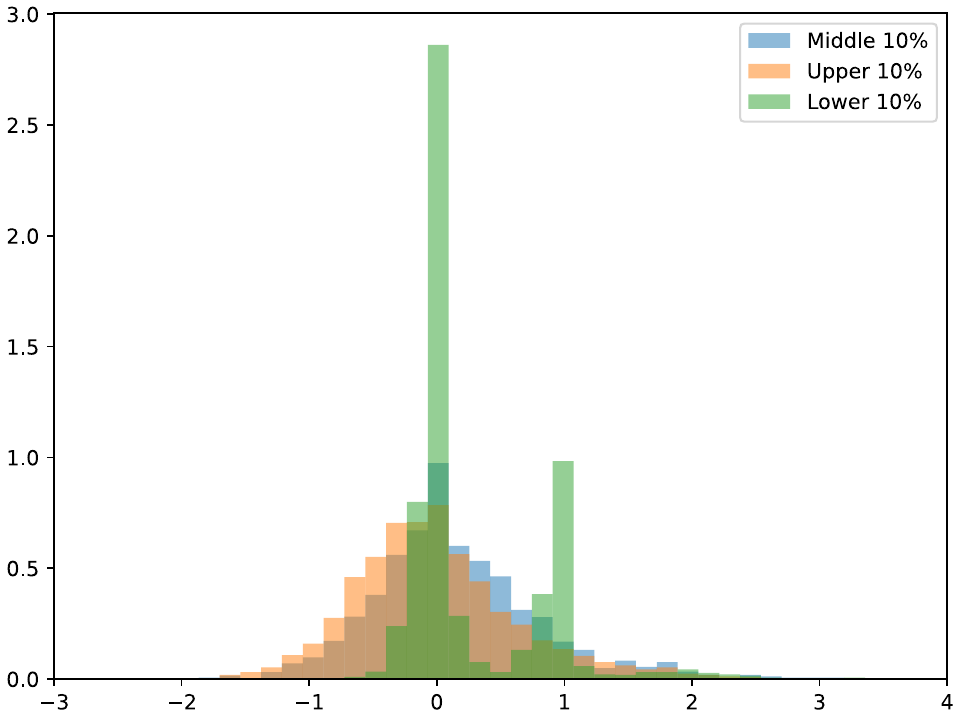}
    \includegraphics[width=0.14\textwidth]{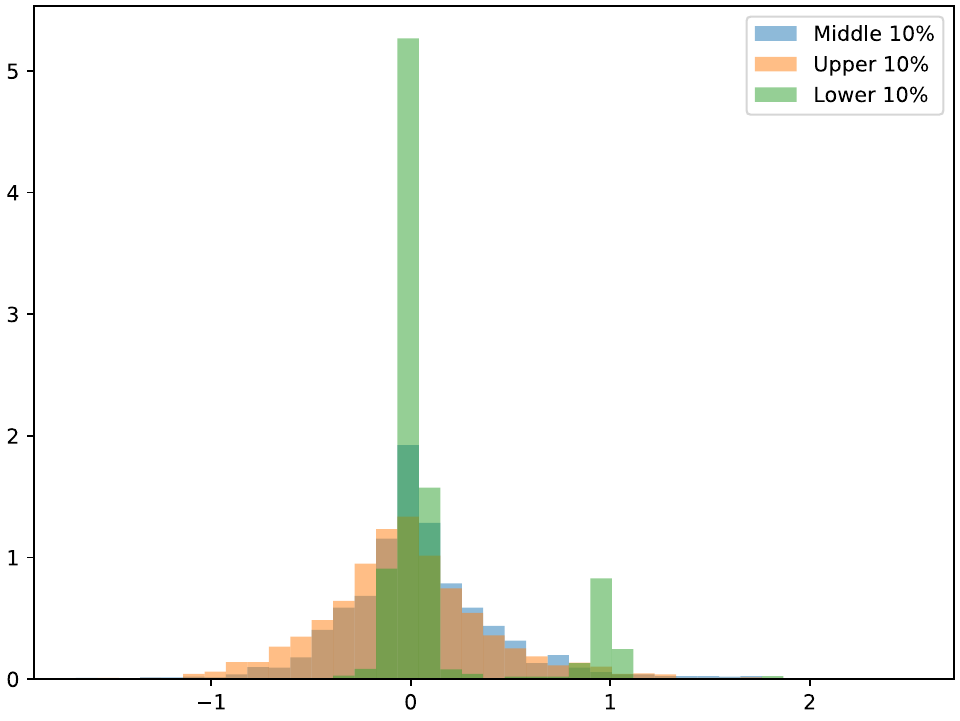}
    \includegraphics[width=0.14\textwidth]{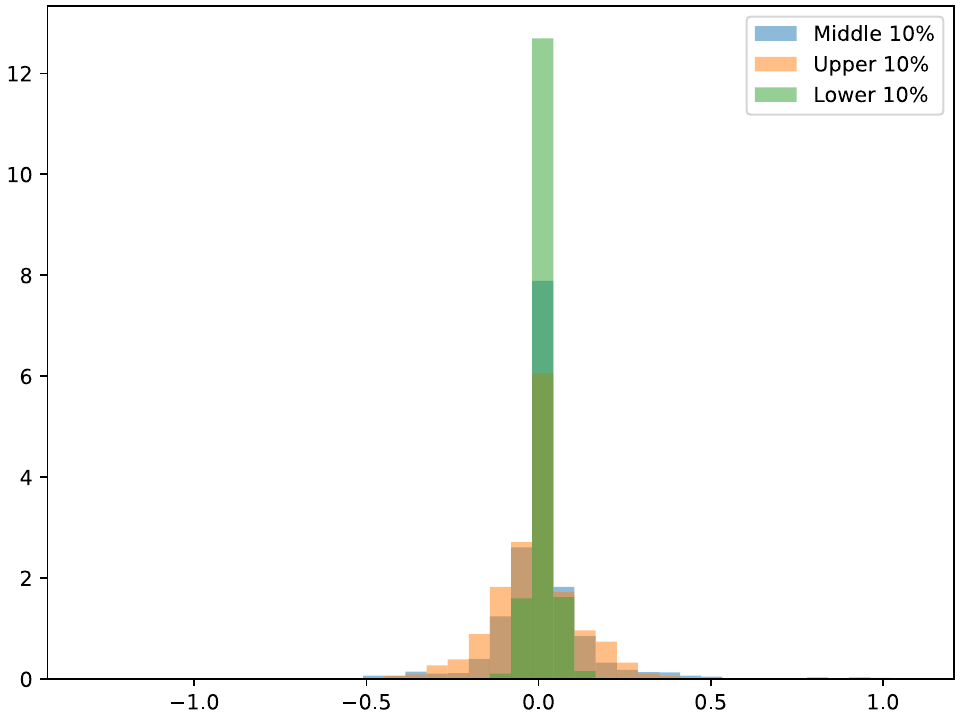}
    \includegraphics[width=0.14\textwidth]{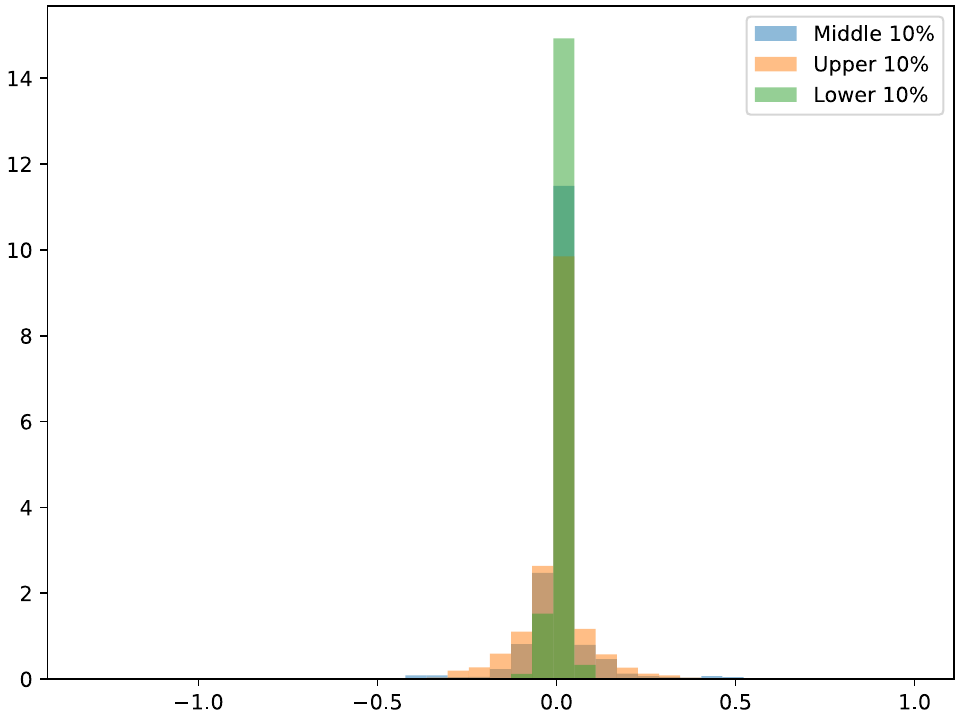}
    \includegraphics[width=0.14\textwidth]{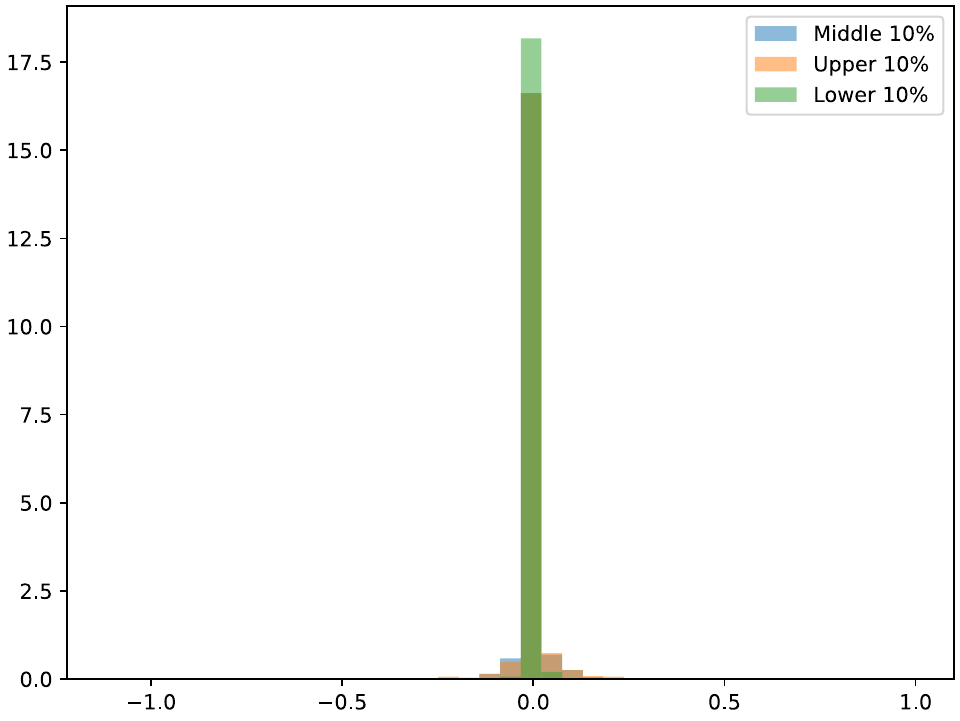}

    \caption{A comparison table for different strategy corresponding to the perturbation result for the \textbf{Amount Past Due} variable in different region.}  
    \label{fig:region_apd}
\end{figure*}

%% file: correlation_matrix.tex
\begin{figure*}[!htbp] \centering
    \makebox[0.01\textwidth]{}
    \makebox[0.3\textwidth]{}
    \makebox[0.3\textwidth]{}
    \makebox[0.3\textwidth]{}

    \raisebox{0.1\height}{\makebox[0.01\textwidth]{\rotatebox{90}{\makecell[c]{\small{ \quad\quad} }}}}
    \includegraphics[width=0.3\textwidth]{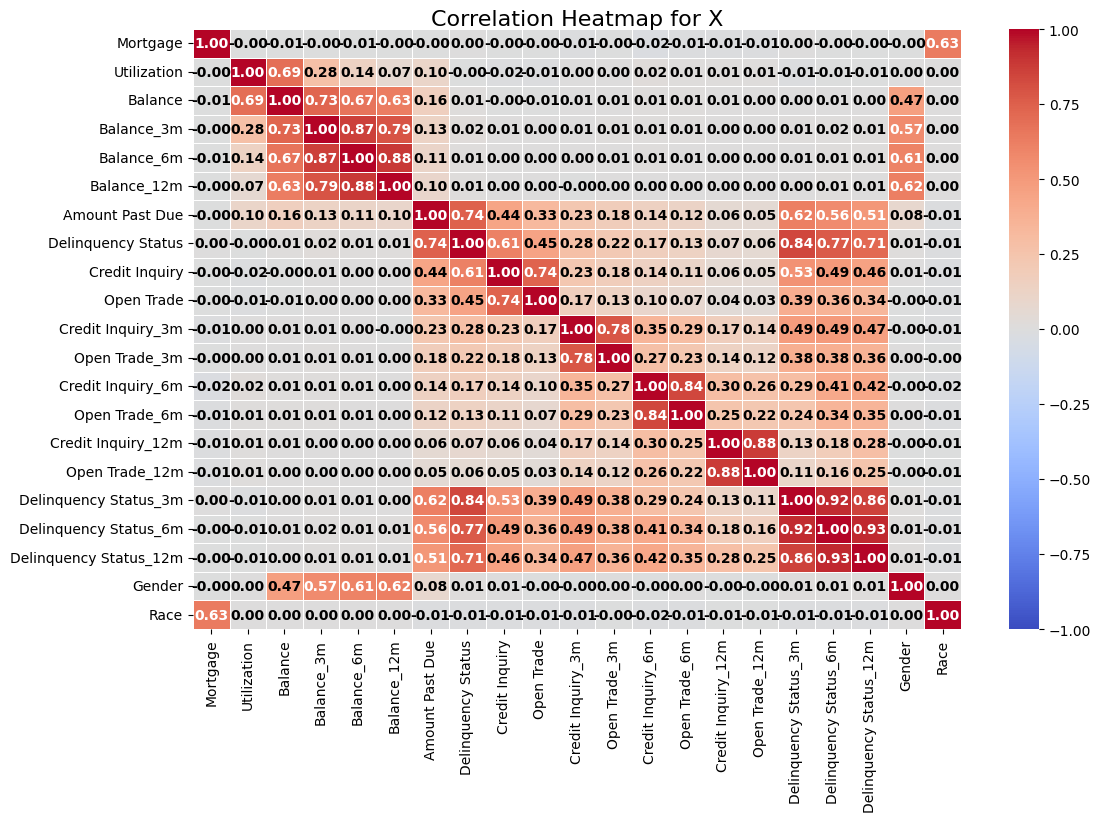}
    \hspace{0.2cm}\includegraphics[width=0.3\textwidth]{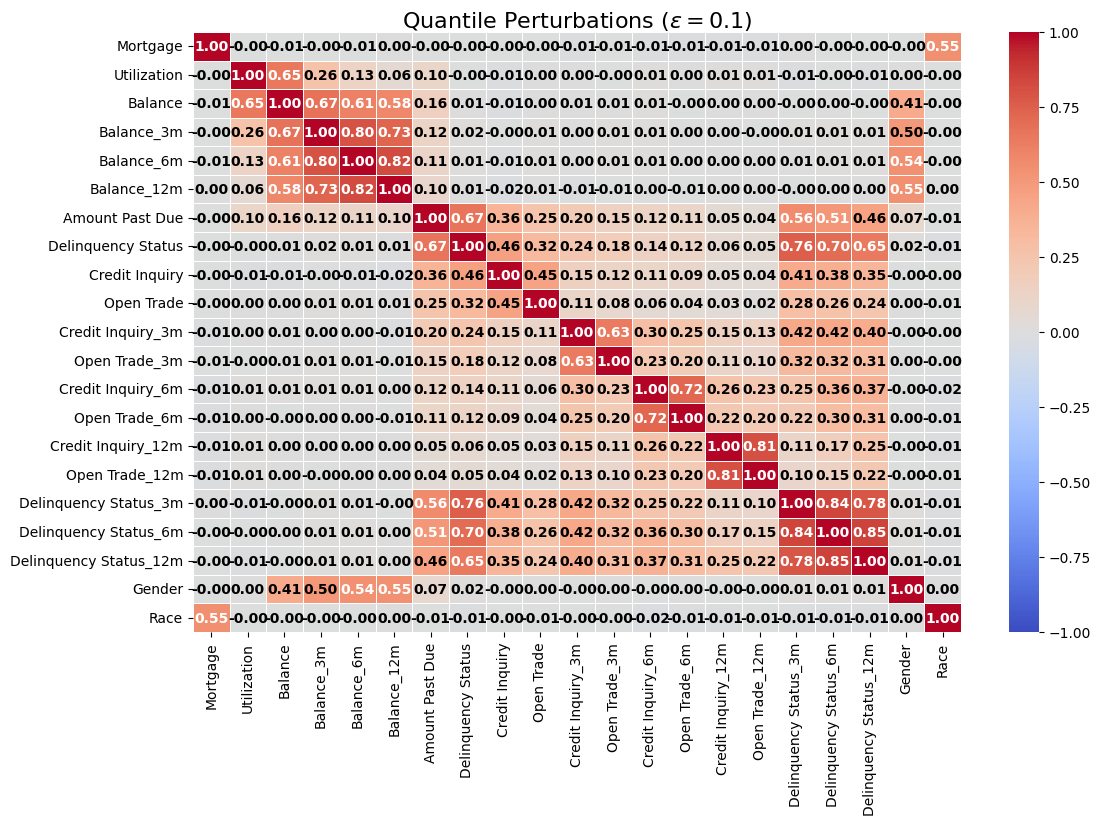}
    \hspace{0.2cm}\includegraphics[width=0.3\textwidth]{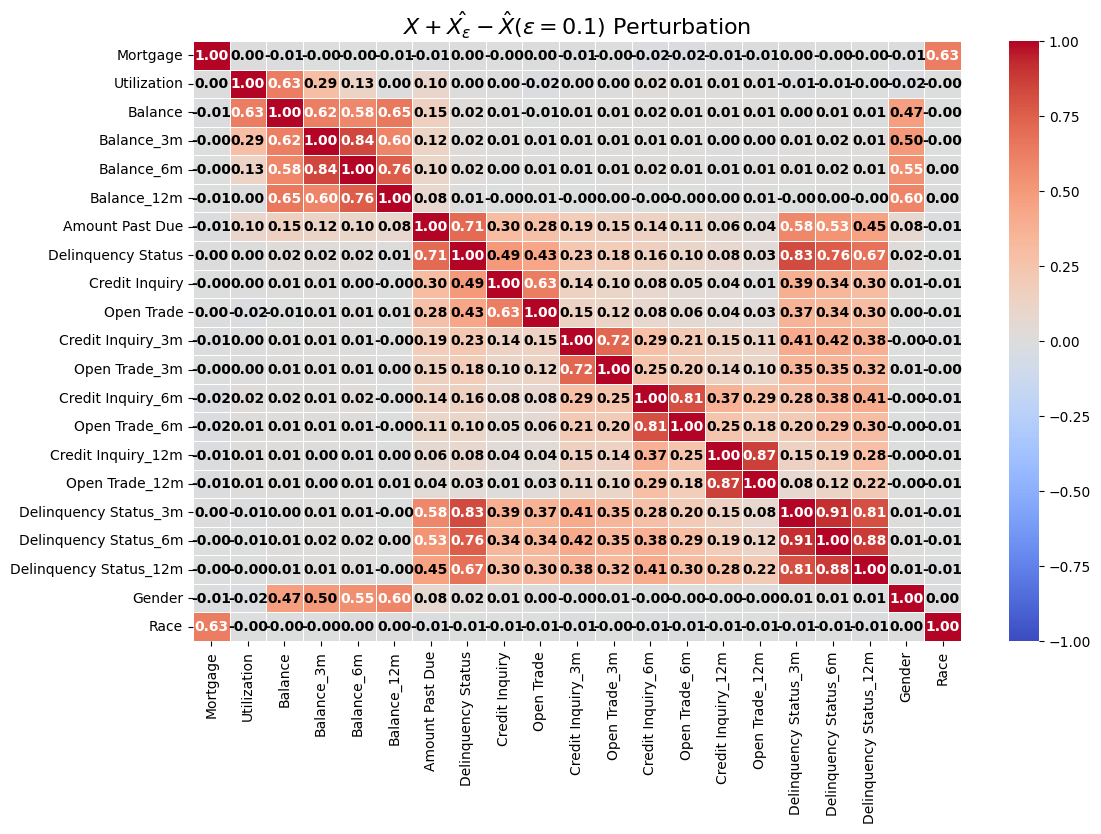}
    
    \caption{From left to right: (A) Correlation maps of ground truth $X$. (B) Correlation map of quantile perturbation $\epsilon=0.1$, (C) Correlation map of model-based perturbation $X+\hat{X_{\epsilon}}-\hat{X}$. $SSIM(A,B)=0.977019, SSIM(A,C)=0.985617$}  
    \label{fig:robust}
\end{figure*}

%% file: r_q_model_based.tex
\begin{figure*}[!htbp] \centering
    \makebox[0.01\textwidth]{}
    \makebox[0.3\textwidth]{}
    \makebox[0.3\textwidth]{}
    \makebox[0.3\textwidth]{}

    \raisebox{0.1\height}{\makebox[0.01\textwidth]{\rotatebox{90}{\makecell[c]{\small{ \quad\quad} }}}}
    \includegraphics[width=0.3\textwidth]{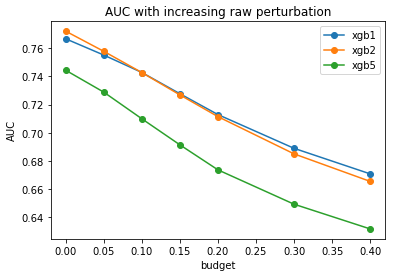}
    \hspace{0.2cm}\includegraphics[width=0.3\textwidth]{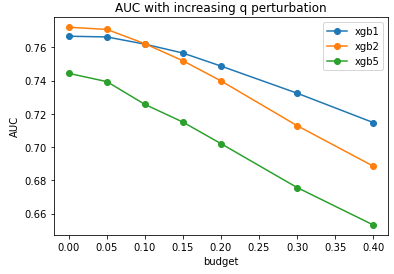}
    \hspace{0.2cm}\includegraphics[width=0.3\textwidth]{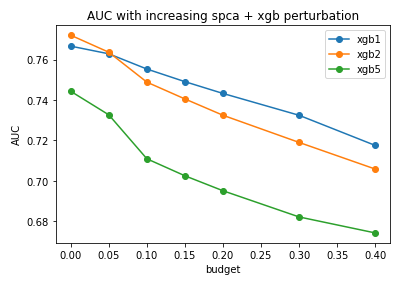}
    
    \caption{The robustness testing results for Raw, Quantile and model based approach for the simulated models.}  
    \label{fig:robust}
\end{figure*}

%% file: appendix.tex
\section{Appendix}
\subsection{\label{ICAs} ICA and related results}
\noindent ICA can unmix certain linear mixing of random variables. We review the contrast between PCA and ICA. From a manifold optimization point of view, PCA maximizes the second order cumulants while ICA maximizes the fourth. That is,
\begin{equation}
\begin{split}
    \text{ICA (Kurtosis)}&: w_{opt}=\text{argmax}_w|\kappa_4(w^TA)|, \|w\|_2=1\\ \text{PCA (Variance)}&: w_{opt}=\text{argmax}_w|\kappa_2(w^TA)|, \|w\|_2=1
\end{split}
\end{equation}
PCA can be done via SVD while ICA is done via manifold optimization which is much more expensive. The main problem for ICA is that it fails on unmixing Gaussian signals. An important class of functional transformation that captures the original random variable characteristic is \textit{moments}. We denote the $k$th moments and the $k$th central moments by
\begin{equation}
\begin{split}
     \mu_n'&=\mathbb{E}[X^n] = \int x^n f_X(x)dx, \\ \mu_n &= \mathbb{E}[(X-\mathbb{E}[X])^n]=\int(x-\mu_1')^n f_X(x)dx
\label{moments}
\end{split}
\end{equation}
Another important class is \textit{cumulants}, which are generated from the cumulant generating function (log of moment generating function)
\begin{equation}
\begin{split}
     K(t) = \text{log}\mathbb{E}[e^{tX}] = \text{log} M_X(t)=\sum_{n=1}^{\infty}\kappa_n t^n
\end{split}
\label{cumulant}
\end{equation}
where $\kappa_n=\frac{d^n}{dt^n}K(t)\lvert_{t=0}$, $\mu_n'=\frac{d^n}{dt^n}M_X(t)\lvert_{t=0}$
Using Equation \ref{moments},\ref{cumulant}, the cumulants can be expressed using \textit{moments}. The first several are
\begin{equation}
\begin{split}
    \kappa_1 &= \mu_1',\kappa_2=\mu_2,\kappa_3=\mu_3,\kappa_4=\mu_4-3\mu_2^2,\\
    \kappa_5 &= \mu_5-10\mu_3\mu_2,\kappa_6=\mu_6-15\mu_4\mu_2-10\mu_3^2+30\mu_2^3
\end{split}
\end{equation}
The cumulants satisfy: (A) Sum of cumulant is the cumulant sum given $X_1,...,X_N$ are independent. (B) Homogeneity to degree $n$
\begin{equation}
\begin{split}
    (A)\kappa_n(X_1+...+X_n)&=\kappa_n(X_1)+...+\kappa_n(X_n)\\
    (B) \kappa_n(cX)&=c^n\kappa_n (X)
\end{split}
\end{equation}

\noindent Since the moment generating function for Gaussian is known to be $M_X(t)=\text{exp}(\mu t+\frac{1}{2}\sigma^2t^2)$, this means that higher cumulants are precisely zero $\kappa_n=0,n\geq 3$. \textit{Thus, ICA can linearly unmix random variables if there are no more than 1 Gaussian variables}. 
\\
\textbf{PCA failure modes} The $\kappa_2$ manifold maximization fails in orthogonal mixing case: $y=Qx$ of zero mean, independent white noise $x$ with $\mathbb{E}[xx^H]=I$. 
Since $\kappa_2 (w^T y)=\kappa_2 (w^T Qx)=\kappa_2 (\tilde{w}^T x)=1$, any $w$ on the circle $\|w\|_2=1$ is a solution. The kurtosis maximization (ICA) partially mitigates this issue as $\kappa_4 (\tilde{w}^T x)$ 
will select a solution $\tilde{w}_{opt}=\pm e_i$ with $i=\text{argmax}_j \lvert \kappa_4(x_j)\rvert$, $x=(x_1,x_2,...,x_n),e_i^Tx=x_j$
if $i$ is 
unique. Thus, $w_{opt}=Q\tilde{w}_{opt}=\pm Q[:,i]$
.\\
In general, ICA can be interpreted as the inverse process for Central Limit theorem by unmixing random variable and discern the direction that maximize non-Gaussianity. 

\subsection{\label{Dimrev}Selected Reviews On Dimension Reduction}
\noindent We present some methodological alternatives of PCA:
\noindent \textbf{Classical MDS} The canonical objective function for classical multidimensional scaling (MDS) is $\min_Z f(Z)$, where
\begin{equation}
    f(Z) = \sum_{i=1}^n\sum_{j=i+1}^n \Big(\underbrace{\|z_i-z_j\|}_{\substack{\text{distance in} \\\text{latent dimension}}}-\underbrace{\|x_i-x_j\|}_{\substack{\text{distance in raw}\\\text{data dimension}}} \Big)^2
    \label{classical-mds}
\end{equation}
The \ul{advantage} of MDS is it is more robust to outliers as the distance will be preserved (while PCA can collapse on the same axis), while the \ul{disadvantage} is that it is much more computational expensive than PCA. Furthermore, it is not convex and sensitive to initialization.\\
\textbf{Metric MDS} One can generalize the distance comparison in (\ref{classical-mds}) and obtain the heuristics for nonlinear dimension reduction algorithms. Namely,
\begin{equation}
    f(Z) = \sum_{i=1}^n\sum_{j=i+1}^n d_3(d_2(z_i,z_j),d_1(x_i,x_j))
\end{equation}
 \textbf{tSNE} The intuition for tSNE is to construct ``ground truth'' similarity metric $p_{ij}$ as \textit{summary statistics} for high dimensional space, and minimize the KL with target metric $q_{ij}$ measure constructed in low dimensional embedding space to enforce low\&high dimensional data structure similarity \cite{wang2021understanding}. From an \textit{metric MDS point of view}, tSNE is a special metric MDS with $d_1$ and $d_2$ being $p_{ij}$ and $q_{ij}$ in probabilistic distance measurement, $d_3$ being the comparison of matching $d_1(\cdot,\cdot)$ and $d_2(\cdot,\cdot)$ using a divergence measure.  In general, tSNE is a metric MDS method matching small distances.\\
\textbf{UMAP} UMAP may also be considered as a special case for metric MDS in the sense that $d_3$ is the cross entropy loss except $d_1(\cdot,\cdot)$ and $d_2(\cdot,\cdot)$ summarizes fuzzy topological representational information \cite{mcinnes2018umap}. Compared to tSNE,
UMAP scales well with the dataset size and improves the running time for tSNE algorithm \cite{wang2021understanding}. The general graph layout procedure is same as tSNE, except UMAP make multiple fine improvements. For example the loss function also encourages global structure preservation and the low dimensional similarity $\frac{1}{(1+a\|z_i-z_j\|_2^{2b})}$ also generalizes the student t-distribution $(1+\|z_i-z_j\|^2)^{-1}$ in tSNE.

